\documentclass[10pt,letterpaper, journal]{IEEEtran}

\usepackage{amsmath}
\usepackage{amsfonts}
\usepackage{amssymb}
\usepackage{graphicx}
\usepackage[T1]{fontenc}    
\usepackage{url}            
\usepackage{booktabs}       
\usepackage{nicefrac}       
\usepackage{microtype}      
\usepackage{bbm}
\usepackage{amsthm}
\usepackage{cite}
\usepackage{subcaption,booktabs}
\usepackage{epsfig}
\newcommand{ \mb}[1]{\mathbf{#1}}

\usepackage{multirow}
\usepackage{color,soul}
\usepackage{algorithm}
\usepackage{algorithmic, bm, textcomp}
\usepackage{etoolbox}
\usepackage{tabularx, booktabs}
\usepackage{xargs}                      
\usepackage[pdftex,dvipsnames]{xcolor} 
\usepackage{xr-hyper}
\usepackage[colorlinks,
linkcolor=blue,
anchorcolor=blue,
citecolor=blue
]{hyperref}

\newcolumntype{P}[1]{>{\centering\arraybackslash}p{#1}}
\newcolumntype{Y}{>{\centering\arraybackslash}m{0.9cm}}
\newcolumntype{G}{>{\bfseries\centering\arraybackslash}m{1.8cm}}
\newcommand{\rk}{{\rm rank}}
\newcommand{\Fr}{{\rm F}}

\newcommand{\spann}{{\rm span}}

\newcommand{\csupp}{\rm csupp}
\newcommand{\vect}{\text{vec}}
\newcommand{\col}{\text{col}}

\newcommand{\cO}{\mathcal{O}}

\usepackage[colorinlistoftodos]{todonotes}
\newcommandx{\unsure}[2][1=]{\todo[linecolor=red,backgroundcolor=red!25,bordercolor=red,#1]{#2}}
\newcommandx{\change}[2][1=]{\todo[linecolor=blue,backgroundcolor=blue!25,bordercolor=blue,#1]{#2}}
\newcommandx{\info}[2][1=]{\todo[linecolor=OliveGreen,backgroundcolor=OliveGreen!25,bordercolor=OliveGreen,#1]{#2}}
\newcommandx{\improvement}[2][1=]{\todo[linecolor=Plum,backgroundcolor=Plum!25,bordercolor=Plum,#1]{#2}}

\def\[#1\]{{$#1$}}
\def\@#1\@{{\mathbf{#1}}}
\def\b#1{{\mathbf{#1}}}
\def\c#1{{\mathcal{#1}}}

\newtheorem{theorem}{Theorem}[]
\newtheorem{lemma}[theorem]{Lemma}
\usepackage{color}
\usepackage[font=footnotesize,labelfont=bf]{caption}

\theoremstyle{definition}
\newtheorem{definition}{Definition D.\ignorespaces}[]

\newtheorem*{remark*}{Remark}

\newcommand{\jdh}[1]{}
\renewcommand{\jdh}[1]{{\color{green}{#1}}} 
\newcommand{\xl}[1]{}
\renewcommand{\xl}[1]{{\color{red}{#1}}} 
\newcommand{\xladd}[1]{}
\renewcommand{\xladd}[1]{{\color{green}{#1}}} 
\newcommand{\remove}[1]{}

\makeatletter
\g@addto@macro \small {%
\setlength\abovedisplayskip{5pt plus 2pt minus 2pt}%
\setlength\belowdisplayskip{5pt plus 2pt minus 2pt}%
}
\makeatother

\makeatletter
\g@addto@macro \scriptsize {%
\setlength\abovedisplayskip{6pt plus 3pt minus 3pt}%
\setlength\belowdisplayskip{6pt plus 3pt minus 3pt}%
}
\makeatother

\makeatletter
\g@addto@macro \footnotesize {%
\setlength\abovedisplayskip{5pt plus 2pt minus 2pt}%
\setlength\belowdisplayskip{5pt plus 2pt minus 2pt}%
}
\makeatother

\makeatletter
\g@addto@macro \normalsize {%
\setlength\abovedisplayskip{5pt plus 2pt minus 2pt}%
\setlength\belowdisplayskip{5pt plus 2pt minus 2pt}%
}
\makeatother

\newcommand{\edit}[1]{\textcolor{black}{#1}}
\newcommand{\RR}{\mathbb{R}}

\begin{document}

\title{A Dictionary-Based Generalization of Robust PCA with Applications to Target Localization in Hyperspectral Imaging}\vspace{-2pt}
\author{\IEEEauthorblockN{Sirisha Rambhatla, Xingguo Li, Jineng Ren and Jarvis Haupt}\\
	\IEEEauthorblockA{Department of Electrical and Computer Engineering,\\ University of Minnesota -- Twin Cities, Minneapolis, MN-55455\\
		{\tt \{rambh002, lixx1661, renxx282, jdhaupt\}@umn.edu}. }}

\author{Sirisha Rambhatla\[^\dagger\],~\IEEEmembership{Student Member,~IEEE,}
	Xingguo Li\[^\ddagger\],\IEEEmembership{}
	Jineng Ren\[^\dagger\],~\IEEEmembership{Student Member,~IEEE,}
	and~\\ Jarvis Haupt\[^\dagger\],~\IEEEmembership{Senior Member,~IEEE} \vspace*{-10pt}
	\thanks{\[^\dagger\]Department of Electrical and Computer Engineering, University of Minnesota, Minneapolis, MN, 55455, USA, e-mail: {\tt \{rambh002, renxx282, jdhaupt\}@umn.edu}, respectively. \[^\ddagger\]Computer Science Department, Princeton University, Princeton, NJ 08540, USA,  email: {\tt{xingguol@cs.princeton.edu}}. The work was done when S. Rambhatla was at the University of Minnesota-Twin Cities.} 
	\thanks{This work was supported by the DARPA YFA, Grant N66001-14-1-4047. Preliminary versions appeared in the proceedings of the 2016 IEEE Global Conference on Signal  \& Information Processing (GlobalSIP), 2017 Asilomar Conference on Signals, Systems, \& Computers, and the 2018 IEEE International Conference on Acoustics, Speech \& Signal Processing (ICASSP).}
		\vspace{-12pt}
}

\maketitle
\begin{abstract}
We consider the decomposition of a data matrix assumed to be a superposition of a low-rank matrix and a component which is sparse in a known dictionary, using a convex demixing method. We consider two sparsity structures for the sparse factor of the dictionary sparse component, namely entry-wise and column-wise sparsity, and provide a unified analysis, encompassing both undercomplete and the overcomplete dictionary cases, to show that the constituent matrices can be successfully recovered under some relatively mild conditions on incoherence, sparsity, and rank. We leverage these results to localize targets of interest in a hyperspectral (HS) image based on their spectral signature(s) using the \textit{a priori} known characteristic spectral responses of the target. We corroborate our theoretical results and analyze target localization performance of our approach via experimental evaluations and comparisons to related techniques.

\end{abstract}
\vspace{-2pt}
\begin{IEEEkeywords}
Low-rank, dictionary learning, target localization, Robust PCA, hyperspectral imaging, sparse representation.
\end{IEEEkeywords}

\vspace*{-8pt}
\section{Introduction} 
\vspace{-1pt}
Leveraging the structure of a given dataset is at the heart of machine learning and data analysis tasks. \textit{A priori} knowledge about the structure often makes the problem well-posed, leading to improvements in the solutions. Perhaps the most common of these, one that is often encountered in practice, is approximate low-rankness of the dataset, which is exploited by the popular principal component analysis (PCA)\cite{Jolliffe02}. The low-rank structure encapsulates the model assumption that the data in fact spans a lower dimensional subspace than the ambient dimension of the data. 
However, in a number of applications, the data may not be inherently low-rank, but may be decomposed as a superposition of a low-rank component, and a component which has a sparse representation in a known \textit{dictionary}. This scenario is encountered in target identification applications in hyperspectral (HS) imaging \cite{Rambhatla17,Li2018}, where the \textit{a priori} knowledge of the target \textit{signatures} (dictionary), can be leveraged for localization.

Hyperspectral (HS) imaging is an imaging modality which senses the intensities of the reflected electromagnetic waves (responses) corresponding to different wavelengths of the electromagnetic spectra, often invisible to the human eye. As the spectral response associated with an object/material is dependent on its composition, HS imaging can be used to identify the said target objects/materials via their characteristic spectra or  \textit{signature} responses, also referred to as \emph{endmembers}. 

Typical applications of HS imaging range from monitoring agricultural use of land, catchment areas of rivers and water bodies, food processing, surveillance, and climate science applications, to detecting various minerals, chemicals, and for presence of life sustaining compounds on distant planets; see \cite{Borengasser2007, Park2015, Rolnick2019tackling}, and references therein for details. However, these spectral \textit{signatures} are often highly correlated, which makes it difficult to detect regions of interest.

In this work, we present two techniques for target localization in HS images by posing it as a matrix demixing task. Here, we first analyze a matrix demixing problem where a data matrix \[ \b{M}\in \mathbb{R}^{n \times m}\] is assumed to be formed via a superposition of a low-rank component \[\b{L} \in \mathbb{R}^{n \times m}\] of rank-\[r\] for \[r < \min(n, m)\], and a dictionary sparse part \[\b{DS} \in \mathbb{R}^{n \times m} \]. Here,  the matrix \[\b{D} \in \mathbb{R}^{n \times d}\] is an \textit{a priori} known dictionary, and \[\b{S} \in \mathbb{R}^{d \times m}\] is an \edit{unknown} \textit{sparse} coefficient matrix. Specifically, we will study the following model for \[\b{M}\]:
\begin{align}
\label{Prob}
\b{M} = \b{L} + \b{DS},
\end{align}
and identify the conditions under which components \[\b{L}\] and \[\b{S}\] can be recovered given $\b{M}$ and $\b{D}$ by solving appropriate convex formulations. We then leverage these theoretical results for the target localization task in HS images; see Section~\ref{sec:intro}.

We consider the demixing problem described above for two different sparsity models on the matrix \[ \b{S}\]. First, we consider a case where \[\b{S}\] has at most \[s_e\] total non-zero entries (entry-wise sparse case), and second where \[\b{S}\] has \[s_c\] non-zero columns (column-wise sparse case). To this end, we develop the conditions under which solving 
\begin{align}\tag{{D-RPCA(E)}}
\underset{\b{L}, \b{S}}{\min~} \|\b{L}\|_* + \lambda_e \|\b{S}\|_1 ~~\text{s.t.}~~ \b{M} = \b{L} + \b{DS}, \label{Pe}
\end{align}
for the entry-wise sparsity case, and 
\begin{align}\tag{{D-RPCA(C)}}
\underset{\b{L}, \b{S}}{\min~} \|\b{L}\|_* + \lambda_c \|\b{S}\|_{1,2} ~\text{s.t.}~\b{M} = \b{L} + \b{DS} ,\label{Pc}
\end{align}
for the column-wise sparse case, will recover \[\b{L}\] and \[\b{S}\] for regularization parameters \[\lambda_e\geq 0\] and \[\lambda_c \geq 0\], respectively, given the data \[\b{M}\] and the dictionary \[\b{D}\]. The known dictionary \[\b{D}\] here can be overcomplete (\textit{fat}, i.e., \[d>n\]) or undercomplete (\textit{thin}, i.e., \[d\leq n\]). Here, ``D-RPCA'' refers to ``Dictionary based Robust Principal Component Analysis'', while ``E'' and ``C'' indicate the entry-wise and column-wise sparsity patterns, respectively. In addition, \[\|.\|_* \], \[\|.\|_1\], and \[\|.\|_{1,2}\] refer to the nuclear norm, \[\ell_1\]- norm of the vectorized matrix, and \[\ell_{1,2}\] norm (sum of column \[\ell_2\] norms), respectively, which serve as convex relaxations of rank, sparsity, and column sparsity inducing regularization, respectively.

These two types of sparsity patterns capture different structural properties of the dictionary sparse component. The entry-wise sparsity model allows individual data points to span low-dimensional subspaces, still allowing the dataset to span the entire space. While in the column-wise sparsity setting, the component \[\b{DS}\] is also column-wise sparse. As a result, this model effectively captures the structured (dictionary dependent) corruptions in the otherwise low-rank structured columns of \[\b{M}\]. Note that the columns of \[\b{S}\] are not restricted to be sparse in the column-wise sparsity model.
\vspace*{-6pt}
\subsection{Background}
\vspace{-1pt}
A wide range of problems can be expressed in the form described in \eqref{Prob}. Perhaps the most celebrated of these is principal component analysis (PCA) \cite{Jolliffe02}, which can be viewed as a special case of \eqref{Prob}, with the matrix \[\b{D}\] set to zero. In the absence of \[\b{L}\], the problem reduces to that of sparse recovery \cite{Natarajan95, Donoho01, Candes05}; see \cite{Rauhut10} and references therein for an overview of related works. Further, the popular framework of Robust PCA tackles a case when the dictionary \[\b{D}\] is identity \cite{Candes11, Chandrasekaran11}, i.e., \[\b{D} = \b{I}\] for an identity matrix \[\b{I}\], Outlier Pursuit (OP) \cite{Xu2010} (\[\b{D} = \b{I}\] and \[\b{S}\] is column-wise sparse,) and others  \cite{Zhou2010, Ding11, Wright13, Chen13, Li15, Li15c, Li15b, Li16_refined, Li2016efficient}. 

The model in \eqref{Prob} is also closely related to the one in \cite{Mardani2012}, which explores the overcomplete dictionary setting with applications to network traffic anomaly detection. However, the analysis therein applies to a case where the  \[\b{D}\] is overcomplete with orthogonal rows, and the coefficient matrix \[\b{S}\] has a small number of non-zero elements per row and column, which may be restrictive assumptions in some applications. To this end, in recent works we analyze the extension of \cite{Mardani2012} to include a case where the dictionary has more rows than columns, i.e., is \textit{thin}, while removing the orthogonality constraint for both the \textit{thin} and the \textit{fat} dictionary cases, for entry-wise sparsity  \cite{Rambhatla2016} and column-wise sparsity \cite{Li2018} cases, respectively.

 In particular, the entry-wise case \eqref{Prob} is propitious in a number of applications.  For example, it can be used for target identification in hyperspectral imaging \cite{Rambhatla17,Li2018}, and in topic modeling applications to identify documents with certain properties, on similar lines as \cite{Min2010}. Further, in source separation tasks, a variant of this model was used in singing voice separation in \cite{Huang2012, Sprechmann2012}.  In addition, we can also envision source separation tasks where \[\b{L}\] is not low-rank, but can in turn be modeled as being sparse in a known \cite{Starck2005} or unknown \cite{Rambhatla2013} dictionary. The column-wise setting, model \eqref{Prob} is also closely related to outlier identification \cite{Xu2010,Li15, Li15c, Rahmani2015}, which is motivated by a number of contemporary ``big data'' applications. Here, the sparse matrix $\b{S}$ (known as ``outliers'') can be used to identify malicious responses in collaborative filtering applications \cite{mehta2008attack}, finding anomalous patterns in network traffic \cite{lakhina2004diagnosing} or estimating visually salient regions of images \cite{Itti:98, Harel:06, Liu:07}; see also \cite{Lerman2018}. 

In Section~\ref{sec:intro} we also analyze and demonstrate the application of the model shown in \eqref{Prob} for a hyperspectral (HS) demixing task. HS image analysis using sparse recovery-based techniques were explored in \cite{Moudden2009, Bobin2009, Kawakami2011, Charles2011}. Applications of compressive sampling have been explored in \cite{Golbabaee2010,Yuan2015}, while \cite{Xing2012} analyzes the case where HS images are noisy and incomplete. Further, in a recent work \cite{Giampouras2016}, the authors study a case where \[\b{L}\] is absent and the sparse matrix \[\b{S}\] is also low-rank for the demixing task \eqref{Prob}.  However, the techniques discussed above focus on identifying all materials in a given HS image. Although sparsity-based target detection was considered in \cite{Chen2011,Zhang2014,Zhang2017,Zhu2019}, the approaches use training samples from both background and the targets for detection, while possessing no recovery guarantees. However, for target localization, the task is to identify only specific target(s) in a given HS image, while the background may be unknown/irrelevant. As a result, there is a need for techniques which localize targets based on their \textit{a priori} known spectral signatures; see also \cite{Nasrabadi2013} and \cite{Li2016survey}.

\vspace*{-3pt}
\subsection{Our Contributions}\label{sec:contribution}
\vspace{-1pt}
As described above, we propose and analyze a dictionary based generalization of robust PCA as shown in \eqref{Prob}. Here, we consider two distinct sparsity patterns of \[\b{S}\], i.e., entry-wise and column-wise sparse $\b{S}$, arising from different structural assumptions on the dictionary sparse component. Our specific contributions are summarized below. 

\vspace{2pt}

\noindent\textbf{Entry-wise case}: 

We make the following contributions towards guaranteeing the recovery of \[\b{L}\] and \[\b{S}\] via the convex optimization problem in \ref{Pe}. First, we analyze the \textit{thin} case (i.e. \[d\leq n\]), where we assume that the matrix \[\b{S}\] has at most {\[s_e = \cO(\tfrac{m}{r})\]} non-zero elements \textit{globally}, i.e.,\[\|\b{S}\|_0 \leq s_e\], where \[\|\cdot\|_0\] represents the number of non-zero entries in \[\b{S}\].
Next, for the \textit{fat} case, we first extend the analysis presented in \cite{Mardani2012} to eliminate the orthogonality constraint on the rows of the dictionary \[\b{D}\]. Further, we relax the sparsity constraints required by \cite{Mardani2012} on rows and columns of the sparse coefficient matrix \[\b{S}\], to study the case when \[\|\b{S}\|_0 \leq s_e\] with at most {\[k = \cO(d/\log(n))\]} non-zero elements per column \cite{Rambhatla2016}. {Hence, we provide a unified analysis for both the \textit{thin} and the \textit{fat} case, making the model \eqref{Prob} amenable to a wide range of applications.} 

\vspace{2pt}
\noindent\textbf{Column-wise case}: {We propose and analyze a dictionary based generalization of \emph{Outlier Pursuit} (OP) \cite{Xu2010},  wherein the coefficient matrix \[\b{S}\] admits a column sparse structure, referred to as ``outliers''; see also \cite{Li2018}.  Note that, in this case there is an inherent ambiguity regarding the recovery of the true component pair \[(\b{L}, \b{S})\] corresponding to the low-rank part and the dictionary sparse component, respectively.  Specifically, any pair \[(\b{L}_0, \b{S}_0)\] satisfying $\b{M} = \b{L}_0 + \b{D}\b{S}_0 = \b{L}+ \b{D}\b{S}$, where $\b{L}_0$ and $\b{L}$ have the same column space, and $\b{S}_0$ and $\b{S}$ have the identical column support, is a solution of \ref{Pc}. To this end, we develop the sufficient conditions for the convex optimization task \ref{Pc} to recover the column space of the low-rank component \[\b{L}\], while identifying the outlier columns of \[\b{S}\]; see Section~\ref{sec:opt_col} for details. Here, the difference between  \ref{Pc} and OP  being the inclusion of the known dictionary. Next, we demonstrate the advantages of leveraging the knowledge of the dictionary via phase transitions in rank and sparsity for recovery of the outlier columns. Specifically, we show that as compared to OP, \ref{Pc} works for potentially higher ranks of \[\b{L}\], when \[s_c\] is a fixed proportion of \[m\].}

{
\vspace{2pt}
\noindent\textbf{The \textit{thin} dictionary case -- an interesting result}: \cite{Mardani2012} suggests that when the dictionary is \textit{thin}, i.e., \[d<n\], one can envision a pseudo-inverse based technique wherein we pre-multiply both sides in \eqref{Prob} with the Moore-Penrose pseudo-inverse \[\b{D}^{\dagger} \in \RR^{d \times n}\], i.e., $\b{D}^{\dagger} \b{D} = \b{I}$ (this is not applicable for the \textit{fat} case due to the non-trivial null space). This operation leads to a formulation which resembles the robust PCA (RPCA) \cite{Candes11, Chandrasekaran11} model for the entry-wise case and Outlier Pursuit (OP) \cite{Xu2010} for the column-wise case, i.e.,

\begin{minipage}{0.24\textwidth}
\begin{align}\tag{{RPCA}$^\dagger$}
\hspace{-0.65in}\b{D}^\dagger\b{M} = \b{D}^\dagger\b{L} + \b{S}, \hspace{-0.55in} \label{RPCA}
\end{align}
\end{minipage}\hspace{-0.15in}
\begin{minipage}{0.24\textwidth}
\begin{align}\tag{{OP}$^\dagger$}
\b{D}^\dagger\b{M} = \b{D}^\dagger\b{L} + \b{S}. \hspace{-0.15in} \label{OP}
\end{align}
\end{minipage}

\vspace{0.05in}

\noindent An interesting finding of our work is that although this transformation algebraically reduces the entry-wise and column-wise sparsity cases to Robust PCA and OP settings, respectively, the specific model assumptions of Robust PCA and OP may not hold  for all choices of dictionary size \[d\] and rank \[r\]. Specifically, we find that in cases where \[d < r\], this pre-multiplication may not lead to a ``low-rank'' \[\b{D}^\dagger\b{L}\]. This suggests that the notion of ``low'' or ``high'' rank is relative to the maximum possible rank of \[\b{D}^\dagger\b{L}\], which in this case is \[\min(d, r)\]. Therefore, if \[d<r\], \[\b{\b{D}^\dagger\b{L}}\] can be full-rank, and the low-rank assumptions of RPCA and OP may no longer hold. As a result, these two models (the pseudo inversed case and the current work) cannot be used interchangeably for the thin dictionary case. We corroborate these via experimental evaluations presented in Section~\ref{sec:simulations}\footnote{The code is made available at \texttt{github.com/srambhatla/Diction ary-based-Robust-PCA}, and the results are reproducible.}.} 

\vspace{0.05in}
\noindent\textbf{Techniques for HS demixing}: Building on our theoretical results, we present two techniques for target detection in a HS image, depending upon different sparsity assumptions on the matrix \[\b{S}\]. Our techniques operate by forming the dictionary \[\b{D}\] using the \textit{a priori} known spectral signatures of the target of interest, and leveraging the approximate low-rank structure of the data matrix \[\b{M}\] \cite{Rambhatla2016, Li2018, Rambhatla17}. We then analyze the performance of these techniques via extensive experimental evaluations on real-world demixing tasks over different datasets and dictionary choices, and compare the performance of the proposed techniques with related works.

The choice of a particular sparsity model, i.e., \textit{entry-wise} and  \textit{column-wise} for this task depends on the properties of the dictionary matrix \[\b{D}\]. In particular, if the target signature admits a sparse representation in the dictionary, entry-wise sparsity structure is preferred. This is likely to be the case when the dictionary is overcomplete (\[n<d\]) or \textit{fat}, and also when the target spectral responses admit a sparse representation in the dictionary. On the other hand, the column-wise sparsity structure is amenable to cases where the representation can use all columns of the dictionary. This potentially arises in the cases when the dictionary is undercomplete (\[n\geq d\]) or \textit{thin}. Note that, in the column-wise sparsity case, the non-zero columns need not be sparse themselves. The applicability of these two modalities is also exhibited in our experimental analysis; see Section~\ref{sec:exp} for further details. 

\begin{figure}[!t]
  \centering
  \begin{tabular}{c}
    \includegraphics[width=0.26\textwidth]{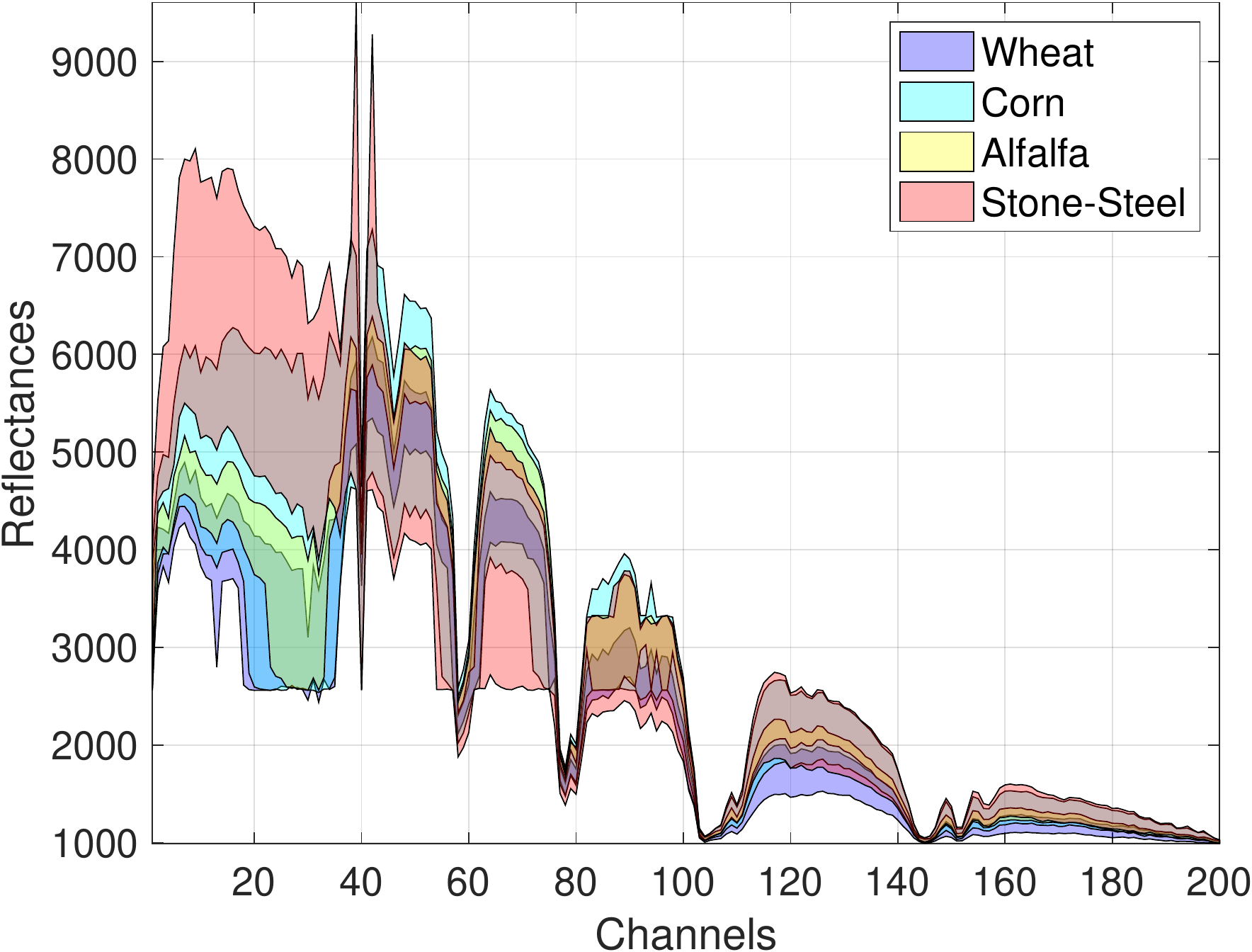}
    \end{tabular}
      \vspace{-8pt}
  \caption{Correlated spectral signatures. The spectral signatures of even different materials are highly correlated. Shown here are spectral signatures of classes from the Indian Pines dataset \cite{HSdat}. Here, the shaded region shows the lower and upper ranges of reflectance values the signatures take. }
  \label{fig:data_corr} 
  \vspace{-18pt}
\end{figure}
\vspace{0.05in}
\noindent\textbf{Demixing Despite Correlated Signatures}:  Since the spectral signatures of even distinct classes are highly correlated to each other this demixing task is particularly challenging. For instance, we plot the spectral signatures of different classes of the ``Indian Pines'' Dataset \cite{HSdat} in Fig.~\ref{fig:data_corr}. The shaded region here shows the upper and lower ranges of different classes. For instance, in Fig.~\ref{fig:data_corr} we observe that the spectral signature of the ``Stone-Steel'' class is similar to that of class ``Wheat''. This correlation between the spectral signatures of different classes results in an approximate low-rank structure of the data, captured by the low-rank component \[\b{L}\], while the dictionary-sparse component \[\b{DS}\] is used to identify the target of interest; see also Fig~\ref{fig:sing_val_data}. We specifically show that such a  decomposition successfully localizes the target despite the high correlation between spectral signatures. It is worth noting that although we consider \textit{thin} dictionaries (\[n\geq d\]) for the purposes of this demixing task, our theoretical results are also applicable for the \textit{fat} case (\[n< d\]) \cite{Rambhatla2016},\cite{Li2018}.
\remove{Interestingly, the seemingly simple change in the sparsity structure from entry-wise to column-wise case leads to significant differences in the conditions and recovery guarantees for the two cases. For example, as discussed above, we can provide explicit conditions under which solving \ref{Pe} recovers the two components \textit{exactly}. However, we can only guarantee that any solution pair \[(\b{L}, \b{S})\] recovered by solving \ref{Pc} has the same column space $\c{U}$ and column support $\c{I}_{\c{S}_c}$ due to its inherent ambiguity. Consequently, the entry-wise case requires more stringent conditions on the known dictionary \[\b{D}\] as compared to the column-wise case (discussed shortly).}

The rest of the paper is organized as follows\footnote{\noindent\textbf{Notation:} Given a matrix \[\b{X}\] and vector \[\b{v}\], we use \[\|\b{X}\|:= \sigma_{\max}(\b{X})\] for the spectral norm, where \[\sigma_{\max}(\b{X})\] denotes the maximum singular value of the matrix, \[\|\b{v}\|_\infty = \underset{i}{\max} ~|\b{v}_i|\], \[\|\b{X}\|_\infty := \underset{i,~j}{\max} |\b{X}_{ij}|\], \[\|\b{X}\|_{\infty, \infty} = \underset{\|\b{v}\|_\infty = 1}{\max}~ \|\b{X}\b{v}\|_{\infty} = 
\underset{i}{\max} ~\|\b{e}^\top_i\b{X}\|_1 \], and \[\|\b{X}\|_{\infty,2} := \underset{i}{\max} \|\b{X}\b{e}_i\|\]. Here, \[\b{X}_{i,j}\] denotes the \[(i,j)\] element of \[\b{X}\] and \[\b{e}_i\] denotes the canonical basis vector with \[1\] at the \[i\]-th location. We also use \[\|\cdot\|\] to denote the \[\ell_2\]-norm in case of vectors and spectral norm for matrices.}. We formalize the problem and describe various considerations on the structure of the component matrices in Section~\ref{sec:prelim}. In Section~\ref{sec:main_result}, we present our main theorems for the entry-wise and column-wise cases along with discussion on the implication of the results, followed by an outline of the analysis in Section~\ref{sec:proof_sketch}. Numerical evaluations on synthetic data are provided in Section~\ref{sec:simulations}, while we explore the application to target localization in HS images in Section~\ref{sec:intro}. Finally, we summarize our contributions and conclude this discussion in Section~\ref{sec:conclusion} with future directions. 

\vspace*{-2pt}
\section{Preliminaries} \label{sec:prelim}
We start formalizing the problem set-up and introduce model parameters pertinent to our analysis. We begin our discussion with our notion of optimality for the two sparsity modalities; we also summarize the notation in Table~\ref{tab:notation} in the appendix.
\vspace*{-8pt}
\subsection{Optimality of the Solution Pair}\label{sec:opt_col}
For the entry-wise case, we recover the low-rank component \[\b{L}\], and the sparse coefficient matrix \[\b{S}\], given the dictionary \[\b{D}\], and data \[\b{M}\] generated according to the model described in \eqref{Prob}. Recall that \[s_e\] is the global sparsity, \[k\] denotes the number of non-zero entries in a column of \[\b{S}\] when the dictionary is \textit{fat}. 

In the the column-wise sparsity setting, due to the inherent ambiguity in the model \eqref{Prob}, as discussed in Section~\ref{sec:contribution}, we can only hope to recover the column-space for the low-rank matrix and the identities of the non-zero columns for the sparse matrix. Therefore, in this case any solution in the \textit{Oracle Model} (defined below) is deemed to be optimal.
\vspace{-0pt}
\begin{definition}[Oracle Model for Column-wise Sparsity Case]\label{def:oracle}
 	\vspace*{-0pt}

Let the pair \[(\b{L}, \b{S})\] be the matrices forming the data \[\b{M}\] as per \eqref{Prob}, and define the oracle model \[\{ \b{M}, \c{U}, \c{I}_{\c{S}_c}  \}\]. Then, any pair \[(\b{L}_0, \b{S}_0)\] is in the \emph{Oracle Model} \[\{ \b{M}, \c{U}, \c{I}_{\c{S}_c}  \}\], if \[\c{P}_{\c{U}}(\b{L}_0) = \b{L}\], \[\c{P}_{\c{S}_c}(\b{D} \b{S}_0) = \b{D} \b{S}\] and \[\b{L}_0 + \b{D} \b{S}_0 = \b{L} + \b{D} \b{S} = \b{M}\] hold simultaneously, where \[\c{P}_{\c{U}}\] and \[\c{P}_{\c{S}_c}\] are projections onto the column space \[\c{U}\] of \[\b{L}\] and column support \[\c{I}_{\c{S}_c} \] of \[\b{S}\], respectively.
 	\vspace*{-5pt}
\end{definition}
\vspace*{-22pt}
\subsection{Conditions on the Dictionary}
We require that the dictionary \[\b{D}\] follows the \textit{generalized frame property} (GFP) defined as follows. 
\vspace{-0pt}
\begin{definition}\label{frame}
 	\vspace*{-0pt}
	A matrix \[\b{D}\] satisfies the \emph{generalized frame property} (GFP), on vectors \[\b{v} \in \c{R}\], if for any fixed vector \[\b{v}\in \c{R}\] where \[\b{v}\neq \b{0}\] and some \[\c{R}\], we have
	\begin{align*}
	\alpha_\ell\|\b{v}\|^2_2 \leq \|\b{Dv}\|^2_2 \leq \alpha_u\|\b{v}\|^2_2, 
	\end{align*}
	where \[\alpha_\ell\] and \[\alpha_u\] are the lower and upper \emph{generalized frame bounds} with \[0 < \alpha_\ell \leq \alpha_u < \infty \].
	 	\vspace*{-0pt}
\end{definition}
\vspace{-0pt}
\noindent The GFP shown above is met as long as the vectors \[\b{v}\] are not in the null-space of the matrix \[\b{D}\] for finite \[ \|\b{D}\| \].  Therefore, for the \textit{thin} dictionary setting \[d \leq n\] for both entry-wise and column-wise cases \[\c{R}\] can be the entire space, and GFP is satisfied as long as \[\b{D}\] has full column rank. For example,  \[\b{D}\] being a \textit{frame}\cite{Duffin1952} suffices; see also \cite{Heil2013}. On the other hand, for the \textit{fat} dictionary setting, we need the space \[\c{R}\] to have a union-of-subspace structure such that GFP is met for both the entry-wise and column-wise sparsity cases. Specifically, for the entry-wise sparsity case, we also require that the frame bounds \[\alpha_u\] and \[\alpha_\ell\] be close to each other. To this end, we assume that \[\b{D}\] satisfies the \textit{restricted isomtery property} (RIP) \cite{Candes05} of order \[k = \cO(d/\log(n))\] with a \emph{restricted isometric constant} (RIC) of \[\delta\] in this case, and that \[\alpha_u = (1 + \delta)\] and \[\alpha_\ell = (1 - \delta)\].

\vspace*{-8pt}
\subsection{Relevant Subspaces}\label{sec:params}

We now define the subspaces relevant for our discussion. For the following discussion, let the pair \[(\b{L_0}, \b{S_0})\] denote the solution to \ref{Pe} in the entry-wise sparse case. Further, for the column-wise sparse setting, let \[(\b{L_0}, \b{S_0})\] denote a solution pair in the oracle model \[\{ \b{M}, \c{U}, \c{I}_{\c{S}_c} \}\] \textcolor{blue}{\textbf{D.\ref{def:oracle}}}, obtained by solving \ref{Pc}. For the low-rank matrix \[\b{L}\], let the compact singular value decomposition (SVD) be defined as
\vspace{-0pt}
	\begin{align*}
	\b{L} = \b{U\Sigma V^\top},
	\end{align*}
where \[\b{U}\in\RR^{n \times r}\] and \[\b{V}\in\RR^{m \times r}\] are the left and right singular vectors of \[\b{L}\], respectively, and \[\b{\Sigma}\] is the diagonal matrix with singular values on the diagonal. Here, matrices \[\b{U}\] and \[\b{V}\] each have orthogonal columns, and the non-negative entries \[\b{\Sigma}_{ii} = \sigma_{i}\] are arranged in descending order. We define \[\c{L}\] as the linear subspace consisting of matrices spanning the same row or column space as \[\b{L}\], i.e., for \[\b{W}_1 \neq 0\] or \[\b{W}_2 \neq 0\],
\begin{align*}
\c{L} := \{ \b{UW}_1^\top + \b{W}_2\b{V}^\top, \b{W}_1  \in \RR^{m \times r}, \b{W}_2 \in \RR^{n \times r}\}.
\end{align*}

Next, let \[\c{S}_e\] (\[\c{S}_c\] for the column-wise sparsity setting) be the space spanned by \[d \times m\] matrices with the same non-zero support (column support, denoted as $\csupp(\cdot)$) as \[\b{S}\], and let \[\c{I}_{\c{S}_c}\] denote the index set containing the non-zero column index set of \[\b{S}\] for the column-wise case, then we denote the space spanned by the dictionary sparse component \[\c{D}\] as
\vspace{-0pt}
\begin{align*}
\c{D} := \{ \b{DH}\}, 
\end{align*}
where \[\b{H} \in \c{S}_e\] for entry-wise case and \[\csupp(\b{H}) \subseteq \c{I}_{\c{S}_c}\] for column-wise case.
Also, we denote the corresponding complements of the spaces described above by appending `$\perp$'.  In addition, we use calligraphic `\[\c{P}_{\c{G}}(\cdot)\]' to denote the projection operator onto a subspace \[\c{G}\], and `\[\b{P}_{\b{G}}\]' to denote the corresponding projection matrix.  For instance, we define \[\c{P}_{\c{U}}(\cdot)\] and \[\c{P}_{\c{V}}(\cdot)\] as the projection operators corresponding to the column space \[\c{U}\] and row space \[\c{V}\] of the low-rank component \[\b{L}\]. Therefore, for a given matrix \[\b{X} \in \RR^{n \times m}\], 
\begin{align*}
\c{P}_{\c{U}}(\b{X}) =\b{P}_{\b{U}}\b{X} ~\text{and}  ~\c{P}_{\c{V}}(\b{X}) =\b{X}\b{P}_{\b{V}},
\end{align*}
where \[\b{P}_{\b{U}} = \b{UU^\top}\] and \[\b{P}_{\b{V}} = \b{VV^\top}\]. With this, the projection operators onto, and orthogonal to, the subspace \[\c{L}\] are respectively defined as
\begin{align*}
&\c{P}_{\c{L }} (\b{X}) = \b{P_U}\b{X} + \b{X} \b{P_V} -   \b{P_U}\b{X}\b{P_V}, 
\end{align*}
and
\begin{align*}
&\c{P}_{\c{L}^\perp}(\b{X}) =  (\b{I} - \b{P_U})\b{X}  (\b{I} - \b{P_V}).
\end{align*}
\vspace{-20pt}
\subsection{Incoherence Measures and Parameters}
We employ various notions of incoherence to identify the conditions under which our procedures succeed. To this end, we first define the incoherence parameter \[\mu\], which characterizes the relationship between the low-rank part \[\b{L}\] and the dictionary sparse part \[\b{DS}\] as
\begin{align} \label{eqn:mu}
\mu := \underset{\b{Z} \in \mathcal{D} \backslash \{\b{0}\}}{\max} \tfrac{\|\mathcal{P}_{\mathcal{L}}(\b{Z})\|_{\rm F}}{\|\b{Z}\|_{\rm F}}.
\end{align}
The parameter \[\mu \in [0, 1]\] is the measure of degree of similarity between the low-rank part and the dictionary sparse component. Here, a larger \[\mu\] implies that the dictionary sparse component is close to the low-rank part, while a small \[\mu\] indicates otherwise. In addition, we also define the parameter \[\beta_{\b{U}}\] as
\begin{align}\label{eqn:beta}
\b{\beta}_{\b{U}}  :=  \underset{\|\b{u}\| = 1}{\max} \tfrac{\|(\b{I} - \b{P}_{\b{U}}) \b{D}\b{u}\|^2}{\|\b{Du}\|^2},
\end{align}
which measures the similarity between the orthogonal complement of the column-space \[\c{U}\] and the dictionary \[\b{D}\].

The next two measures of incoherence can be interpreted as a ways to identify the cases where for \[\b{L}\] with SVD as \[ \b{L}= \b{U \Sigma V^\top}\]: (a) \[\b{U}\] resembles the dictionary  \[\b{D}\], and/or (b) \[\b{V}\] resembles the sparse coefficient matrix \[\b{S}\]. In these cases, the low-rank part may mimic the dictionary sparse component. To this end, similar to \cite{Mardani2012}, we define the following to measure these properties respectively as
\begin{align}\label{eqn:gamma}
\text{(a)} ~\gamma_{\b{U}}  :=  \underset{i}{\max} \tfrac{\|\b{P}_\b{U} \b{D}\b{e}_{i}\|^2}{\|\b{De}_{i}\|^2} ~{\normalsize\text{and}}~
\text{(b)}~ \gamma_{\b{V}} := \underset{i}{\max} \|\mathbf{P}_{\b{V}}\mathbf{e}_{i}\|^2.
\end{align}
Here, \[\gamma_{\b{U}} \in [0, 1]\], and achieves the upper bound when a dictionary element is exactly aligned with the column space \[\c{U}\] of \[\b{L}\]. Moreover, \[\b{\gamma}_{\b{V}}  {\in [r/m, 1]}\] achieves the upper bound when the row-space of \[\b{L}\] is ``spiky,'' i.e., a certain row of \[\b{V}\] is \[1\]-sparse, meaning that a column of \[\b{L}\] is supported by (can be expressed as a linear combination of) a column of \[\b{U}\]. The lower bound here is attained when it is ``spread-out,'' i.e., each column of \[\b{L}\] is a linear combination of all columns of \[\b{U}\]. In general, our recovery of the two components is easier when the incoherence parameters \[\gamma_{\b{U}}\] and \[\b{\gamma}_{\b{V}}\] are closer to their lower bounds.

Further, for notational convenience, we define
\begin{align}\label{eqn:xi}
\xi_e := \|\b{D}^\top \b{UV}^\top\|_\infty ~~\text{and}~
\xi_c &:= \|\b{D}^\top \b{UV}^\top\|_{\infty,2}.
\end{align}
Here, \[\xi_e\] is the maximum absolute entry of \[\b{D}^\top \b{UV}^\top\] for the entry-wise case, which measures how close columns of \[\b{D}\] are to the singular vectors of $\b{L}$. Similarly, for the column-wise case, \[\xi_c\] measures the closeness of columns of \[\b{D}\] to the singular vectors of $\b{L}$ under a column-wise maximum \[\ell_2\]-norm metric.

\vspace{-2pt}
\section{Main Results }
\label{sec:main_result}
We present the main results corresponding to each sparsity structure of $\b{S}$ in this section.
\vspace*{-8pt}
\subsection{Exact Recovery for Entry-wise Sparsity Case}
Our main result establishes the existence of a regularization parameter \[\lambda_e\] for which solving the optimization problem \ref{Pe} will recover the components \[\b{L}\] and \[\b{S}\] exactly. To this end, we will show that such a \[\lambda_e\] belongs to a non-empty interval \[[\lambda_e^{\min}, \lambda_e^{\max}]\] with \[\lambda_e^{\min}\] and \[\lambda_e^{\max}\] defined as
\begin{align}\label{eqn:lam}
\lambda_{e}^{\min} := \tfrac{1 + C_e}{1-C_e} ~\xi_e ~\text{and}~\lambda_e^{\max} := 
\tfrac{\sqrt{\alpha_\ell} (1-\mu) -\sqrt{r \alpha_u} \mu}{\sqrt{s_e}},
\end{align}
where \[0 \leq C_e < 1\] is a constant that captures the relationship between different model parameters, and is defined as
\begin{align*}
C_e := \tfrac{c}{\alpha_\ell(1 - \mu)^2 - c},
\end{align*}
and \[c\] is defined as 
\begin{align}\label{eqconst:c_entry}
\hspace{-0.16cm}{\footnotesize c := 
\begin{cases}
c_t = \tfrac{ \alpha_u \left((1 + 2 \gamma_{\b{U}} )(\min(s_e, d)  + s_e \gamma_{\b{V}} ) +2 \gamma_{\b{V}}\min(s_e, m)\right)}{2} \\
\hspace{1.2in}- \tfrac{\alpha_\ell \left(\min(s_e, d)  + s_e \gamma_{\b{V}} \right)}{2}, ~\text{for  \[d \leq n\]}, \\
c_f = \tfrac{ \alpha_u \left((1 + 2 \gamma_{\b{U}} )(k  + s_e \gamma_{\b{V}} ) +2 \gamma_{\b{V}}\min(s_e, m)\right)}{2} \\
\hspace{1.2in}- \tfrac{\alpha_\ell \left(k  + s_e \gamma_{\b{V}} \right)}{2}, \hspace{0.4in}~\text{for  \[ d > n\]} 
\end{cases}}\hspace{-10pt}
\end{align}

\vspace{-0pt}
Given these definitions, we formalize the theorem for the entry-wise case as following; a proof sketch is provided in Section~\ref{sec:pf:thm_entry}.
\vspace{-2pt}
\begin{theorem} \label{theorem_entry}
Suppose \[\b{M} = \b{L} + \b{DS}\], where $\rk(\b{L})=r$ and \[\b{S}\] has at most \[s_e\] non-zeros, i.e., \[\|\b{S}\|_0 \leq s_e \leq s_e^{\max} := \tfrac{(1 - \mu)^2}{2}\tfrac{m}{r}\]. Given \[\mu \in [0, 1)\], \[ \gamma_{\b{U}}\in [0, 1]\],  \[ \gamma_{\b{V}} \in [r/m, 1]\], \[\xi_e\] defined in \eqref{eqn:mu}, \eqref{eqn:gamma}, \eqref{eqn:xi}, and any \[\lambda_e \in [\lambda_e^{\min}, \lambda_e^{\max} ]\] with \[\lambda_e^{\max} >\lambda_e^{\min}\geq 0\] defined in \eqref{eqn:lam}, and asssuming the dictionary \[\b{D} \in \mathbb{R}^{n \times d}\] obeys the generalized frame property \textcolor{blue}{\textbf{D.\ref{frame}}} with frame bounds \[[\alpha_\ell, \alpha_u ]\], solving \ref{Pe} will recover matrices \[\b{L}\] and \[\b{S}\] in the following cases:

\noindent$\bullet$ For \[d\leq n\], \[\c{R}\] may contain the entire space 
and \[ \gamma_{\b{U}}\] follows 
\begin{align}\label{A2}
\gamma_{\b{U}} \leq
\begin{cases}
\tfrac{(1 - \mu)^2 - 2s_e \gamma_{\b{V}}}{2s_e( 1 +  \gamma_{\b{V}})}, \text{ for }     s_e \leq \min ~(d, s_e^{\max})\\
\tfrac{(1 - \mu)^2 - 2s_e \gamma_{\b{V}}}{2(d + s_e \gamma_{\b{V}})}, \text{ for }   d<s_e \leq s_e^{\max} 
\end{cases};
\end{align}

\noindent$\bullet$ For \[d> n > C_1~k \log (n)\] for a constant \[C_1\], \[\c{R}\] consists of all \[k\] sparse vectors, and \[ \gamma_{\b{U}}\] follows
\vspace{-2pt}
\begin{align}\label{A4}
\gamma_{\b{U}} \leq  \tfrac{(1 - \mu)^2 - 2s_e \gamma_{\b{V}}}{2(k + s_e \gamma_{\b{V}})}.
\end{align}
\end{theorem}
\vspace{-0pt}
Theorem~\ref{theorem_entry} establishes the sufficient conditions for the existence of { $\lambda_e$} to guarantee recovery of \[(\b{L, S} )\] for both the \textit{thin} and the \textit{fat} cases. The conditions on \[ \gamma_{\b{U}}\] dictated by \eqref{A2} and \eqref{A4}, for the thin and fat case, respectively, arise from ensuring that \[\lambda_e^{\min} \geq 0\]. Further, the condition \[\lambda_e^{\min}< \lambda_e^{\max}\], translates to the following sufficient condition on rank \[r\] in terms of the sparsity \[s_e\] for \[\mu>0\], 
\begin{align}
\label{rankSpar1}
r <& \bigg(\sqrt{\tfrac{\alpha_\ell}{\alpha_u}} \tfrac{ 1-\mu}{\mu} - \tfrac{\xi_e}{\sqrt{\alpha_u}\mu} \tfrac{1 + C_e}{1 - C_e} \sqrt{s_e}\bigg)^2,
\end{align}
for the recovery of \[(\b{L, S} )\]. This relationship matches with our empirical evaluations and will be revisited in Section~\ref{sec:exp_entry}.

For both, \textit{thin} and \textit{fat} dictionary cases, smaller incoherence measures (\[\mu\], \[ \gamma_{\b{V}}\], and \[ \gamma_{\b{U}}\]) between the low-rank part, \[\b{L}\], the dictionary, \[\b{D}\], and the sparse component \[ \b{S}\] are sufficient for recovery. Our theoretical results for the fat case are similar to \cite{Mardani2012} without its restrictions (e.g. orthogonality of rows and columns of \[\b{D}\], and sparsity requirements). By extending the analysis to thin dictionaries, we consider the worst case deterministic setting as opposed to Robust PCA analysis such as \cite{Chandrasekaran11} which imposes randomness assumptions on the components.  The algorithm works beyond these constrains in practice since we consider sufficient conditions under the worst-case deterministic setting; see Section~\ref{sec:simulations}. One sanity check is to consider the case when the low-rank part is orthogonal to the dictionary, i.e., \[\mu, \gamma_{\b{U}}, \xi_e=0\]. From \eqref{eqn:lam}, we see that the condition \[\lambda_e^{\min}< \lambda_e^{\max}\], no longer constraints rank and sparsity, and we need \[s_e \leq s_e^{\max} = \c{O}(\tfrac{m}{r})\]. However, the rank and sparsity are still restricted, i.e., with increase in rank the dictionary choice may be restricted to maintain orthogonality.

\vspace*{-8pt}
\subsection{Exact Recovery for Column-wise Sparsity Case}\label{sec:rpca_c}

Recall that we consider the oracle model in this case as described in \textcolor{blue}{\textbf{D.\ref{def:oracle}}} owing to the intrinsic ambiguity in recovery of \[(\b{L}, \b{S})\]; see our discussion in Section~\ref{sec:contribution}. To demonstrate its recoverability, the following lemma establishes the sufficient conditions for the existence of an optimal pair \[(\b{L}_0, \b{S}_0)\]. The proof is provided in Appendix~\ref{pf:lem:unique}.
\vspace{-2pt}
\begin{lemma}\label{lem:unique}
Given \[\b{M}\], \[\b{D}\], and \[\left( \c{L}, {\c{S}_c}, \c{D} \right)\], any pair \[( \b{L}_0, \b{S}_0 ) \in \{ \b{M}, \c{U}, \c{I}_{\c{S}_c}  \}\] satisfies \[\spann\{\col(\b{L}_0) \} = \c{U}\] and \[\csupp(\b{S}_0) = \c{I}_{\c{S}_c} \] if \[\mu < 1\].
\end{lemma}
\vspace{-2pt}
Analogous to the entry-wise case, we show the existence of a non-empty interval \[[\lambda_c^{\min}, \lambda_c^{\max}]\] for the regularization parameter \[\lambda_c\], for which solving \ref{Pc} recovers an optimal pair as per Lemma~\ref{lem:unique}. Here, for a constant $C_c := \tfrac{ \alpha_u}{\alpha_\ell}\tfrac{1}{{(1 - \mu)^2}}\gamma_{\b{V}}\beta_{\b{U}}$, \[\lambda_c^{\min}\] and \[\lambda_c^{\max}\] are defined as 
\begin{align}\label{eqn:lamb_c}
\hspace{-0.05in}\lambda_c^{\min} := \tfrac{\xi_c + \sqrt{r s_c \alpha_u}\mu C_c}{1 - s_cC_c}~\text{and}~
\lambda_c^{\max} := \tfrac{\sqrt{\alpha_{\ell}} (1 - \mu) - \sqrt{r \alpha_{u}}\mu}{\sqrt{s_c}}.
\end{align}

Then, our main result for the column-wise case is as follows; a proof sketch is provided in Section~\ref{sec:pf:thm_col}.
\vspace{-2pt}
\begin{theorem}\label{theorem_col}
	Suppose \[\b{M} = \b{L} + \b{D}\b{S}\] with \[(\b{L}, \b{S})\] defining the oracle model \[\{ \b{M}, \c{U}, \c{I}_{\c{S}_c} \}\], where \[\rk(\b{L})=r\], and \[|\c{I}_{\c{S}_c}|=s_c\] for \[s_c \leq s_c^{\max} := \tfrac{\alpha_\ell}{ \alpha_u\gamma_{\b{V}}}\cdot\tfrac{(1 - \mu)^2}{\beta_{\b{U}}} \]. Given \[\mu \in [0,1)\], \[\beta_{\b{U}}\], \[\gamma_{\b{V}} \in [r/m, 1]\], \[\xi_c\] defined in \eqref{eqn:mu}, \eqref{eqn:beta}, \eqref{eqn:gamma}, \eqref{eqn:xi}, and any \[\lambda_c \in [\lambda_c^{\min}, \lambda_c^{\max}]\], for \[\lambda_c^{\max} > \lambda_c^{\min} \geq 0\] defined in \eqref{eqn:lamb_c}, solving \ref{Pc} will recover a pair of components \[(\b{L}_0, \b{S}_0)  \in \{ \b{M}, \c{U}, \c{I}_{\c{S}_c}  \}\], if the space \[\c{R}\] is structured such that the dictionary \[\b{D} \in \mathbb{R}^{n \times d}\] obeys the generalized frame property \textcolor{blue}{\textbf{D.\ref{frame}}} with frame bounds \[[\alpha_\ell, \alpha_u ]\], for \[\alpha_\ell>0\].	
\end{theorem}
\vspace{-2pt}

Theorem~\ref{theorem_col} states the conditions under which the solution to the optimization problem \ref{Pc} will be in the oracle model defined in \textcolor{blue}{\textbf{D.\ref{def:oracle}}}. The condition on the column sparsity \[s_c \leq s_c^{\max}\] is a result of the constraint that \[\lambda_c^{\min}\geq0\]. Similar to \eqref{rankSpar1}, requiring  \[\lambda^{\max}_c > \lambda_c^{\min}\] leads to the following sufficient condition on the rank \[r\] in terms of the sparsity $s_c$ for \[\mu>0\],
\begin{align}\label{rankSparCol}
r <
\left(\sqrt{\tfrac{\alpha_{\ell}}{\alpha_u}}\tfrac{1 - \mu}{\mu} - \tfrac{\xi_c}{\sqrt{ \alpha_u}\mu}\sqrt{s_c}\right)^2.
\end{align}
For \[\mu=0\] the conditions are similar to the entry-wise case, namely, that \[s_c \leq s_c^{\max}\]. 
Moreover, suppose that \[\alpha_{l}\] and  \[\alpha_{u} \] are both close to \[1\], which can be easily met by a tight frame when \[d<n\], or a RIP type condition when \[d>n\]. Then, if \[\tfrac{(1-\mu)^2}{\beta_{\b{U}}}\] is a constant, since \[\gamma_{\b{V}} = \Theta(\tfrac{r}{m})\], we have that \[s_c^{\max} = \c{O}(\tfrac{m}{r})\]. This is of the same order with the upper bound of \[s_c\] in the Outlier Pursuit (OP) \cite{Xu2010}. Our numerical results in Section~\ref{sec:simulations} further show that \ref{Pc} can be much more robust than OP, and may recover \[\{ \c{U}, \c{I}_{\c{S}_c} \}\] even when the rank of \[\b{L}\] is high and outliers \[s_c\] are a constant proportion of \[m\]. 
\vspace{2pt}

\noindent\emph{Remark:} In essence, Theorems~\ref{theorem_entry} and \ref{theorem_col} guarantee recovery of the components as long as the incoherence parameters, \[\mu\], \[ \gamma_{\b{V}}\], and \[ \gamma_{\b{U}}\] are small. As stated in Section~\ref{sec:params}, these parameters measure if the low-rank component and the dictionary sparse component can be teased apart from the given data. Specifically, here \[\mu\] measures how close the low-rank component is to the dictionary sparse component. Both \[\gamma_{\b{U}}\] and \[\beta_{\b{U}}\] measure how close the column space of the low-rank part \[\c{U}\] is to the dictionary \[\b{D}\], while \[ \gamma_{\b{V}}\] measures if the row space of \[\b{L}\] is sparse. These measures ensure that the components can be identified successfully. Furthermore, we see that the global sparsity in the column-wise case can be higher than the entry-wise case.
\vspace*{-2pt}
\section{Proof of Main Results}
\label{sec:proof_sketch}
\subsection{Proof of Theorem~\ref{theorem_entry}}\label{sec:pf:thm_entry}
We use dual certificate construction procedure to prove the main result in Theorem.~\ref{theorem_entry}; the proofs of all lemmata used here are given in Appendix~\ref{app:entry}. To this end, we start by constructing a dual certificate for the convex problem shown in \ref{Pe}. Here, we first show the conditions the dual certificate needs to satisfy via the following lemma.
\vspace{-3pt}
\begin{lemma}
	\label{DualCert}
	If there exists a dual certificate \[\b{\Gamma} \in \RR^{n \times m }\] satisfying
		\begin{align*}
		\centering
		\begin{tabular}{ll}
\normalfont{\textbf{(C1)}} \[\c{P}_{\c{L}}(\b{\Gamma})= \b{UV^\top}\], &\hspace{-0.15in}
		\normalfont{\textbf{(C2)}} \[\c{P}_{\c{S}_e}(\b{D^\top \Gamma})= \lambda_e ~\text{sign}(\b{S}_0)\],\\
		\normalfont{\textbf{(C3)}} \[\|\c{P}_{\c{L}^\perp} (\b{\Gamma})\| < 1\], ~\textit{and} &\hspace{-0.15in}
		\normalfont{\textbf{(C4)}} \[\|\c{P}_{\c{S}_e^\perp} (\b{D^\top\Gamma})\|_\infty < \lambda_e\]. \\
		\end{tabular}
		\end{align*}
	then the pair \[( \b{L}_0,~\b{S}_0)\] is the unique solution of \ref{Pe}.
\end{lemma}
\vspace{-3pt}
We will now proceed with the construction of the dual certificate which satisfies the conditions outlined by \textbf{(C1)-(C4)} by Lemma~\ref{DualCert}. Using the analysis similar to \cite{Mardani2012} (Section V. B.), we construct the dual certificate as 
\begin{align*}
\b{\Gamma} =\b{UV^\top} + (\b{I- P_U})\b{X} \b{(I - P_V)},
\end{align*}
for arbitrary \[\b{X} \in \RR^{n\times m}\]. 
The condition \textbf{(C1)} is readily satisfied by our choice of \[\b{\Gamma}\]. For \textbf{(C2)}, we substitute the expression for \[\b{\Gamma}\] to arrive at
\begin{align}	
\c{P}_{\c{S}_e}(\b{D^\top UV^\top}) + \c{P}_{\c{S}_e}(\b{D^\top(I- P_U)}&\b{X} \b{(I - P_V)}) \nonumber \\
&= \lambda_e ~\text{sign}(\b{S}_0). \label{eqn:c2}
\end{align}
\vspace{-2pt}
Letting $\b{Z} := \b{D^\top}(\b{I- P_U})\b{X} \b{(I - P_V)}$
and 
\begin{align*}
\b{B_{\c{S}_e}} :=\lambda_e ~\text{sign}(\b{S}_0) - \c{P}_{\c{S}_e} (\b{D^\top UV^\top}),
\end{align*}
 we can write \eqref{eqn:c2} as $\c{P}_{\c{S}_e}(\b{Z}) = \b{B_{\c{S}_e}}$. Further, we can vectorize the equation above as $\c{P}_{\c{S}_e}(\text{vec}(\b{Z})) = \text{vec}(\b{B_{\c{S}_e}})$. Let \[\b{b}_{\c{S}_e}\] be a length \[s_e\] vector containing elements of  \[\b{B}_{\c{S}_e}\] corresponding to the support of \[\b{S}_0\].
Now, note that \[\text{vec}(\b{Z})\] can be represented in terms of a Kronecker product as follows,
\begin{align*}
\text{vec}(\b{Z}) = [\b{(I - P_V)}\otimes \b{D^\top}(\b{I- P_U})] \text{vec}(\b{X}).
\end{align*}
On defining \[\b{A}:= \b{(I - P_V)}\otimes \b{D^\top}(\b{I- P_U}) \in \mathbb{R}^{md \times mn}\], we have $\text{vec}(\b{Z}) = \b{A} \text{vec}(\b{X})$.

Further, let \[\b{A_{\c{S}_e}} \in \RR^{s \times nm}\] denote the rows of \[\b{A}\] that correspond to support of \[\b{S}_0\], and let \[\b{A_{\c{S}_e^\perp}}\] correspond to the remaining rows of \[\b{A}\].  Using these definitions and results, we have $\b{A}_{\c{S}_e} \text{vec}(\b{X}) = \b{b}_{\c{S}_e}$. Thus, for conditions \textbf{(C1)} and \textbf{(C2)} to be satisfied, we need
\begin{align}
\label{vecX}
\text{vec}(\b{X}) = \b{A}_{\c{S}_e}^\top(\b{A}_{\c{S}_e}\b{A}_{\c{S}_e}^\top)^{-1}\b{b}_{\c{S}_e}.
\end{align}
Here, the following result ensures the existence of the inverse.
\vspace{-2pt}
\begin{lemma}\label{lower_sigmaMin}
	If \[\mu <1\] and \[\alpha_\ell >0 \], \[\sigma_{\min}{(\b{A}_{\c{S}_e})}\] satisfies the bound $\sigma_{\min}{(\b{A}_{\c{S}_e})} \geq \sqrt{\alpha_{\ell}} (1- \mu)$.
\end{lemma}
\vspace{-2pt}

Now, we look at the condition \textbf{(C3)} $\|\c{P}_{\c{L}^\perp} (\b{\Gamma})\| < 1$. This is where our analysis departs from \cite{Mardani2012}; we write
\begin{align*}
\|\c{P}_{\c{L}^\perp}(\b{\Gamma})\| &= \|(\b{I- P_U})\b{X} \b{(I - P_V)}\|\\
&\leq \|\b{X}\| \leq \|\b{X}\|_{\rm F}
\leq \|\b{A}_{\c{S}_e}^\top (\b{A}_{\c{S}_e}\b{A}_{\c{S}_e}^\top)^{-1}\| \|\b{b}_{\c{S}_e}\|_2,
\end{align*}
where we have used the fact that \[\|(\b{I- P_U})\|\leq1\] and \[\| \b{(I - P_V)}\| \leq 1\]. Now, as \[\b{A}_{\c{S}_e}^\top (\b{A}_{\c{S}_e}\b{A}_{\c{S}_e}^\top)^{-1}\] is the pseudo-inverse of \[\b{A}_{\c{S}_e}\], i.e., \[\b{A}_{\c{S}_e}\b{A}_{\c{S}_e}^\top (\b{A}_{\c{S}_e}\b{A}_{\c{S}_e}^\top)^{-1} = \b{I}\], we have that \[\|\b{A}_{\c{S}_e}^\top (\b{A}_{\c{S}_e}\b{A}_{\c{S}_e}^\top)^{-1}\| = 1/{\sigma_{\min}{(\b{A}_{\c{S}_e})}}\], where \[\sigma_{\min}{(\b{A}_{\c{S}_e})}\] is the smallest singular value of \[\b{A}_{\c{S}_e}\]. Therefore, we have
\begin{align}
\|\c{P}_{\c{L}^\perp}(\b{\Gamma})\| \leq \tfrac{\|\b{b}_{\c{S}_e}\|_2}{\sigma_{\min}{(\b{A}_{\c{S}_e})}}. \label{eqn:P_L1}
\end{align}
The following lemma establishes an upper bound on \[\|\b{b}_{\c{S}_e}\|_2\].
\vspace{-2pt}
\begin{lemma}\label{upper_bOmega}
An upper-bound on	\[\|\b{b}_{\c{S}_e}\|_2\] is given by $ \|\b{b}_{\c{S}_e}\|_2 \leq \lambda_e \sqrt{s_e} + \sqrt{r \alpha_u} \mu$.
\end{lemma}
\vspace{-2pt}
\noindent Combining \eqref{eqn:P_L1}, Lemma~\ref{lower_sigmaMin}, and Lemma~\ref{upper_bOmega}, we have
\begin{align}
\|\c{P}_{\c{L}^\perp}(\b{\Gamma})\| \leq \tfrac{\lambda_e \sqrt{s_e} + \sqrt{r \alpha_u} \mu.}{\sqrt{\alpha_\ell} {(1- \mu)}}. \label{eqn:P_L}
\end{align}
Now, combining \eqref{eqn:P_L} and the upper bound on \[\lambda_e\] defined in \eqref{eqn:lam}, we have that \textbf{(C3)} holds. Now, we move on to finding conditions under which \textbf{(C4)} is satisfied by our dual certificate. For this we will bound \[\|\c{P}_{{\c{S}_e}^\perp} (\b{D^\top\Gamma})\|_\infty \]. Our analysis follows the similar procedure as employed in deriving (16) in \cite{Mardani2012}, reproduced here for completeness. First, by the definition of \[\b{\Gamma}\] and properties of the \[\|.\|_\infty\] norm, we have
\begin{align}
\hspace{-0.08in}\|\c{P}_{{\c{S}_e}^\perp} (\b{D^\top\Gamma})\|_\infty 
\hspace{-0.02in}\leq\hspace{-0.02in} \|\c{P}_{{\c{S}_e}^\perp} (\b{Z}) \|_\infty \hspace{-0.03in} + \hspace{-0.02in} \|\c{P}_{{\c{S}_e}^\perp} (\b{D^\top UV})\|_\infty. \hspace{-0.02in} \label{eqn:P_DTao}
\end{align}
We now focus on simplifying the term \[\|\c{P}_{{\c{S}_e}^\perp} (\b{Z}) \|_\infty \].
By definition of \[\b{A}\], and using the fact that \[	\text{vec}(\b{Z}) = \b{A} \text{vec}(\b{X})\], we have $\c{P}_{{\c{S}_e}^\perp} (\b{Z}) = \b{A}_{{\c{S}_e}^\perp} \text{vec}(\b{X})$, which implies
\begin{align*}
\|\c{P}_{{\c{S}_e}^\perp} (\b{Z}) \|_\infty &= \|\b{A}_{{\c{S}_e}^\perp} \text{vec}(\b{X})\|_\infty\\
&= \|\b{A}_{{\c{S}_e}^\perp} \b{A}_{\c{S}_e}^\top (\b{A}_{\c{S}_e}\b{A}_{\c{S}_e}^\top)^{-1}\b{b}_{\c{S}_e}\|_\infty,
\end{align*}
where we have used the result on \[\text{vec}(\b{X})\] shown in \eqref{vecX}.  

Further, we can write \[\|\b{b}_{\c{S}_e}\|_\infty\] as
\begin{align*}
\|\b{b}_{\c{S}_e}\|_\infty = \|\b{B}_{\c{S}_e}\|_\infty  = \| \lambda_e \text{sign}(\b{A}_0) - \c{P}_{\c{S}_e} (\b{D}^\top\b{UV}^\top)\|_\infty.
\end{align*}
Moving on, we derive an upper bound on \[\|\b{b}_{\c{S}_e}\|_\infty\].
\vspace{-2pt}
\begin{lemma}\label{lbOmegaInf}
	An upper-bound on \[\|\b{b}_{\c{S}_e}\|_\infty\] is given by \[ \|\b{b}_{\c{S}_e}\|_\infty \leq \lambda_e  + \|\c{P}_{\c{S}_e} (\b{D}^\top\b{UV}^\top)\|_\infty\].
\end{lemma}
\vspace{-2pt}
\noindent Then, on defining \[\b{Q} := \b{A}_{{\c{S}_e}^\perp} \b{A}_{\c{S}_e}^\top (\b{A}_{\c{S}_e}\b{A}_{\c{S}_e}^\top)^{-1},\] we have
\begin{align*}
\|\c{P}_{{\c{S}_e}^\perp} (\b{Z}) \|_\infty &= \|\b{Qb}_{\c{S}_e}\|_\infty \leq  \|\b{Q}\|_{\infty, \infty} \|\b{b}_{\c{S}_e}\|_\infty\\
&= \|\b{Q}\|_{\infty, \infty} \| \lambda_e \text{sign}(\b{A}_0) - \c{P}_{\c{S}_e} (\b{D}^\top\b{UV}^\top)\|_\infty,\\
&\leq  \|\b{Q}\|_{\infty, \infty} (\lambda_e  + \|\c{P}_{\c{S}_e} (\b{D}^\top\b{UV}^\top)\|_\infty),
\end{align*}
where we have the following bound for $\|\b{Q}\|_{\infty, \infty}$.
\vspace{-2pt}
\begin{lemma}\label{Q_inf}
An upper-bound on {$\|\b{Q}\|_{\infty, \infty}$} is given by \[\|\b{Q}\|_{\infty, \infty} \leq C_e(\alpha_u, \alpha_\ell,  \gamma_{\b{U}},  \gamma_{\b{V}}, s_e, d, k, \mu)\], where
	\begin{align*}
	 C_e := \tfrac{c}{\alpha_\ell(1 - \mu)^2 - c}
	\end{align*}
	where \[0 \leq C_e < 1\] and \[c\] is defined in \eqref{eqconst:c_entry}.
\end{lemma}  
\vspace{-2pt}
\noindent Combining this with \eqref{eqn:P_DTao} and Lemma~\ref{Q_inf}, we have
\begin{align}
\|\c{P}_{{\c{S}_e}^\perp} (\b{D^\top\Gamma})\|_\infty \nonumber
&\leq C_e \Big(\lambda_e  + \|\c{P}_{\c{S}_e} (\b{D}^\top\b{UV}^\top)\|_\infty\Big)\\
&\hspace{0.3in}+ \|\c{P}_{{\c{S}_e}^\perp}(\b{D^\top UV^\top})\|_\infty. \label{eqn:P_DTao2}
\end{align}

\noindent By simplifying \eqref{eqn:P_DTao2}, we arrive at the lower bound \[\lambda_e^{\min}\] for \[\lambda_e\] as in \eqref{eqn:lam}, from which \textbf{(C4)} holds. Gleaning from the expressions for \[\lambda^{\max}_e\] and \[\lambda^{\min}_e\], we observe that \[\lambda^{\max}_e > \lambda^{\min}_e \geq 0\] for the existence of \[\lambda_e\] that can recover the desired matrices. This completes the proof. \qed

\vspace{0.05in}

\noindent\textbf{Characterizing \[\lambda_e^{\min}\]:} 
In the previous section, we characterized the \[\lambda_e^{\min}\] and \[\lambda_e^{\max}\] based on the dual certificate construction procedure. For the recovery of the true pair \[(\b{L}, \b{S})\], we require $\lambda_e^{\max} >\lambda_e^{\min} \geq 0$. 
Since \[\xi_e \geq 0\] and \[c\geq 0\] by definition, we need \[ 0 \leq C_e<1\] for \[\lambda_e^{\min} >0\], i.e., 
\begin{align}
\label{lam_min_con}
c < \tfrac{1}{2}\alpha_\ell(1 - \mu)^2 \leq \tfrac{\alpha_\ell}{2}.
\end{align}

\vspace{0.05in}

\noindent\textbf{\textit{Conditions for \textit{thin} \[\b{D}\]}}: 
To simplify the analysis we assume, without loss of generality, that \[d < m\]. Specifically, we will assume that \[d \leq \tfrac{m}{\alpha r}\], where \[\alpha >1\] is a constant. With this assumption in mind, we will analyze the following cases for the global sparsity, when \[s_e \leq d\] and \[d < s_e \leq m\].
\\
\textit{Case 1: \[s_e \leq d\]}. 

From \eqref{eqconst:c_entry} and \eqref{lam_min_con}, we have $\alpha_\ell(1 - \mu)^2 - 2c_t > 0$,

which leads to

\begin{align*}
\tfrac{\alpha_u}{\alpha_\ell} < \tfrac{(1 - \mu)^2 + s_e (1  +  \gamma_{\b{V}} )}{s_e(1 + 2 \gamma_{\b{U}} )(1  +  \gamma_{\b{V}} ) +2s_e \gamma_{\b{V}}}.  
\end{align*}
As per the GFP  of \textcolor{blue}{\textbf{D.\ref{frame}}}, we also require that \[{\alpha_u}/{\alpha_\ell} \geq 1\]. Therefore we arrive at
\begin{align*}
\gamma_{\b{U}} < \tfrac{(1 - \mu)^2 - 2s_e \gamma_{\b{V}}}{2s_e( 1 +  \gamma_{\b{V}})} .
\end{align*}
Further, since \[ \gamma_{\b{U}}\geq 0\], we require the numerator to be positive, and since the lower bound on \[ \gamma_{\b{V}} \geq \tfrac{r}{m}\], we have
\begin{align*}
s_e \leq \tfrac{(1 - \mu)^2 }{2}\tfrac{m}{r} := s_e^{\max}, 
\end{align*}
which also implies \[s_e\leq m\]. Now, the condition \[c_t \geq 0\] implies
\begin{align*}
\tfrac{\alpha_u}{\alpha_\ell} \geq \tfrac{1 +  \gamma_{\b{V}} }{(1 + 2 \gamma_{\b{U}} )(1  +  \gamma_{\b{V}} ) +2 \gamma_{\b{V}} }. 
\end{align*}
Since, the R.H.S. of this inequality is upper bounded by \[1\] (achieved when \[ \gamma_{\b{U}}\] and \[ \gamma_{\b{V}}\] are zero). This condition on \[c_t\] is satisfied by our assumption that \[{\alpha_u}/{\alpha_\ell} \geq 1\].

\noindent\textit{Case 2: \[d< s_e \leq m\]}. 

Again, due to the requirement that \[{\alpha_u}/{\alpha_\ell} \geq 1\], following a similar argument as in the previous case we conclude that
\begin{align*}
\gamma_{\b{U}} \leq \tfrac{(1 - \mu)^2 - 2s_e \gamma_{\b{V}}}{2(d + s_e \gamma_{\b{V}})} ~\text{and}~ s_e \leq \tfrac{(1 - \mu)^2}{2}\tfrac{m}{r}.
\end{align*}

\vspace{0.05in}

\noindent\textbf{\textit{Conditions for \textit{fat} \[\b{D}\]}}: 
To simplify the analysis, we suppose that \[k < m\]. Note that in this case, we require that the coefficient matrix \[\b{S}\] has {\[k\]}-sparse columns. Now, \[c = c_f\]. Using similar arguments as above

\begin{align*}
\gamma_{\b{U}} <\tfrac{(1 - \mu)^2 - 2s_e \gamma_{\b{V}}}{2( k + s_e \gamma_{\b{V}})} ~\text{and}~ s_e \leq \tfrac{(1 - \mu)^2 }{2}\tfrac{m}{r}.
\end{align*}

\vspace{0.05in}

\noindent\textbf{Characterizing \[\lambda_e^{\max}\]:} 
Further, the condition \[\lambda_e^{\min}< \lambda_e^{\max}\] translates to a relationship between rank \[r\], and the sparsity \[s_e\], as shown in \eqref{rankSpar1} for \[s_e \leq  s_e^{\max}\].

\vspace*{-5pt}
\subsection{Proof of Theorem~\ref{theorem_col}} \label{sec:pf:thm_col}
In this section we prove Theorem~\ref{theorem_col}; the proofs of lemmata are provided in Appendix~\ref{app:col}. The Lagrangian of the nonsmooth optimization problem \ref{Pc} is
\begin{align}\label{eqn:lagmain}
\c{F}(\b{L}, \b{S}, \b{\Lambda}) = \| \b{L} \|_* + \lambda_c \| \b{S} \|_{1,2} + \langle \b{\Lambda}, \b{M} - \b{L} - \b{D} \b{S} \rangle,
\end{align}
where \[\b{\Lambda} \in \RR^{n \times m}\] is a dual variable. The subdifferentials of \eqref{eqn:lagmain} with respect to \[(\b{L}, \b{S})\] are
\begin{align}
&\hspace{-0.1in}\partial_{\b{L}} \c{F}(\b{L}, \b{S}, \b{\Lambda}) = \resizebox{0.35\textwidth}{0.0153\textwidth}{$ \left\{ \b{U} \b{V}^\top + \b{W} - \b{\Lambda}, \| \b{W} \|_2 \leq 1, \c{P}_{\c{L}} (\b{W}) = \b{0} \right\} $}, \nonumber \\
&\hspace{-0.1in}\partial_{\b{S}} \c{F}(\b{L}, \b{S}, \b{\Lambda}) = \Big\{ \lambda_c \b{H} + \lambda_c \b{F} - \b{D}^\top \b{\Lambda},\c{P}_{{\c{S}_c}}(\b{H}) = \b{H}, \notag \\ 
&\hspace{1.5cm}\c{P}_{{\c{S}_c}}(\b{F}) = \b{0}, \| \b{F} \|_{\infty,2} \leq 1, \b{H}_{:,j} = \tfrac{\b{S}_{:,j}}{\| \b{S}_{:,j} \|_2} \Big\}. \label{eqn:subdifC}
\end{align}
We claim that a pair \[(\b{L}, \b{S})\] is an optimal point of \ref{Pc} if and only if the following hold by the optimality conditions:
\begin{align}
\b{0}_{n \times m} &\in \partial_{\b{L}} \c{F}(\b{L}, \b{S}, \b{\Lambda}) ~\text{and}~ \label{eqn:optL} \\
\b{0}_{d \times m} &\in \partial_{\b{S}} \c{F}(\b{L}, \b{S}, \b{\Lambda}). \label{eqn:optC}
\end{align}

The following lemma states the optimality conditions for the optimal solution pair \[(\b{L}, \b{S})\].
\vspace{-2pt}
\begin{lemma}\label{lem:dualcertify}
Given \[\b{M}\] and \[\b{D}\], let \[(\b{L}, \b{S})\] define the oracle model \[\{ \b{M}, \c{U}, \c{I}_{\c{S}_c} \}\]. Then any solution \[(\b{L}_0, \b{S}_0) \in \{ \b{M}, \c{U}, \c{I}_{\c{S}_c} \}\] is the an optimal solution pair of \ref{Pc}, if there exists a dual certificate \[\b{\Gamma} \in \RR^{n \times m}\] that satisfies\\
$(\b{C1})$ \[\c{P}_{\c{L}} (\b{\Gamma}) = \b{U} \b{V}^\top\], $(\b{C2})$ \[\c{P}_{\c{S}_c} (\b{D}^\top \b{\Gamma})  = \lambda_c \b{H}\], where \[\b{H}_{:,j} = \b{S}_{:,j} / \| \b{S}_{:,j} \|_2\] for all \[j \in \c{I}_{\c{S}_c}\]; \[\b{0}\], otherwise,\\
$(\b{C3})$ \[\| \c{P}_{\c{L}^\perp} (\b{\Gamma}) \|_2 < 1\], and $(\b{C4})$ \[\| \c{P}_{{\c{S}_c}^\perp} (\b{D}^\top \b{\Gamma}) \|_{\infty,2} < \lambda_c\].
\end{lemma}
\vspace{-2pt}

\noindent
We first propose \[\b{\Gamma}\] as the dual certificate, where
\begin{align*}
\b{\Gamma} = \b{U} \b{V}^\top  + \left( \b{I} - \b{P}_{\b{U}} \right) {\b{X}} \left( \b{I} - \b{P}_{\b{V}} \right),~\text{for any}~\b{X} \in \RR^{n \times m}.
\end{align*}
Hence, the condition \textbf{(C1)} is readily satisfied by our choice of \[\b{\Gamma}\]. Now, the condition \textbf{(C2)}, defined as $\c{P}_{\c{S}_c} (\b{D}^\top \b{\Gamma})  = \lambda_c \tilde{\b{S}}$, where \[\tilde{\b{S}}_{:,j} = \tfrac{\b{S}_{:,j}}{\| \b{S}_{:,j} \|_2}\] for all \[j \in \c{I}_{\c{S}_c}\]; \[\b{0}\], otherwise. Substituting the expression for \[\b{\Gamma}\], we need the following condition to hold
\begin{align}
\hspace{-0.08in}\c{P}_{\c{S}_c} (\b{D}^\top \b{U} \b{V}^\top) \hspace{-0.02in}+ \hspace{-0.02in} \c{P}_{\c{S}_c} (\b{D}^\top ( \b{I} - \b{P}_{\b{U}}) {\b{X}} \left( \b{I} - \b{P}_{\b{V}} \right) )  \hspace{-0.02in}= \hspace{-0.02in} \lambda_c \tilde{\b{S}}.  \hspace{-0.02in} \label{eqn:dual1}
\end{align}
Letting \[\b{Z}:= \b{D}^\top \left( \b{I} - \b{P}_{\b{U}} \right) {\b{X}} \left( \b{I} - \b{P}_{\b{V}} \right)\] and \[\b{B}_{\c{S}_c} := \lambda_c \tilde{\b{S}} - \c{P}_{\c{S}_c} (\b{D}^\top \b{U} \b{V}^\top) \], we have $\c{P}_{\c{S}_c} \left(\b{Z} \right) = \b{B}_{\c{S}_c}$. Further, vectorizing the equation above, we have 
\begin{align}
\label{b_sc}
\c{P}_{\c{S}_c} \left(\vect(\b{Z}) \right) = \b{b}_{\c{S}_c},
\end{align}
where \[\b{b}_{\c{S}_c} := \vect(\b{B}_{\c{S}_c})\]. Next, by letting \[\b{A} := \b{(I - P_V)}\otimes \b{D^\top}(\b{I- P_U})\], using the definition of \[\b{Z}\] and the properties of the Kronecker product we have $\text{vec}(\b{Z}) = \b{A} \text{vec}(\b{X})$. Now, let \[\b{A}_{\c{S}_c}\] denote the rows of \[\b{A}\] corresponding to the non-zero rows of \[\vect(\b{S})\] and \[\b{A}_{\c{S}_c^\perp}\] denote the remaining rows, then
\begin{align}\label{Asc_vecX}
\c{P}_{\c{S}_c} \left(\vect(\b{Z}) \right) = \b{A}_{\c{S}_c}\text{vec}(\b{X}).
\end{align}
From \eqref{b_sc} and \eqref{Asc_vecX}, we have $\b{A}_{\c{S}_c}\vect(\b{X}) = \b{b}_{\c{S}_c}$. Therefore, we need the following
\begin{align}
\label{vecX_sc}
\text{vec}(\b{X}) = \b{A}_{\c{S}_c}^\top(\b{A}_{\c{S}_c}\b{A}_{\c{S}_c}^\top)^{-1}\b{b}_{\c{S}_c},
\end{align}
which corresponds to the least norm solution i.e., \[\b{X} = \text{argmin}_{\b{X}} ~\|{\b{X}}\|_{\rm F}\], s.t. \[\b{A}_{\c{S}_c} \vect({\b{X}}) = \b{b}_{\c{S}_c}\]. For this choice of \[\b{X}\] \eqref{eqn:dual1} is satisfied and consequently so is the condition \textbf{(C2)}. Here, the existence of the inverse is ensured by the following.
\begin{lemma}\label{lower_sigmaMin_col}
If \[\mu<1\] and \[\alpha_\ell >0\], the minimum singular value of \[\b{A}_{\c{S}_c}\] is bounded away from \[0\] and is given by \[\sqrt{\alpha_{\ell}} (1 - \mu)\]
\end{lemma}
\begin{figure}[!]
	\centering
	{\small
		\begin{tabular}{cc}
			\hspace{-0.4cm} {\small\rotatebox{90}{ ~~~~~~Recovery of $\b{L}$}}
			\includegraphics[width=0.21\textwidth]{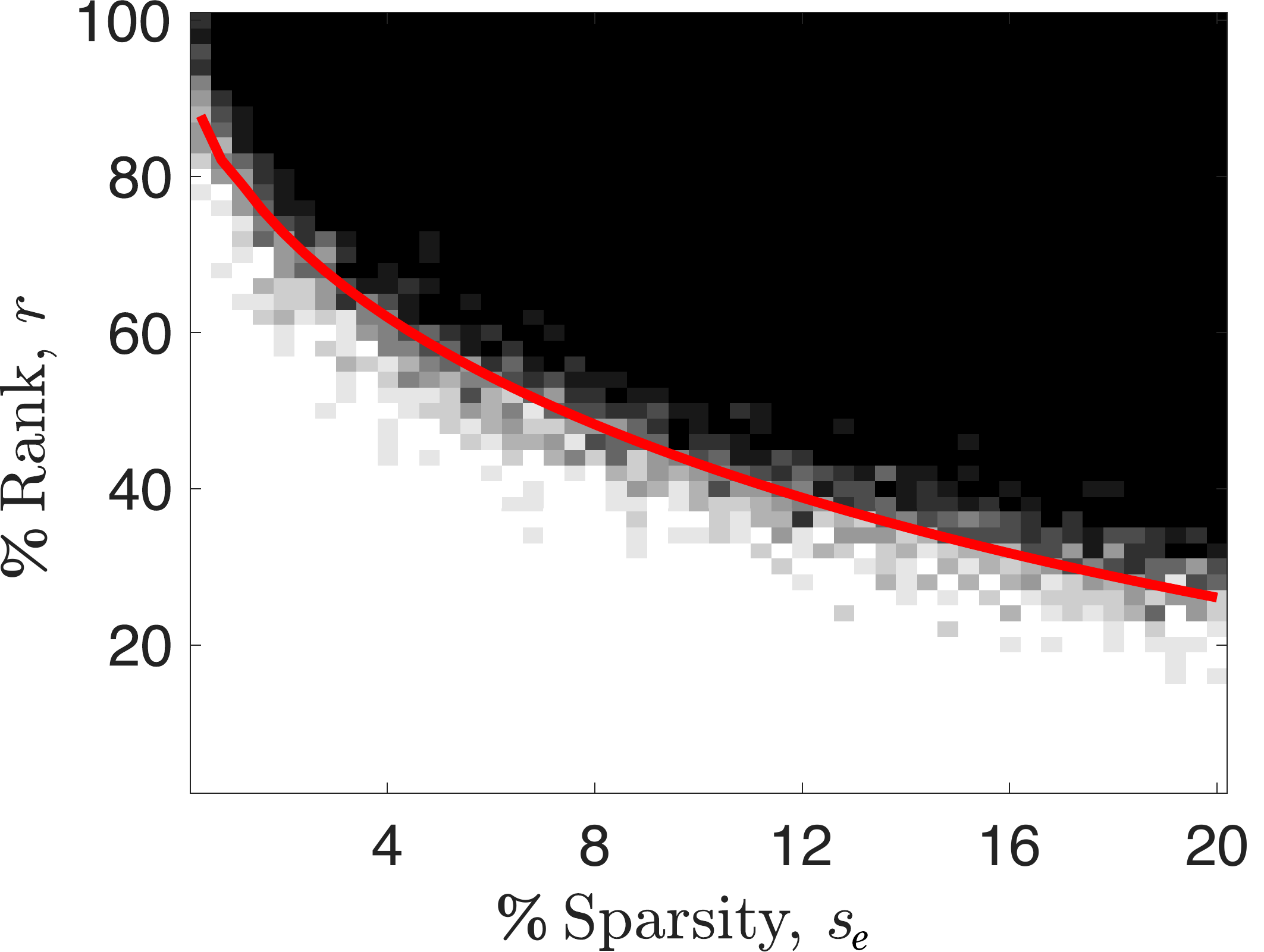} & \hspace{-0.2cm} \includegraphics[width=0.21\textwidth]{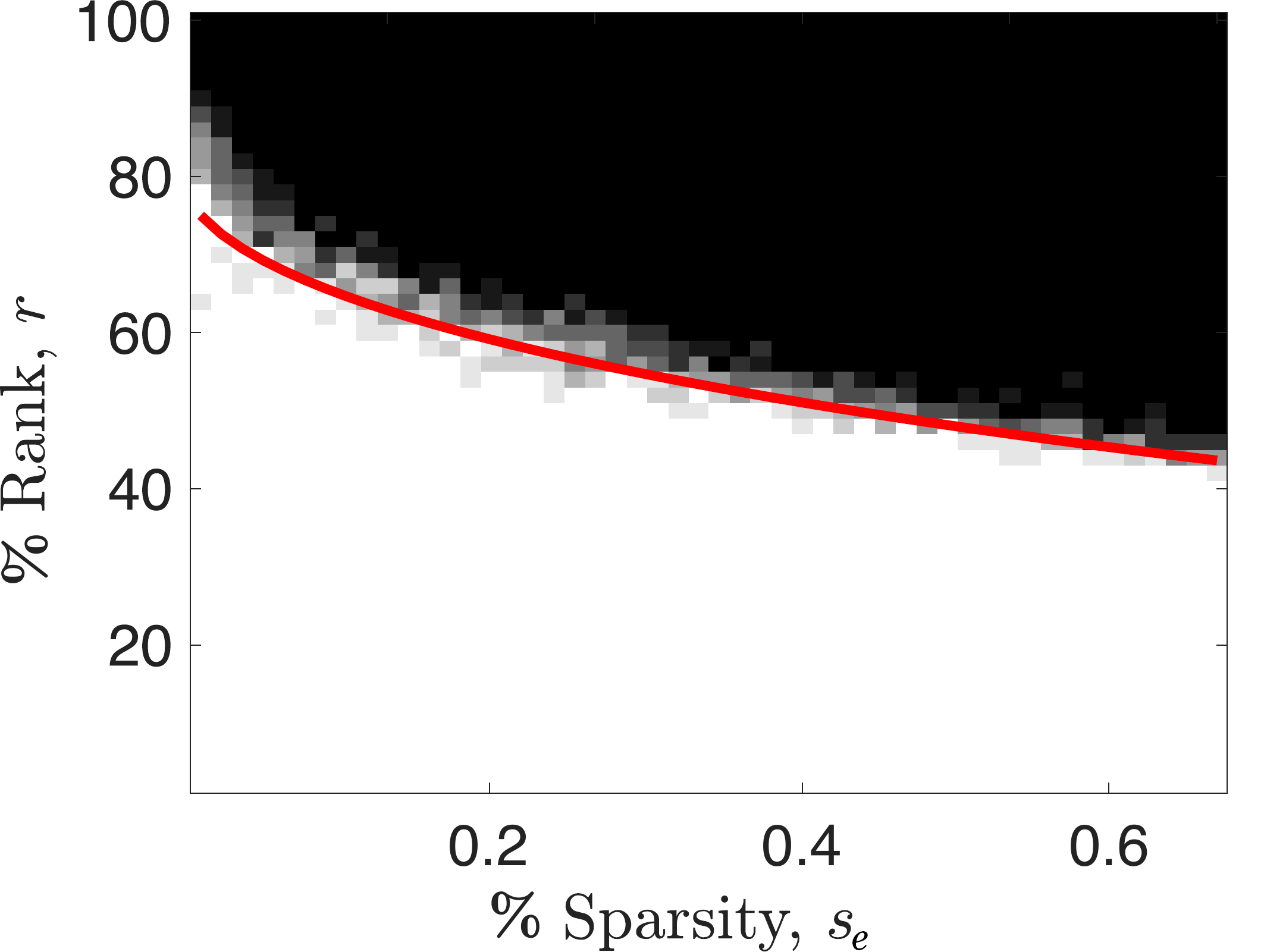} \vspace{-2pt}\\
			~(a) $d = 5$ & ~(b) $d = 150$ \vspace{0.02in} \\
			\hspace{-0.2cm}{\small\rotatebox{90}{ ~~~~~~Recovery of $\b{S}$}}
			\includegraphics[width=0.21\textwidth]{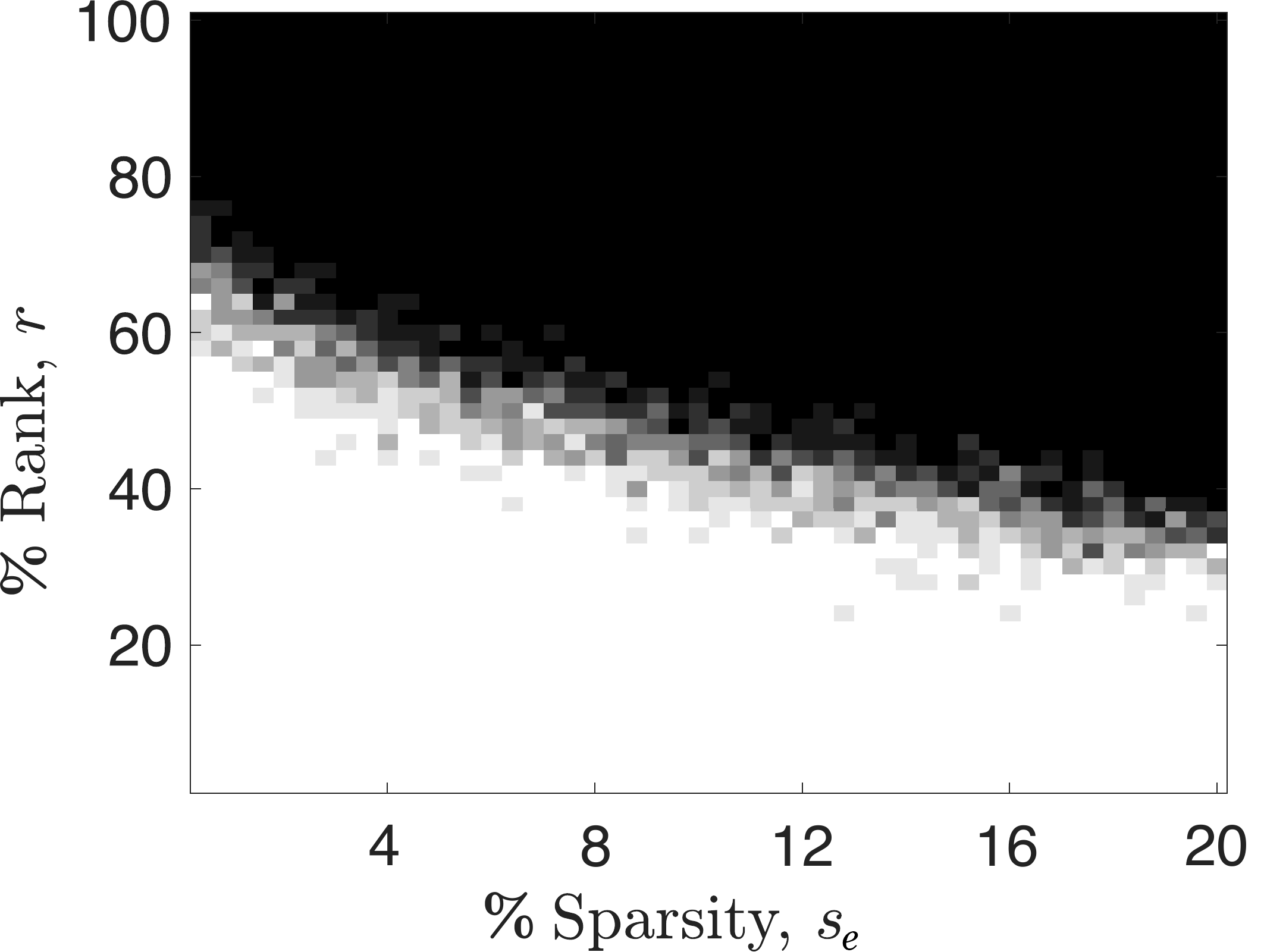} 
			& \hspace{-0.2cm} \includegraphics[width=0.21\textwidth]{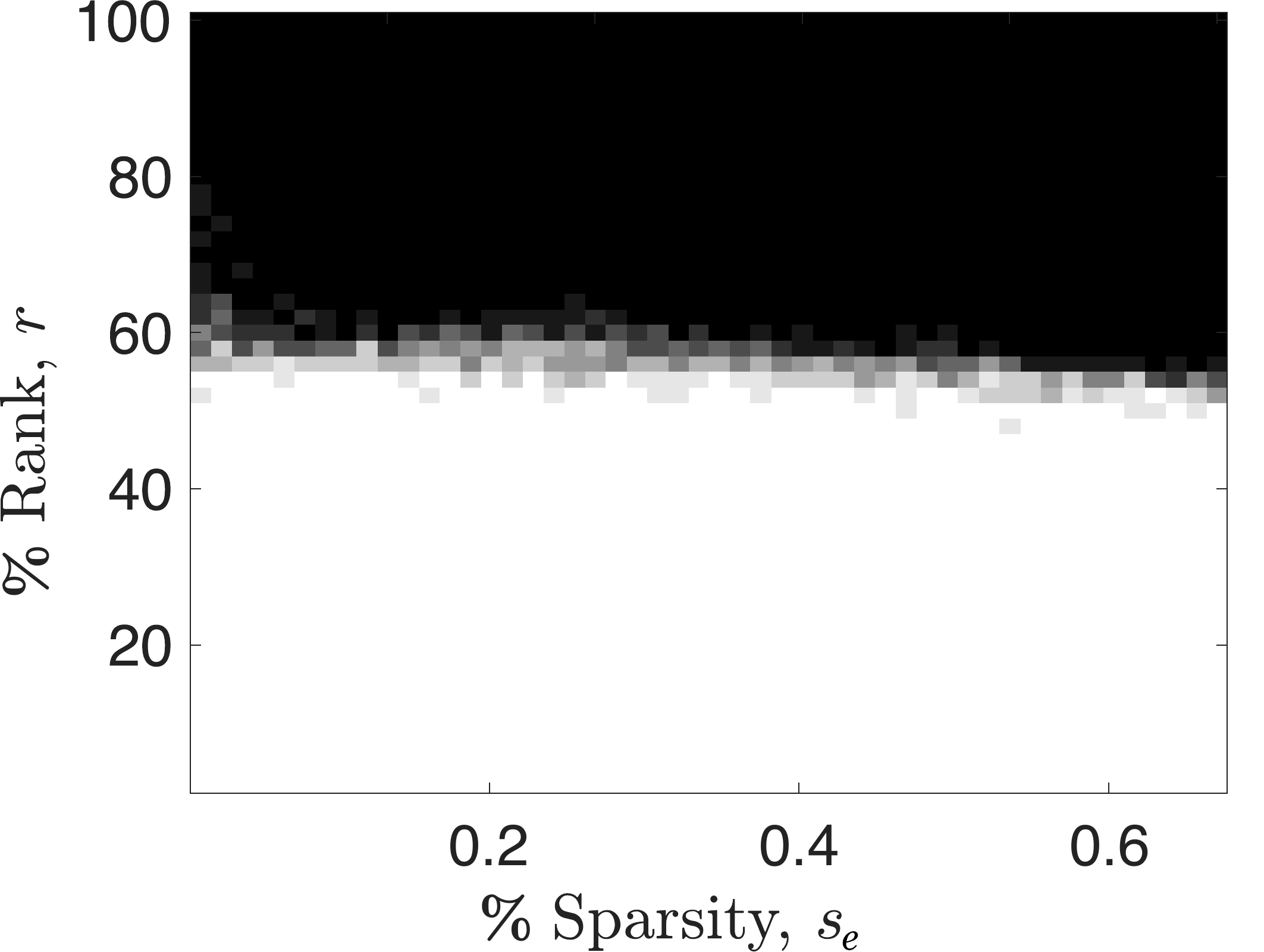} \vspace{-2pt}\\
			~(c) $d = 5$ & ~(d) $d = 150$ \\
			\vspace{-0.5cm}
	\end{tabular}}
	\caption{\footnotesize Recovery for varying rank of {\footnotesize$\b{L}$}, sparsity of {\footnotesize $\b{S}$} and number of dictionary elements in {\footnotesize $\b{D}$} as per Theorem~\ref{theorem_entry}. Each plot shows average recovery across $10$ trials for varying ranks and sparsity up to {\footnotesize $s_e^{\max} = m$}, where $n = m = 100$ and the white region represents correct recovery. We declare success if {\footnotesize $\| \b{L} - \hat{\b{L}}\|_{\rm F}/\| \b{L} \|_{\rm F} \leq 0.02$} and {\footnotesize $\| \b{S} - \hat{\b{S}}\|_{\rm F}/\| \b{S} \|_{\rm F} \leq 0.02$}, where {\footnotesize $\hat{\b{L}}$} and {\footnotesize $\hat{\b{S}}$} are the recovered {\footnotesize $\b{L}$} and {\footnotesize $\b{S}$}, respectively. Panels (a)-(b) show the recovery of the low-rank part {\footnotesize $\b{L}$} along with the predicted trend between rank {\small $r$} and sparsity {\small $s_e$} as per eq.\eqref{rankSpar1} (in red), and (c)-(d) show the recovery of the sparse part with varying dictionary sizes $d = 5 ~\text{and}~ 150$, respectively.}
	\vspace{-0.6cm}
	\label{fig:phaseTr}
\end{figure} 
\noindent
Upon the existence of such \[\b{X}\] as defined in \eqref{vecX_sc}, \textbf{(C3)} is satisfied if the following condition holds
\begin{align*}
&\| \c{P}_{\c{L}^\perp} (\b{\Gamma}) \|_2 = \| \left( \b{I} - \b{P}_{\b{V}} \right) \b{X} \left( \b{I} - \b{P}_{\b{U}} \right) \|_2 \\
&\leq \| \b{I} - \b{P}_{\b{V}} \|_2 \| \b{X} \|_2 \| \b{I} - \b{P}_{\b{U}} \|_2= \| \b{X} \|_2 \leq \| \b{X} \|_{\rm F} < 1.
\end{align*}
From \eqref{vecX_sc}, this condition translates to 
\begin{align*}
\|\b{A}_{\c{S}_c}^\top(\b{A}_{\c{S}_c}\b{A}_{\c{S}_c}^\top)^{-1}\| \| \b{b}_{\c{S}_c}\|_2 < 1.
\end{align*}
Now, since \[\|\b{A}_{\c{S}_c}^\top(\b{A}_{\c{S}_c}\b{A}_{\c{S}_c}^\top)^{-1}\| = 1/\sigma_{\min}(\b{A}_{\c{S}_c})\] (see the analogous analysis for the entry-wise case), we need
\begin{align*}
\| \c{P}_{\c{L}^\perp} (\b{\Gamma}) \|_2 \leq \tfrac{\| \b{b}_{\c{S}_c}\|_2}{\sigma_{\min}(\b{A}_{\c{S}_c})} < 1.
\end{align*}
Now, using Lemma~\ref{lower_sigmaMin_col}  and the following bound on \[\| \b{b}_{\c{S}_c} \|_2\],
\begin{lemma}\label{upper_bOmega_col}
An upper-bound on \[\| \b{b}_{\c{S}_c} \|_2\] is given by \[\lambda_c \sqrt{s_c} + \sqrt{r \alpha_{u}}\mu\].
\end{lemma}
\vspace{-4pt}
\noindent we have that the condition \textbf{(C3)} holds if
\begin{align*}
\| \c{P}_{\c{L}^\perp} (\b{\Gamma}) \|_2 \leq \tfrac{\lambda_c \sqrt{s_c} + \sqrt{r \alpha_{u}}\mu}{\sqrt{\alpha_{\ell}} (1 - \mu)} <1,
\end{align*}
which is satisfied by our choice of \[\lambda_c^{\max}\] \eqref{eqn:lamb_c}. Now, for the condition \textbf{(C4)} we need the following condition to hold true:
\begin{align*}
&\| \c{P}_{\c{S}_c^\perp} (\b{D}^\top \b{\Gamma}) \|_{\infty,2} \\
&\leq \| \c{P}_{\c{S}_c^\perp}\hspace{-0.02in}(\b{D}^\top \hspace{-0.02in}\b{U} \b{V}^\top\hspace{-0.02in}) \|_{\infty,2} \hspace{-0.02in}+\hspace{-0.02in} \| \c{P}_{\c{S}_c^\perp}(\b{D}^\top \hspace{-0.02in}( \b{I} \hspace{-0.01in}-\hspace{-0.01in} \b{P}_{\b{U}} \hspace{-0.02in}) \hspace{-0.01in}\b{X} \hspace{-0.01in}( \b{I} \hspace{-0.01in}-\hspace{-0.01in} \b{P}_{\b{V}} \hspace{-0.02in})) \|_{\infty,2} \\
&= \| \c{P}_{\c{S}_c^\perp} (\b{D}^\top \b{U} \b{V}^\top) \|_{\infty,2} + \|\c{P}_{\c{S}_c^\perp}(\b{Z}) \|_{\infty,2} < \lambda_c.
\end{align*}
Note that, here \[\|\c{P}_{\c{S}_c}(\b{D}^T \b{U} \b{V}^T)\|_{\infty,2} \leq \xi_c\]. Further, the following result establishes an upper-bound on \[ \|\c{P}_{\c{S}_c^\perp}(\b{Z}) \|_{\infty,2}\].
\vspace{-11pt}
\begin{lemma}\label{lem:step3}
An upper bound on \[ \|\c{P}_{\c{S}_c^\perp}(\b{Z}) \|_{\infty,2}\] is given by
\[( \lambda_c s_c + \sqrt{r \alpha_{u}s_c}\mu)C_c .\]
\end{lemma}
\vspace{-2pt}
\noindent In light of this, the condition \textbf{(C4)} implies that,
\begin{align*}
\xi_c + \tfrac{ \alpha_u}{\alpha_\ell(1 - \mu)^2}\sqrt{s_c} \gamma_{\b{V}}\beta_{\b{U}} ( \lambda_c \sqrt{s_c} + \sqrt{r \alpha_{u}}\mu) <\lambda_c.
\end{align*}
\noindent To this end, if we let \[C_c := \tfrac{ \alpha_u}{\alpha_\ell(1 - \mu)^2} \gamma_{\b{V}}\beta_{\b{U}}\], \textbf{(C4)} is satisfied by $\lambda_c^{\min}$ defined in \eqref{eqn:lamb_c}. This completes the proof. \qed

\begin{figure}[t]
	\centering
	{\small
		\begin{tabular}{cc}
			\hspace{-0.2cm} {\small\rotatebox{90}{ ~~~~~~Recovery of $\b{L}$}}
			\includegraphics[width=0.21\textwidth]{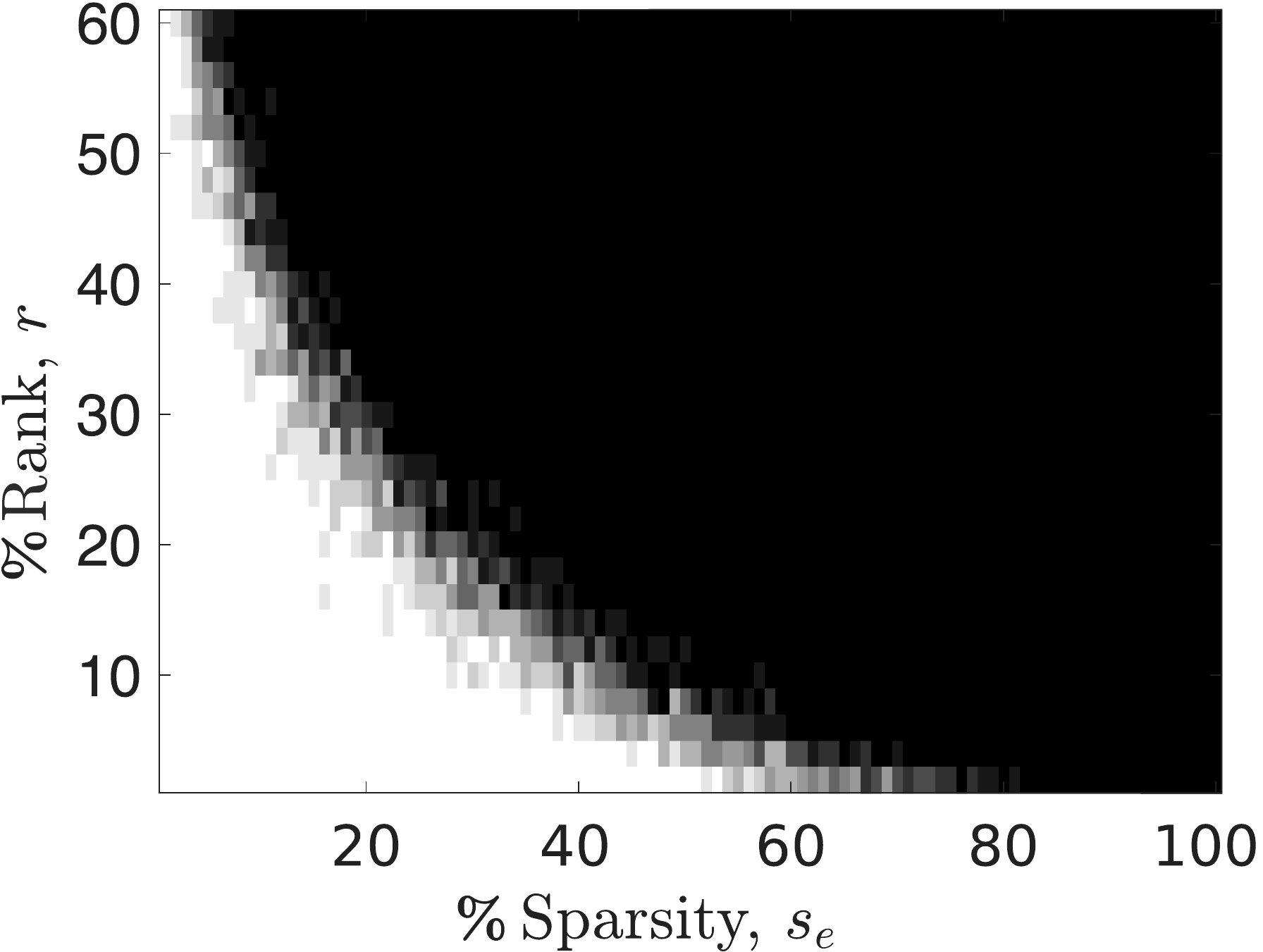} &\hspace{-0.2cm} \includegraphics[width=0.205\textwidth]{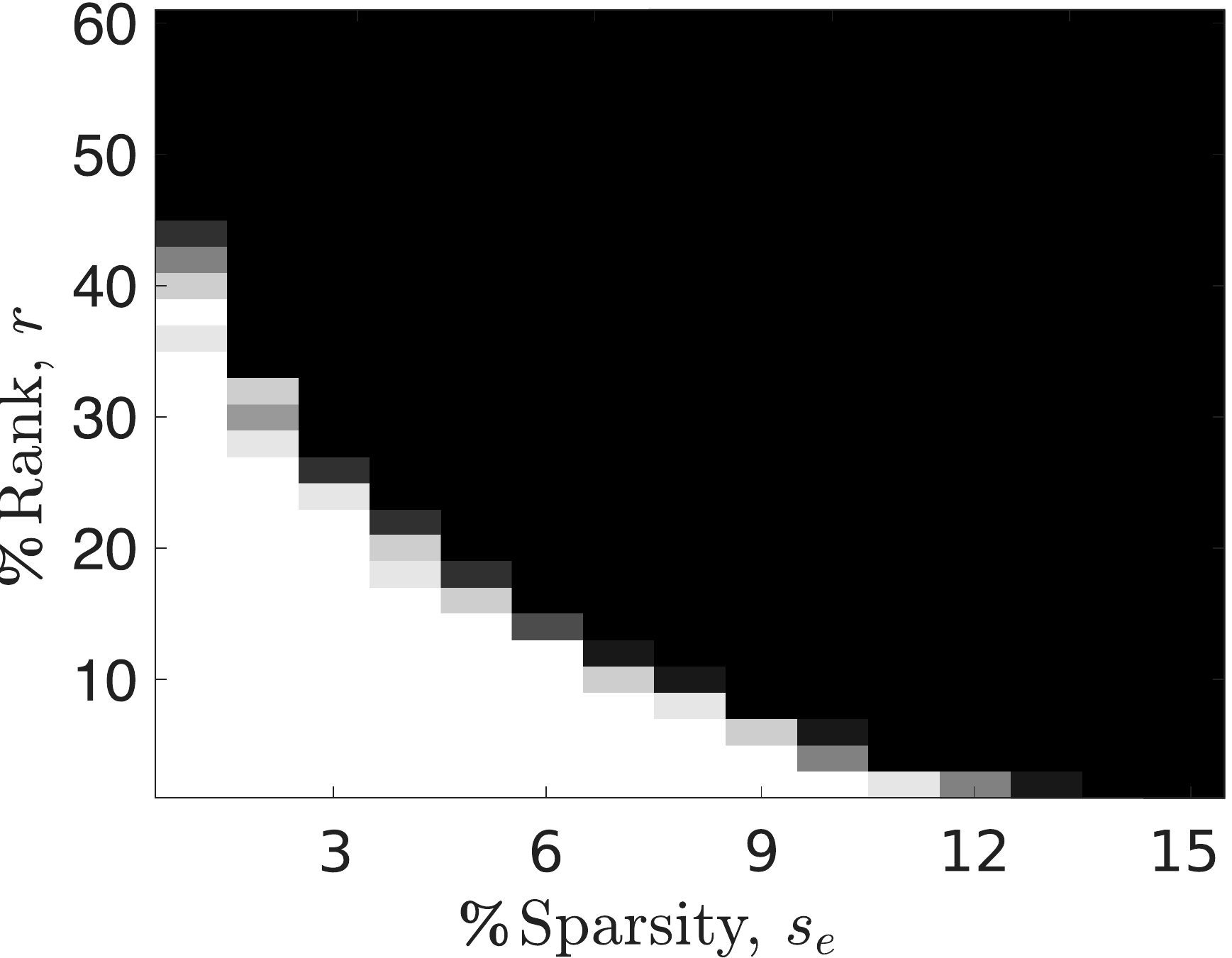} \vspace{-2pt}\\
			~(a) $d = 5$ & ~(b) $d = 150$ \vspace{0.02in} \\
			\hspace{-0.2cm}{\small\rotatebox{90}{ ~~~~~~Recovery of $\b{S}$}}
			\includegraphics[width=0.21\textwidth]{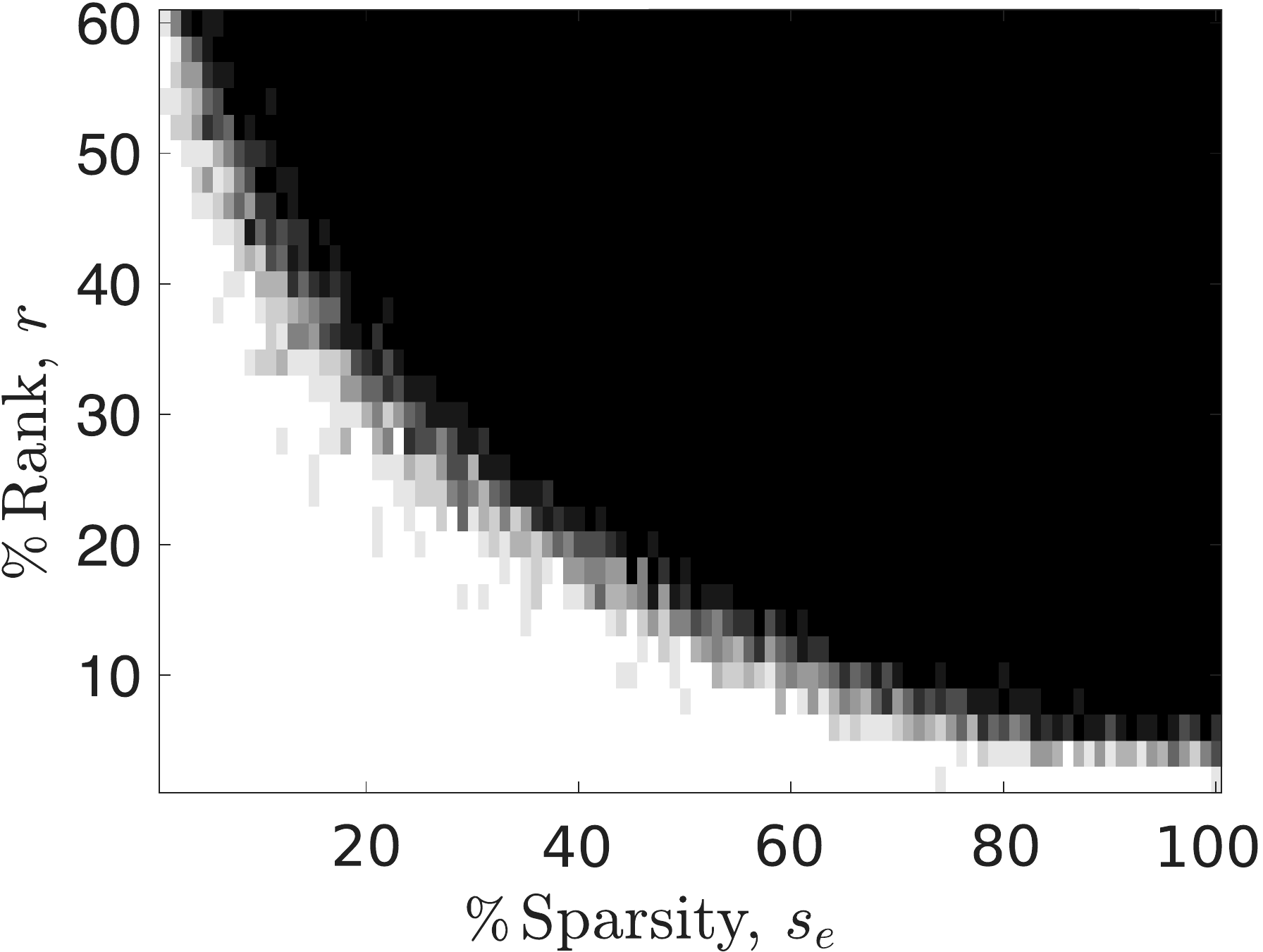} & \hspace{-0.2cm} \includegraphics[width=0.205\textwidth]{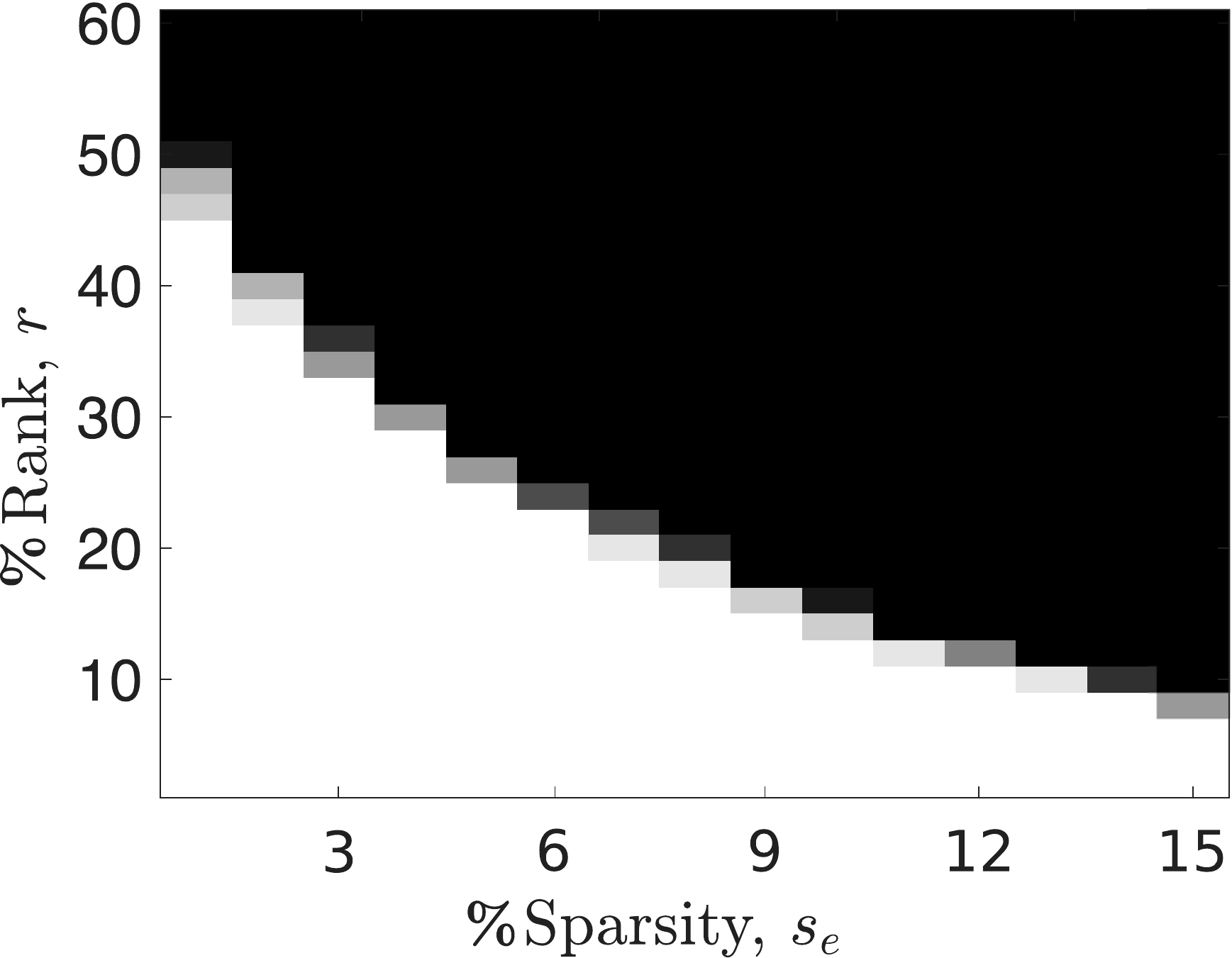} \vspace{-2pt}\\
			~(c) $d = 5$ & ~(d) $d = 150$
	\end{tabular}}
	\vspace{-0.1cm}
	\caption{\footnotesize Recovery for varying rank of {\footnotesize$\b{L}$}, sparsity of {\footnotesize $\b{S}$} and number of dictionary elements in {\footnotesize $\b{R}$}. Panels (a)-(b) show the recovery of the low-rank part {\footnotesize $\b{L}$} and (c)-(d) show the recovery of the sparse part with varying dictionary sizes $d = 5 ~\text{and}~ 150$, respectively. The experimental set-up and the success metric remains the same as in Fig.~\ref{fig:phaseTr}.}
	\vspace{-0.6cm}
	\label{fig:phaseTr_full}
\end{figure}
\vspace{0.05in}

\noindent\textbf{Characterizing \[\lambda_c^{\min}\]:} From \eqref{eqn:lamb_c}, we need $\lambda_c^{\min} := \tfrac{\xi_c + \sqrt{r s_c \alpha_u}\mu C_c}{1 - s_cC_c} \geq 0$, where \[C_c := \tfrac{ \alpha_u}{\alpha_\ell(1 - \mu)^2} \gamma_{\b{V}}\beta_{\b{U}} \geq 0\]. Then from \[s_cC_c < 1\], we require $s_c < s_c^{\max} := \tfrac{\alpha_\ell(1 - \mu)^2}{ \alpha_u \gamma_{\b{V}}\beta_{\b{U}}}$.

\vspace{0.05in}

\noindent\textbf{Characterizing \[\lambda_c^{\max}\]:} Since we need \[\lambda_c^{\min} < \lambda_c^{\max}\], substituting the expressions for \[\lambda_c^{\min}\] and \[\lambda_c^{\max}\], and using the fact that \[s_cC_c < 1\], we arrive at  \eqref{rankSparCol}. 

\vspace{-2pt}
\section{Numerical Simulations on Synthetic Data}
\label{sec:simulations}
In this section, we empirically evaluate the properties of \ref{Pe} and \ref{Pc} via phase transition in rank and sparsity, and compare its performance to related techniques, and to the behavior predicted by Theorem~\ref{theorem_entry} and Theorem~\ref{theorem_col} in \eqref{rankSpar1} and \eqref{rankSparCol}, respectively. 

\vspace{-8pt}
\subsection{Entry-Wise Sparsity Case}\label{sec:exp_entry}
\noindent\textbf{Experimental Set-up:}  We employ the accelerated proximal gradient (APG) algorithm outlined in Algorithm~\ref{algo} to solve the optimization problem \ref{Pe}. For these evaluations, we fix $n = m = 100$, and generate the low-rank part $\b{L}$ by outer product of two column normalized random matrices of sizes $n \times r$ and $m \times r$, with entries drawn from the standard normal distribution. In addition, we choose $s_e$ non-zero locations of the sparse component $\b{S}$ randomly, and draw the values at these non-zero entries from the Rademacher distribution, and the dictionary $\b{D}$ from the standard normal distribution with normalized columns. We then run $10$ Monte-Carlo trials for each pair of rank and sparsity, and for each of these, we scan across $100$ values of $\lambda_e$s in the range of  $[\lambda_e^{\min}, \lambda_e^{\max}]$ to find the best pair of \[(\b{L}_0, \b{S}_0)\] to compile the results. For ease of computation we run on modest values of $n$ and $m$. Here, the white and dark region correspond to correct recovery and failure, respectively.

\noindent\textbf{Discussion:}
Phase transition in rank and sparsity averaged over {$10$} trials for dictionaries of sizes {$d = 5$ } (thin) and {$d = 150$} (fat), are shown in  Fig.~\ref{fig:phaseTr} and Fig.~\ref{fig:phaseTr_full}, respectively. We note from Fig.~\ref{fig:phaseTr} that indeed the empirical relationship between rank and sparsity for the recovery of \[(\b{L_0, S_0})\] has the same trend as predicted by \eqref{rankSpar1} in Section~\ref{sec:main_result} for $s_e \leq s_e^{\max}$.  Here, the parameters corresponding to the predicted trend (shown in red) have been hand-tuned for best fit. In fact, as shown in Fig.~\ref{fig:phaseTr_full}, this trend continues for sparsity levels much greater than $s_e^{\max}$. This can be potentially attributed to the worst case deterministic analysis considered here. 
{\footnotesize
\begin{figure}[t]
	\centering
	{\small
		\begin{tabular}{cc}
			\hspace{-0.2cm}{\small\rotatebox{90}{ ~~~~~~Recovery of $\b{S}$}}
			\includegraphics[width=0.21\textwidth]{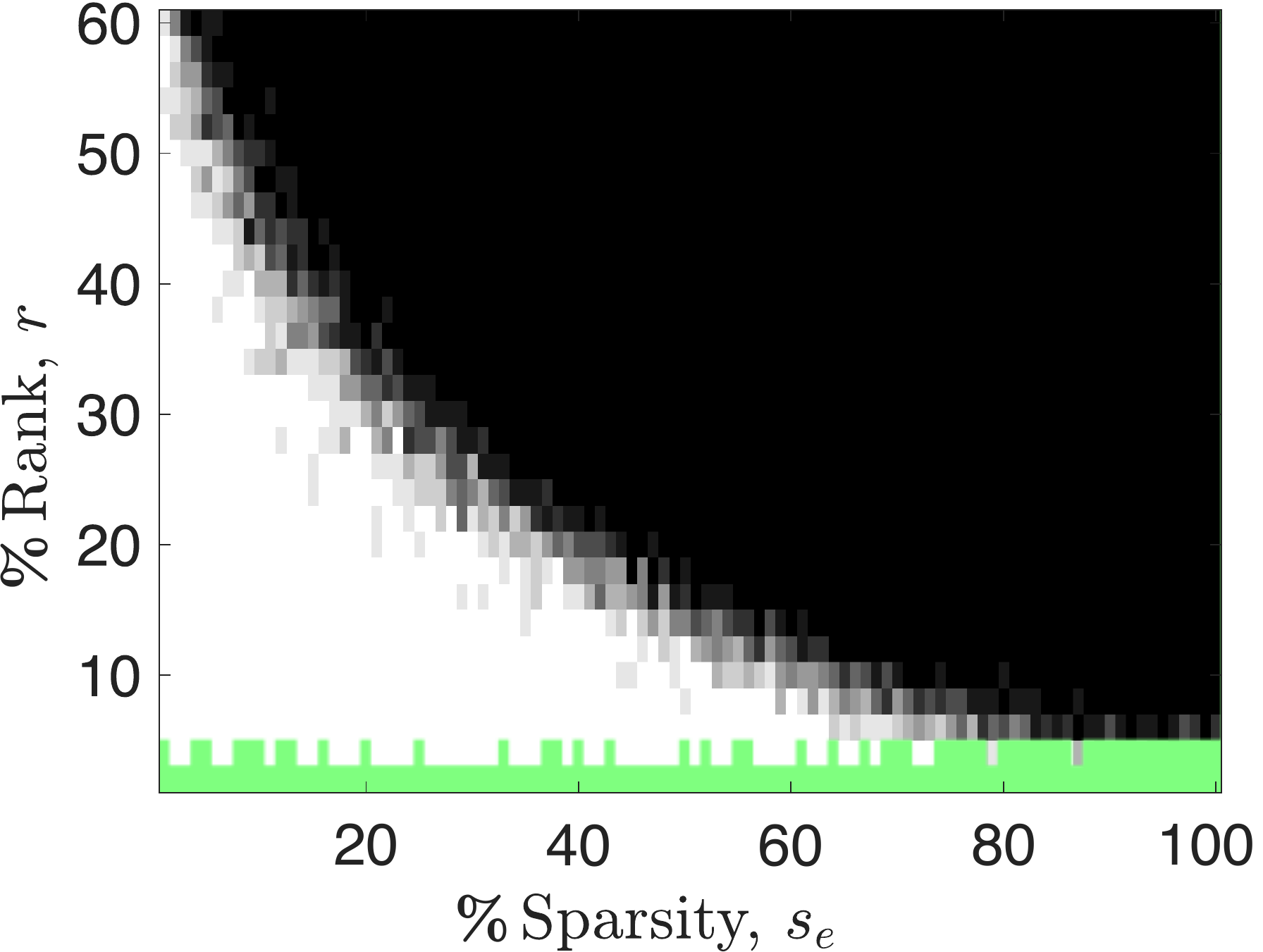} & \hspace{-0.2cm} 
			\includegraphics[width=0.21\textwidth]{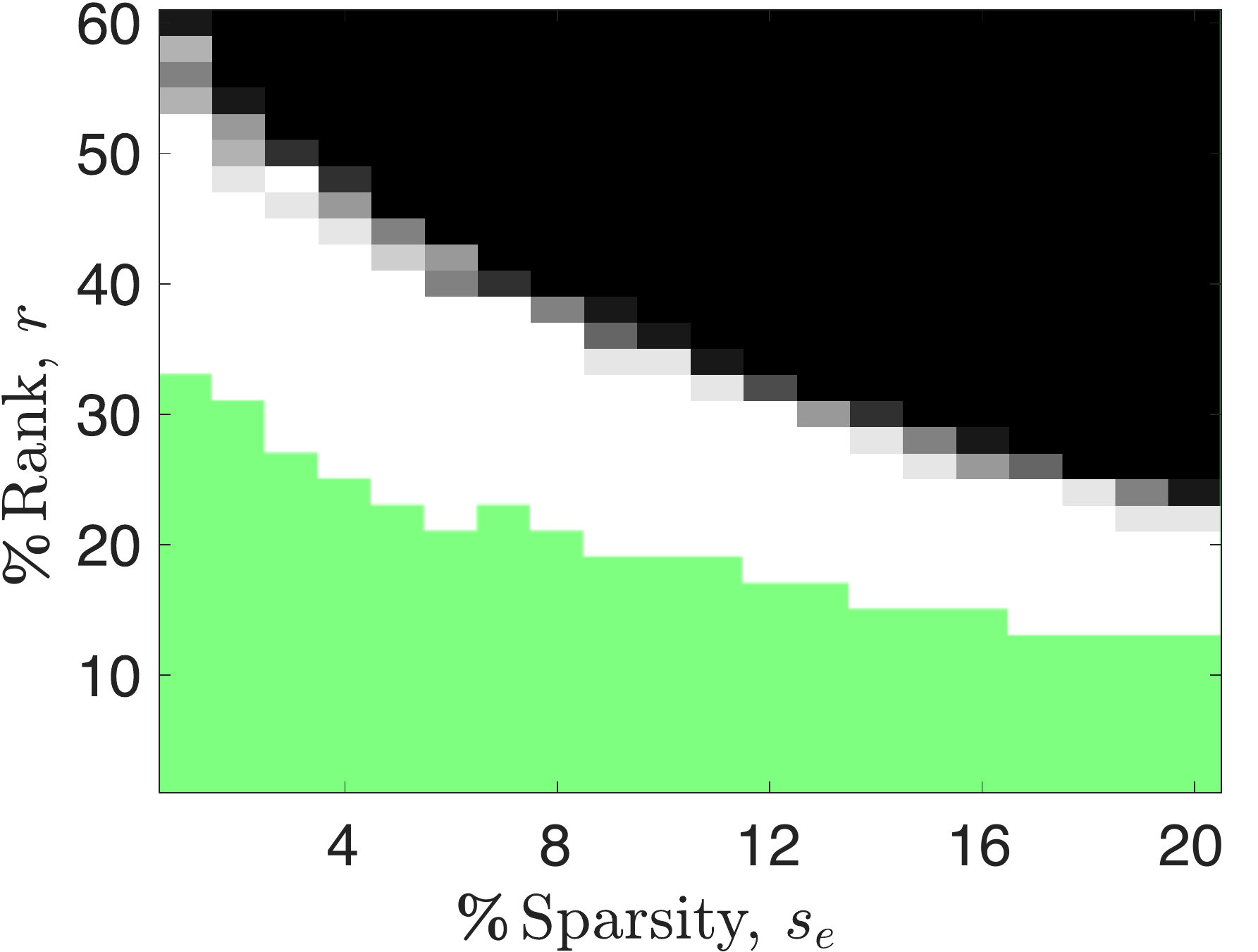} \\
			~(a) $d=5$ & ~(b) $d=50$
	\end{tabular}}
	\vspace{-6pt}
	\caption{\footnotesize Comparision of phase transitions in rank and sparsity between \ref{Pe} and \ref{RPCA} for recovery of $\b{S}$ for different dictionary sizes. Panels (a) and (b) correspond to $d=5$ and $d=50$, respectively. Experimental set-up remains same as Fig.~\ref{fig:phaseTr_full}. The area in green corresponds to recovery by \ref{RPCA} where at least $1$ out of the $10$ Monte-Carlo trials succeeds.}
	\vspace{-20pt}
	\label{fig:phaseTr_comp}
\end{figure}}

Further, Fig.~\ref{fig:phaseTr_comp} shows the results of \ref{RPCA} (in green, shows the area where at least one of the $10$ Monte-Carlo simulations succeeds) in comparision to the results obtained by \ref{Pe} for $d=5$ and $d=50$. We observe that \ref{Pe} outperforms \ref{RPCA} across the board.  In fact, we notice that the \ref{RPCA} technique only succeeds when $r < d$. We believe that this is because when $d < r$ the component $\b{D}^\dagger\b{L}$ is not low-rank (full-rank in this case) w.r.t. the maximum potential rank of $\b{D}^\dagger\b{L}$. As a result, the model assumptions of the robust PCA problem do not apply; see Section~\ref{sec:contribution}. In contrast, the proposed framework of \ref{Pe} can handle these cases effectively (see Fig.~\ref{fig:phaseTr_comp}) since $\b{L}$ is low-rank irrespective of the dictionary size. This highlights the applicability of the our approach to cases where $d<r$, and simultaneous recovery of the low-rank component in one-shot.

\vspace*{-8pt}
\subsection{Column-wise Sparsity Case}
We now present phase transition in rank $r$ and number of outliers$s_c$ to evaluate the performance of \ref{Pc}. In particular, we compare with Outlier Pursuit (OP) \cite{Xu2010} that solves \ref{Pc} with $\b{D} = \b{I}$, and \ref{OP} to demonstrate that the \textit{a priori} knowledge of the dictionary provides superior recovery properties. 
\vspace{6pt}
\noindent\textbf{Experimental Set-up:}

Again, we employ a variant of the APG algorithm outlined in Algorithm~\ref{algo} to solve the optimization problem \ref{Pc}. We set $n = 100$, $m = 1000$, and for each pair of $r$ and $s_c$ we run $10$ Monte-Carlo trials for $r \in \{5, 10, 15 \dots, 100\}$ and $s_c \in \{50, 100, 150, \dots, 900\}$. For our experiments, we form $\b{L} = [\b{U} \b{V}^\top ~|~\b{0}_{n\times s_c}] \in \RR^{n \times m}$, where $\b{U} \in \RR^{n \times r}$, $\b{V} \in \RR^{(m-s_c) \times r}$ have i.i.d. $\c{N}(0,1)$ entries, which are then normalized column-wise. Next, we generate $\b{S} = [\b{0}_{d \times (m - s_c)} ~|~ \b{W}] \in \RR^{d \times m}$ where each entry of $\b{W} \in \RR^{d \times s_c}$ is i.i.d. $\c{N}(0,1)$. Also, the known dictionary $\b{D} \in \RR^{n \times d}$ is formed by normalizing the columns of a random matrix with  i.i.d. $\c{N}(0,1)$ entries. For each method, we scan through $100$ values of the regularization parameter \[\lambda_c \in [\lambda_c^{\min}, \lambda_c^{\max}]\] to find a solution pair \[(\b{L}_0, \b{S}_0)\] with the best precision, i.e. $\rm(True ~Positives/(True ~Positives + False ~Positives))$. We declare an experiment successful if it acheives a precision of $0.99$ or higher. Here, we threshold the column norms at $2\times10^{-3}$ before we evaluate the precision. 

\begin{figure}[t]
	\centering
	\begin{tabular}{ccc}
		$d=50$ & $d=50$ & $d=150$\\
		\hspace{-5pt}\includegraphics[width=0.32\linewidth]{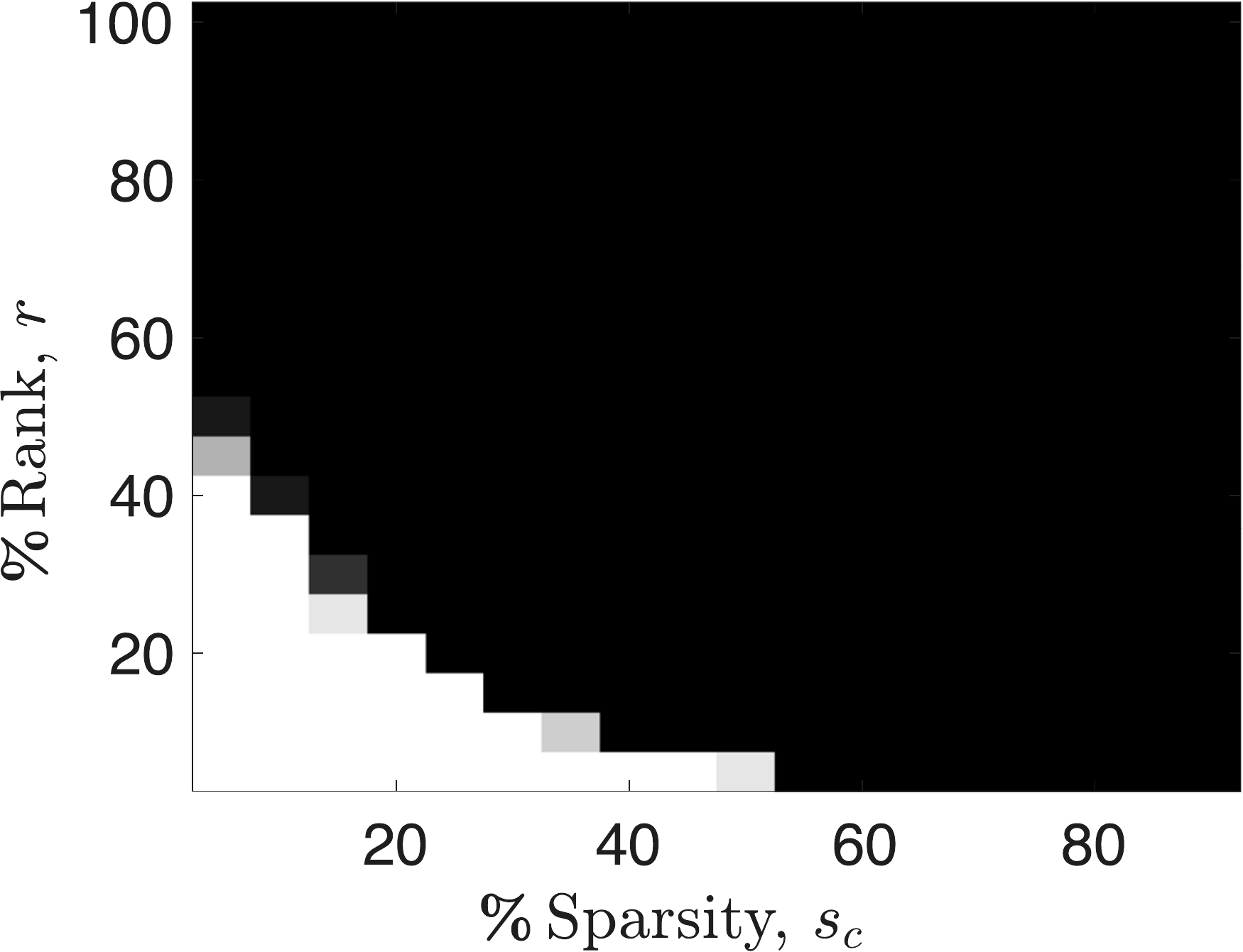} & \hspace{-10pt}
		\includegraphics[width=0.32\linewidth]{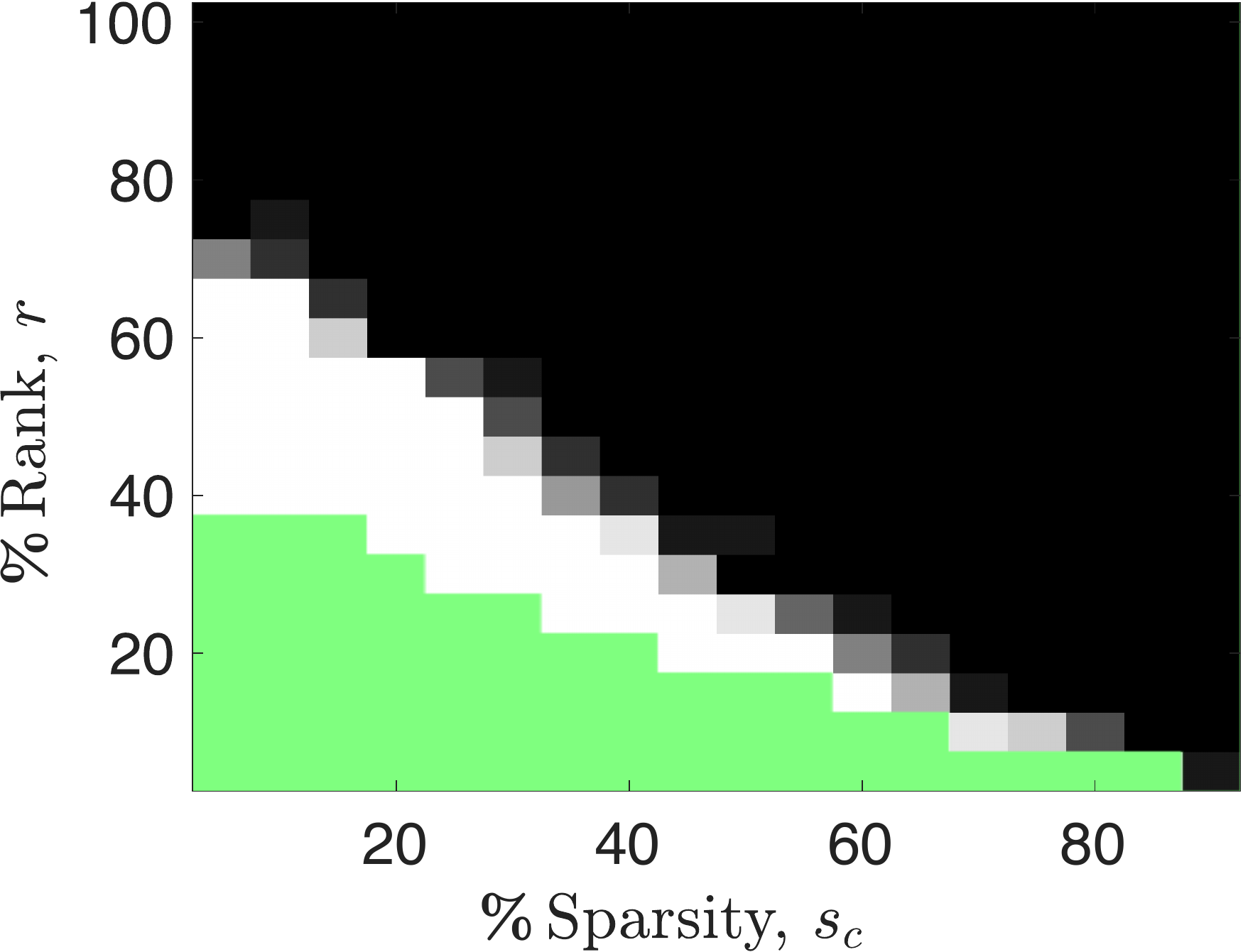} & \hspace{-15pt}
		\includegraphics[width=0.32\linewidth]{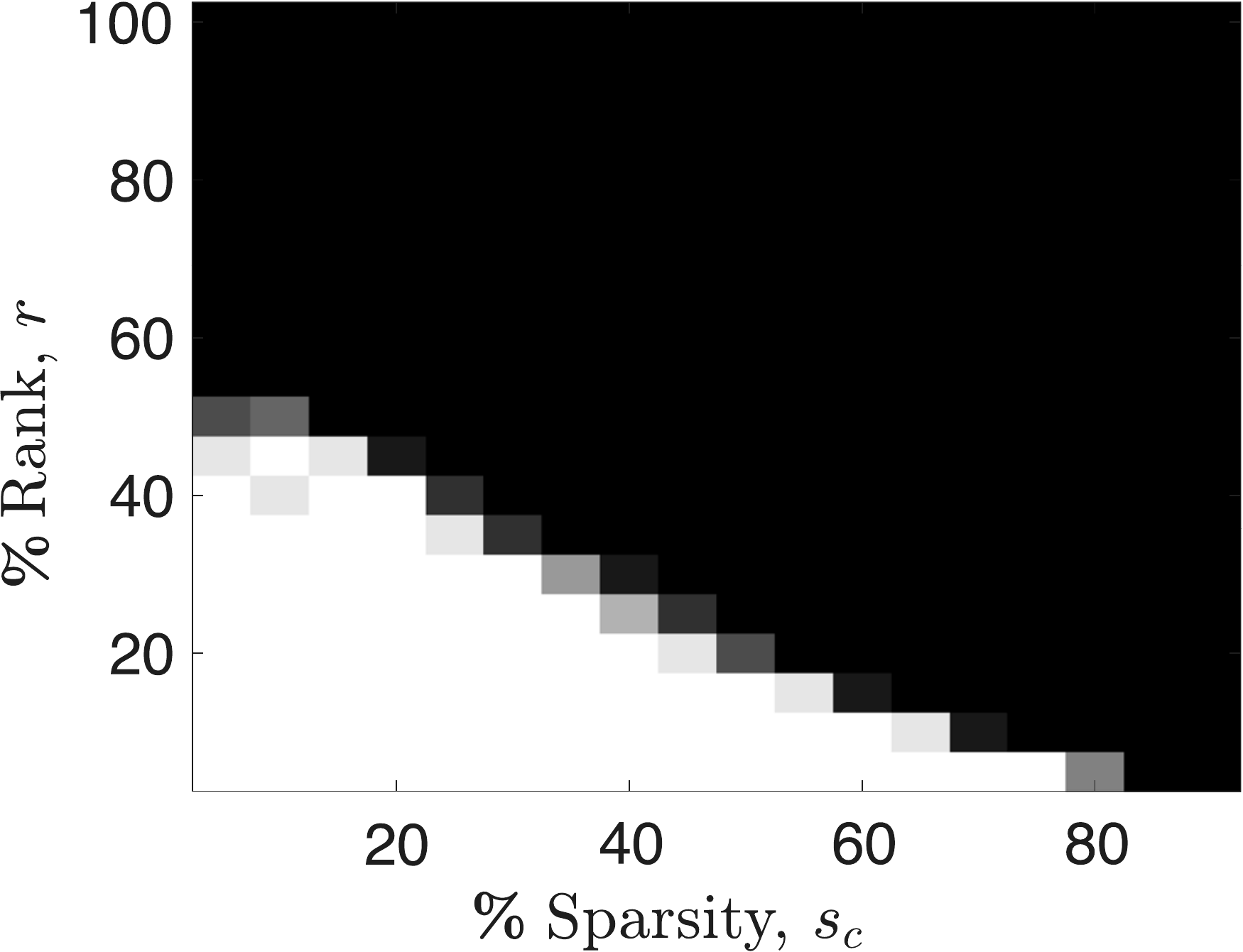} \\
		(a) OP & (b) \ref{Pc}; & (c) \ref{Pc} 
	\end{tabular}
	\vspace{-3pt}
	\caption{\footnotesize Phase transitions in rank $r$ and column sparsity $s_c$ across $10$ Monte-Carlo simulations. Panels (a), (b), (c) show the precision i.e., $\rm(True ~Positives/(True ~Positives + False ~Positives))$ in identifying the outlier columns of $\b{S}$ using (a) OP and (b) \ref{Pc} for \[d=50\], and \ref{Pc} for $d=150$, respectively. In addition, panel (b) also shows the performance by \ref{OP} for $d=50$ in green, marking the region where precision is greater than $0$, super imposed over \ref{Pc}. Here, we threshold the column norms of the recovered $\b{S}$ at $2\times 10^{-3}$ before computing the precision, and a trial is declared successful if it achieves a precision of $0.99$ or higher.}
	\label{fig:phase_dop}
	\vspace{-20pt}  
\end{figure}

\noindent\textbf{Discussion:}
Fig.~\ref{fig:phase_dop} (a)-(c) shows the phase transition in rank $r$ and column-sparsity $s_c$ for the outlier identification performance (in terms of precison) of OP for $d=50$, \ref{Pc} for $d=50$ (and \ref{OP} in green, marking the region where precision is greater than $0$), and  \ref{Pc} for $d=150$, respectively. We observe that the \textit{a priori} knowledge of the dictionary \[\b{D}\] significantly boosts the performance of \ref{Pc} as compared to OP. This showcases the superior outlier identification properties of the proposed technique \ref{Pc}. Further, similar to the entry-wise case, we note that the pseudo-inverse based technique \ref{OP} (in green) fails when $r>d$. For the $d=150$ case the proposed technique \ref{Pc} is able to identify the outlier columns with high precision. Meaning that our technique succeeds even when the outlier columns are not sparse. 

\vspace*{-2pt}
\section{Evaluation of Real-World Dataset: Target Localization in Hyperspectral Imaging}
\label{sec:intro}

\begin{figure}[t]
  \centering
  \begin{tabular}{c}
    \includegraphics[width=0.2\textwidth]{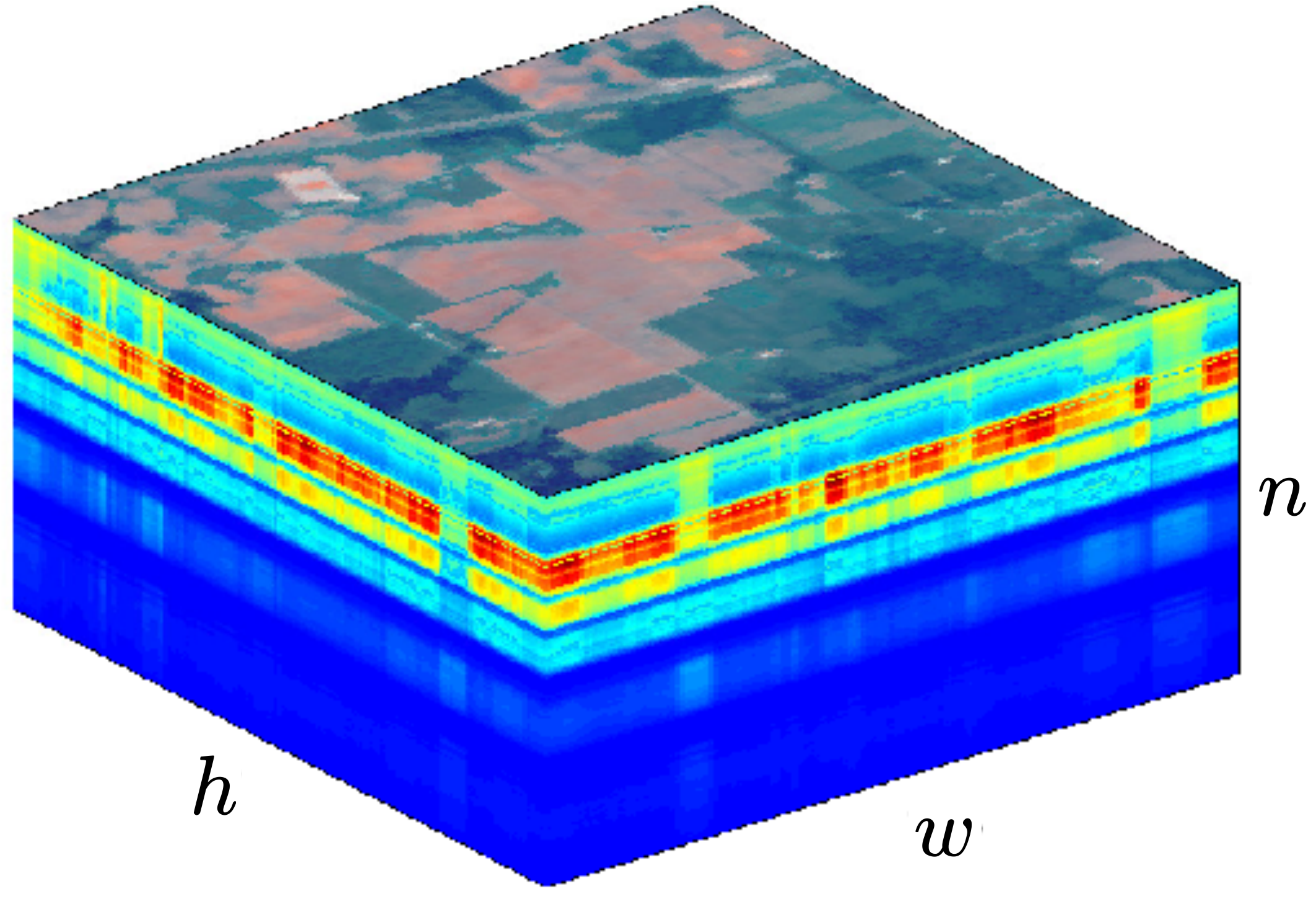}
    \vspace{-2pt}
    \end{tabular}
  \caption{ The HS image data-cube corresponding to the Indian Pines dataset.} 
      \vspace{-4pt}
  \label{fig:data} 
\end{figure}

A HS sensor records the response of a region to different frequencies of the electromagnetic spectrum. As a result, each HS image $\mb{I} \in \mathbb{R}^{ h\times w \times n}$, can be viewed as a data-cube formed by stacking $n$ matrices of size {$h \times w$}, as shown in Fig.~\ref{fig:data}. Here,  $n$ is determined by the number of channels or frequency bands across which measurements of the reflectances are made. Therefore, each volumetric element or \textit{voxel}, of a HS image is a vector of length $n$ corresponding to response of the material to $n$ measurement channels. 

HS images  (when represented as a matrix) are approximately low-rank since a particular scene is composed of only a limited type of objects/materials \cite{Keshava2002}. For instance, while imaging an agricultural area, one would expect to record responses from materials like biomass, farm vehicles, roads, houses, water bodies, and so on. Moreover, the spectra of complex materials can be assumed to be a linear mixture of the constituent materials \cite{Keshava2002, Greer2012}, i.e. the received HS responses can be viewed as being generated by a linear mixture model \cite{Xing2012}. For the target localization task at hand, this approximate low-rank structure is used to decompose a given HS image into a low-rank part, and a component that is sparse in a known dictionary -- a \textit{dictionary sparse} part -- wherein the dictionary is composed of the spectral signatures of the target of interest. We consider the \emph{thin} dictionary setting for the rest of this discussion, since often we aim to localize targets based on a few \textit{a priori} known spectral signatures, although a similar analysis applies for the \textit{fat} case; see Section~\ref{sec:main_result} and \cite{Rambhatla2016}.

Formally, let {$\mb{M}\in\mathbb{R}^{ n \times m}$}, where $m = hw$ be formed by \textit{unfolding} the HS image {$\mb{I}$}, such that, each column of {$\mb{M}$} corresponds to a voxel of the data-cube. We then model {$\mb{M}$} as a superposition of a low-rank component {$\mb{L} \in \mathbb{R}^{n \times m}$} with rank { $r$}, and a dictionary-sparse component, {$\mb{DS}$}, i.e.,
	\begin{align}
	\label{Prob_hs}
	\mb{M} = \mb{L} + \mb{DS}. \vspace{-9pt}
	\end{align}%
Here, {$\mb{D} \in \mathbb{R}^{n \times d}$} represents an \textit{a priori} known dictionary composed of appropriately normalized characteristic responses of the material/object (or the constituents of the material), we wish to \edit{localize}, and {$\mb{S} \in \mathbb{R}^{d \times m}$} refers to the \textit{sparse} coefficient matrix (also referred to as \emph{abundances} in the literature).
Note that {$\mb{D}$} can also be constructed by learning a dictionary based on the known spectral signatures of a target; see \cite{Olshausen97,Aharon05, Mairal10,Lee2007,Rambhatla2019}. 

We now discuss the implementation specifics corresponding to the target localization task. We begin by presenting the algorithm used to solve the optimization problems \ref{Pe} and \ref{Pc}, before discussing the experimental details.

\vspace*{-5pt}

\setlength{\textfloatsep}{5pt}

\begin{algorithm}[!t]
	\caption{APG Algorithm for \ref{Pe} and \ref{Pc}, adapted from \cite{Mardani2012}}
	\label{algo}
	\begin{algorithmic}
		\REQUIRE $\b{M}$, $\b{D}$, $\lambda$, $v$, $\nu_0$, $\bar{\nu}$, and $L_f = \lambda_{max}\left([\b{I} ~~\b{D}]^\top [\b{I} ~~\b{D}] \right)$
		
		\STATE \hspace{-1.25em} \textbf{Initialize:} $\b{L}[0] =  \b{L}[-1] = \textbf{0}_{n\times m}$, $\b{S}[0] = \b{S}[-1] = \textbf{0}_{d\times m}$, $t[0] = t[-1] = 1$, and set $k = 0$.    
		\hspace{-3em} \WHILE {not converged}
		\vspace{6pt}
		\STATE Generate points \[\b{T}_L[k]\] and \[\b{T}_S[k]\] using momentum:
		\vspace{3pt}
		\STATE ~~~~~~$\b{T}_L[k] = \b{L}[k] + \frac{t[k-1]-1}{t[k]}(\b{L}[k] - \b{L}[k-1])$,
		\STATE ~~~~~~$\b{T}_S[k] = \b{S}[k] + \frac{t[k-1]-1}{t[k]}(\b{S}[k] - \b{S}[k-1])$.
		\vspace{6pt}
		\STATE Take a gradient step using these points:
		\vspace{3pt}
		\STATE ~~~~~~$\b{G}_{L}[k] = \b{T}_{L}[k] + \frac{1}{L_f}(\b{M} - \b{T}_L[k]-\b{D}\b{T}_S[k])$,
		\STATE ~~~~~~$\b{G}_{S}[k] = \b{T}_{S}[k] + \frac{1}{L_f}\b{D}^\top(\b{M} - \b{T}_L[k]-\b{D}\b{T}_S[k])$.
		\vspace{6pt}
		\STATE Update Low-rank part via singular value thresholding:
		\vspace{3pt}
		\STATE ~~~~~~$\b{U} \b{\Sigma} \b{V}^\top = \text{svd}(\b{G}_L[k])$,
		\STATE ~~~~~~$\b{L}[k+1] = \b{U} \mathcal{S}_{\nu[k]/L_f}(\b{\Sigma})\b{V}^\top$.
		\vspace{6pt}
		\STATE Update the Dictionary Sparse part:
		\vspace{3pt}
		\STATE ~~~$\b{S}[k+1] = \begin{cases}
		\mathcal{S}_{\nu[k]\lambda_e/L_f}(\b{G}_S[k]), &\text{for \ref{Pe}},\\
		\mathcal{C}_{\nu[k]\lambda_c/L_f}(\b{G}_S[k]), &\text{for \ref{Pc}}.\end{cases}$
		\vspace{6pt}
		\STATE Update the momentum term parameter \[t[k+1]\]:
		\vspace{3pt}
		\STATE ~~~~~~$t[k+1] = \tfrac{1+\sqrt{4t^2[k]+1}}{2}$.
		\vspace{6pt}
		\STATE Update the continuation parameter \[\nu[k+1]\]:
		\vspace{3pt}
		\STATE ~~~~~~$\nu[k+1] = \max\{v\nu[k],\bar{\nu}\}$.
		\vspace{6pt}	 
		\STATE $k$ $\leftarrow$ $k + 1$   
		\hspace{-3em} \ENDWHILE
		\STATE \hspace{-1.25em} \textbf{return} $\b{L}[k]$, $\b{S}[k]$
		\end{algorithmic}
		\end{algorithm}
		\vspace*{-8pt}
		
\subsection{Algorithmic Considerations}
\vspace{-0pt}
\label{sec:optimization}
The optimization problems of interest, \ref{Pe} and \ref{Pc}, for the entry-wise and column-wise case, respectively, are convex but non-smooth. To solve for the components of interest, we adopt the accelerated proximal gradient (APG) algorithm, as shown in Algorithm~\ref{algo}. We here present a unified APG-based algorithm for \ref{Pe} for the both sparsity and dictionary cases, which includes the case considered by \cite{Mardani2012}.


\subsubsection{Discussion of Algorithm~\ref{algo}}

For the optimization problem of interest, we solve an unconstrained problem by transforming the equality constraint to a least-square term which penalizes the fit. In particular, we will accomplish the demixing task by solving the following via the APG-based Algorithm~\ref{algo}.
\begin{align}
\underset{\b{L}, \b{S}}{\min~} \nu\|\b{L}\|_* + \nu\lambda_e \|\b{S}\|_1 + \tfrac{1}{2} \|\b{M} - \b{L} - \b{DS}\|_{\rm F}^2 \label{Pe_apg}
\end{align}
for the entry-wise sparsity case, and
\begin{align}
\underset{\b{L}, \b{S}}{\min~} \nu\|\b{L}\|_* + \nu\lambda_c \|\b{S}\|_{1,2} + \tfrac{1}{2}\|\b{M} - \b{L} - \b{DS}\|_{\rm F}^2, \label{Pc_apg}
\end{align}
for the column-wise sparsity case.

We note that although for the HS application at hand, the thin dictionary case with ($n\geq d$) might be more useful in practice, Algorithm~\ref{algo} allows for the use of fat dictionaries ($n<d$) as well. Specifically, the APG algorithm requires that the gradient of the smooth part,
\begin{align*}
f(\b{L},\b{S}) := \tfrac{1}{2}\|\b{M} - \b{L} - \b{DS}\|_{\rm F}^2 = \tfrac{1}{2}\|\b{M} - \begin{bmatrix}
\b{I} & \b{D}
\end{bmatrix}
\begin{bmatrix}
\b{L}\\\b{S}
\end{bmatrix}\|_{\rm F}^2
\end{align*}
of the convex objectives shown in \eqref{Pe_apg} and \eqref{Pc_apg} is Lipschitz continuous with minimum Lipschitz constant \[L_f\]. Now, since the gradient \[\nabla f(\b{L},\b{S})\] with respect to \[\begin{bmatrix}
\b{L} &\b{S}
\end{bmatrix}\hspace{-2pt}^\top \]is given by 
\begin{align*}
\nabla f(\b{L},\b{S}) =\begin{bmatrix}
\b{I} & \b{D}
\end{bmatrix}^\top(\b{M} - \begin{bmatrix}
\b{I} & \b{D}
\end{bmatrix}
\begin{bmatrix}
\b{L}\\\b{S}
\end{bmatrix}),
\end{align*}
we have that the gradient \[\nabla f\] is Lipschitz continuous as
\begin{align*}
\|\nabla f(\b{L}_1,\b{S}_1) - \nabla f(\b{L}_2,\b{S}_2) \| \leq L_f \|\begin{bmatrix}
\b{L}_1\\\b{S}_1
\end{bmatrix} - \begin{bmatrix}
\b{L}_2\\\b{S}_2
\end{bmatrix}\|,
\end{align*}
for all \[(\b{L}_1,\b{S}_1) , (\b{L}_2,\b{S}_2)\] ~\text{in the domain of} \[f\], where
\begin{align*}
L_f = \|\begin{bmatrix}
\b{I} & \b{D}
\end{bmatrix}^\top\begin{bmatrix}
\b{I} & \b{D}
\end{bmatrix}\|= \lambda_{\max} (\begin{bmatrix}
\b{I} & \b{D}
\end{bmatrix}^\top\begin{bmatrix}
\b{I} & \b{D}
\end{bmatrix}).
\end{align*}

The update of the low-rank component  and the sparse matrix \[\b{S}\] for the entry-wise case both involve a soft thresholding step, \[\mathcal{S}_{\tau}(.)\], where for a matrix \[\b{Y}\], \[\mathcal{S}_{\tau}(\b{Y}_{ij})\] is defined as 
\begin{align*}
\mathcal{S}_{\tau}(\b{Y}_{ij}) = {\rm sgn}{(\b{Y}_{ij})}\max({|\b{Y}_{ij} - \tau|, 0}).
\end{align*}
In case of the low-rank part we apply this function to the singular values (therefore referred to as \textit{singular value thresholding}) \cite{Toh10}, while for the update of the dictionary sparse component, we apply it to the sparse coefficient matrix \[\b{S}\].

The low-rank update step remains the same as for the entry-wise case. For the update of the column-wise case, we threshold the columns of \[\b{S}\] based on their column norms, i.e., for a column \[\b{Y}_j\] of a matrix \[\b{Y}\], the column-norm based soft-thresholding function, \[\mathcal{C}_{\tau}(.)\] is defined as 
\begin{align*}
\mathcal{C}_{\tau}(\b{Y}_{j}) = \max({\b{Y}_{j} - \tau{\b{Y}_{j}}/{\|\b{Y}_{j}\|}}).
\end{align*}


\subsubsection{Parameter Selection}\label{sec:opt_ew_param}

We adopt a grid-search strategy over the range of admissible values to find the best values of the regularization parameters. 

\noindent\emph{Selecting parameters for the entry-wise case:} The choice of parameters \[\nu\] and \[\lambda_e\] in Algorithm~\ref{algo} is based on the optimality conditions of the optimization problem shown in \eqref{Pe_apg}. As presented in \cite{Mardani2012}, the range of parameters \[\nu\] and \[\nu\lambda_e\] associated with the low-rank part \[\b{L}\] and the sparse coefficient matrix \[\b{S}\], respectively, lie in \[\nu \in \{0,\|\b{M}\|\}\] and \[\nu\lambda_e \in \{0,\|\b{D}^\top\b{M}\|_\infty\}\], i.e., for Algorithm~\ref{algo} \[\nu_0 = \|\b{M}\|\].

These ranges for are derived using the optimization problem shown in \eqref{Pe_apg}. Specifically, we find the largest values of these regularization parameters which yield a \[(\b{0}, \b{0})\] solution for the pair \[(\b{L}_0, \b{S}_0)\] by analyzing the optimality conditions of \eqref{Pe_apg}. This value of the regularization parameter then defines the upper bound on the range. For instance,  the optimality condition for \[\lambda_*:=\nu\] and \[\lambda_1 := \nu\lambda_e\], is given by
\begin{align*}
\lambda_*\partial_\b{L} \|\b{L}\|_*   - (\b{M} - \b{L} - \b{DS}) = 0,
\end{align*}
where the sub-differential set \[\partial_\b{L} \|\b{L}\|_* \] is defined as
\begin{align*}
\partial_\b{L} \|\b{L}\|_* \Bigr|_{\b{L} = \b{L}_0} = \{ \b{UV^\top} + \b{W} : \|\b{W}\| \leq 1, \c{P}_{\c{L}} (\b{W}) = \b{0} \}.	
\end{align*}
Therefore, for a zero solution pair \[(\b{L}_0, \b{S}_0)\] we have that
\begin{align*}
\{ \lambda_*\b{W} =  \b{M}: \|\b{W}\| \leq 1, \c{P}_{\c{L}} (\b{W}) = \b{0} \},
\end{align*}
which yields the condition that \[\|\b{M}\| \leq \lambda_*\]. Therefore, the maximum value of \[\lambda_*\] which drives the low-rank part to an all-zero solution is \[\|\b{M}\|\]. Similarly,  the optimality condition for the dictionary sparse component to choose \[\lambda_1\] is given by
\begin{align*}
\lambda_1 \partial_\b{S} \|\b{S}\|_1  - \b{D}^\top (\b{M} - \b{L} - \b{DS}) = 0,
\end{align*}
where the the sub-differential set \[\partial_\b{S} \|\b{S}\|_1\] is defined as
\begin{align*}
\partial_\b{S} \|\b{S}\|_1 \Bigr|_{\b{S} = \b{S}_0}  = \{ \text{sign}(\b{S}_0) + \b{F} : \|\b{F}\|_{\infty} \leq 1, \c{P}_{\c{S}_e} (\b{F}) = \b{0} \}.
\end{align*}
Again, for a zero solution pair \[(\b{L}_0, \b{S}_0)\] we need that
\begin{align*}
\{ \lambda_1\b{F} = \b{D}^\top \b{M} : \|\b{F}\|_{\infty} \leq 1, \c{P}_{\c{S}_e} (\b{F}) = \b{0} \},
\end{align*}
which implies that \[\|\b{D}^\top \b{M}\|_\infty \leq \lambda_1\], i.e. the maximum value of \[\lambda_1\] that drives the dictionary sparse part to zero is \[\|\b{D}^\top \b{M}\|_\infty\]. 
\begin{figure}[!t]
	\centering
	\scalebox{0.9}{
	\begin{tabular}{cc}
		\hspace{-4pt}\includegraphics[width=0.58\linewidth]{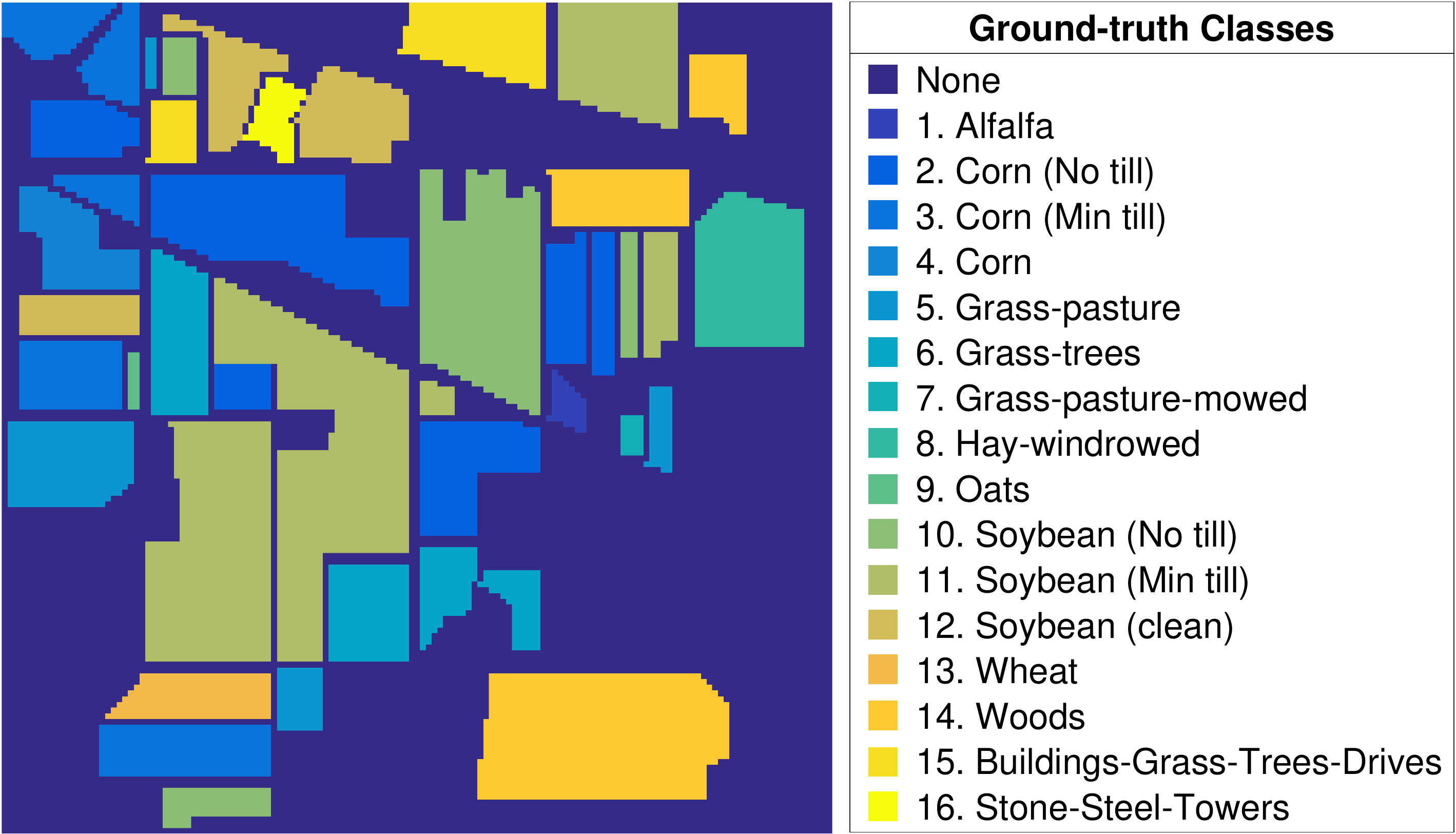}&\hspace{-14pt}
		\includegraphics[width=0.405\linewidth]{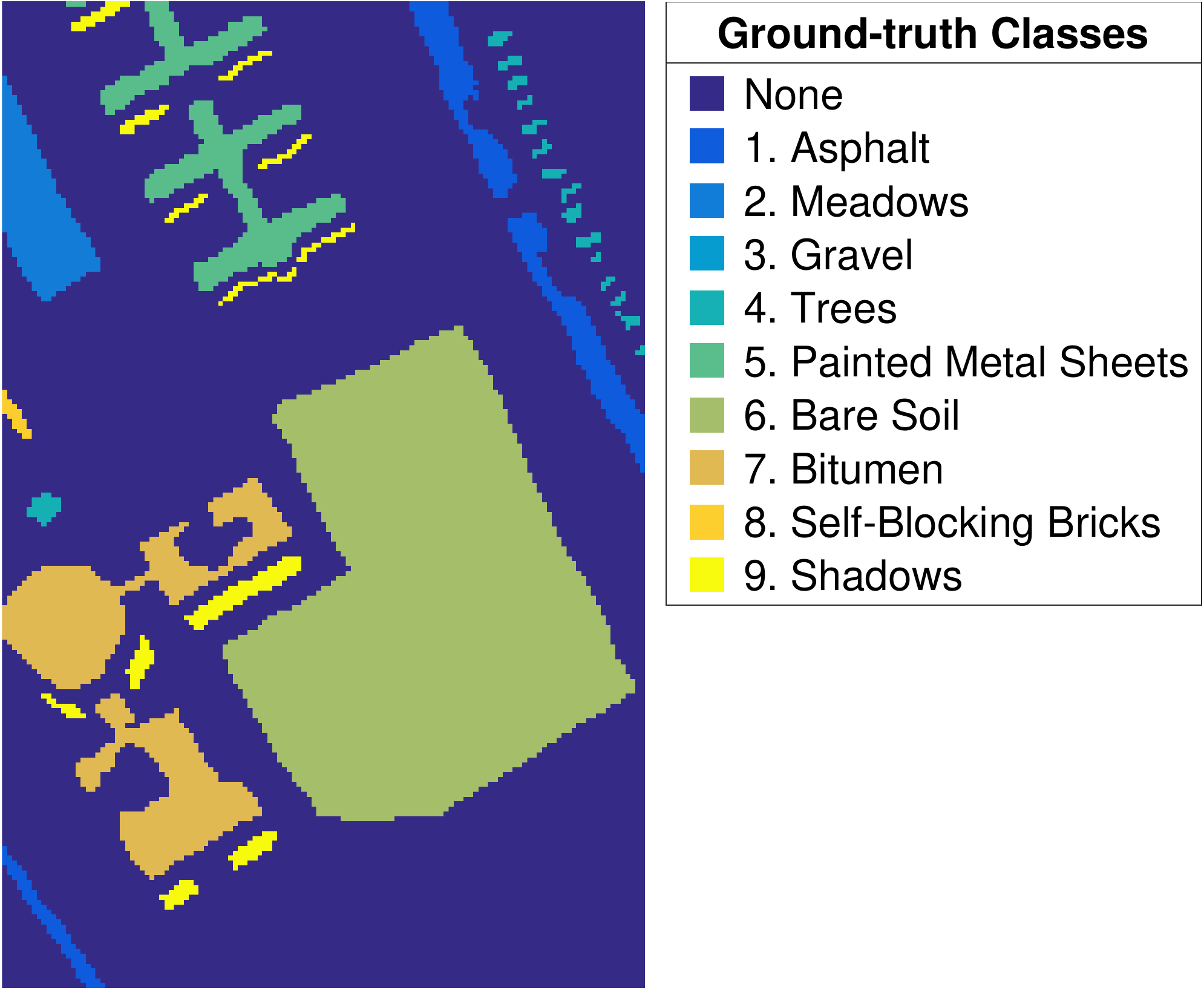}\\
		\hspace{-4pt}	(a) Indian Pines &\hspace{-14pt} (b) Pavia University\\
	\end{tabular}}
		\vspace{-2pt}
	\caption{Ground-truth classes. Panels (a) and (b) show the ground truth classes for the Indian Pines dataset \cite{HSdat} and Pavia University dataset \cite{PaviaUSdat}, respectively.} 
	\label{fig:gt_data} 
\end{figure}

\begin{figure}[!b]
	\centering
	\begin{tabular}{cc}
		\hspace{-4pt}\includegraphics[width=0.35\linewidth]{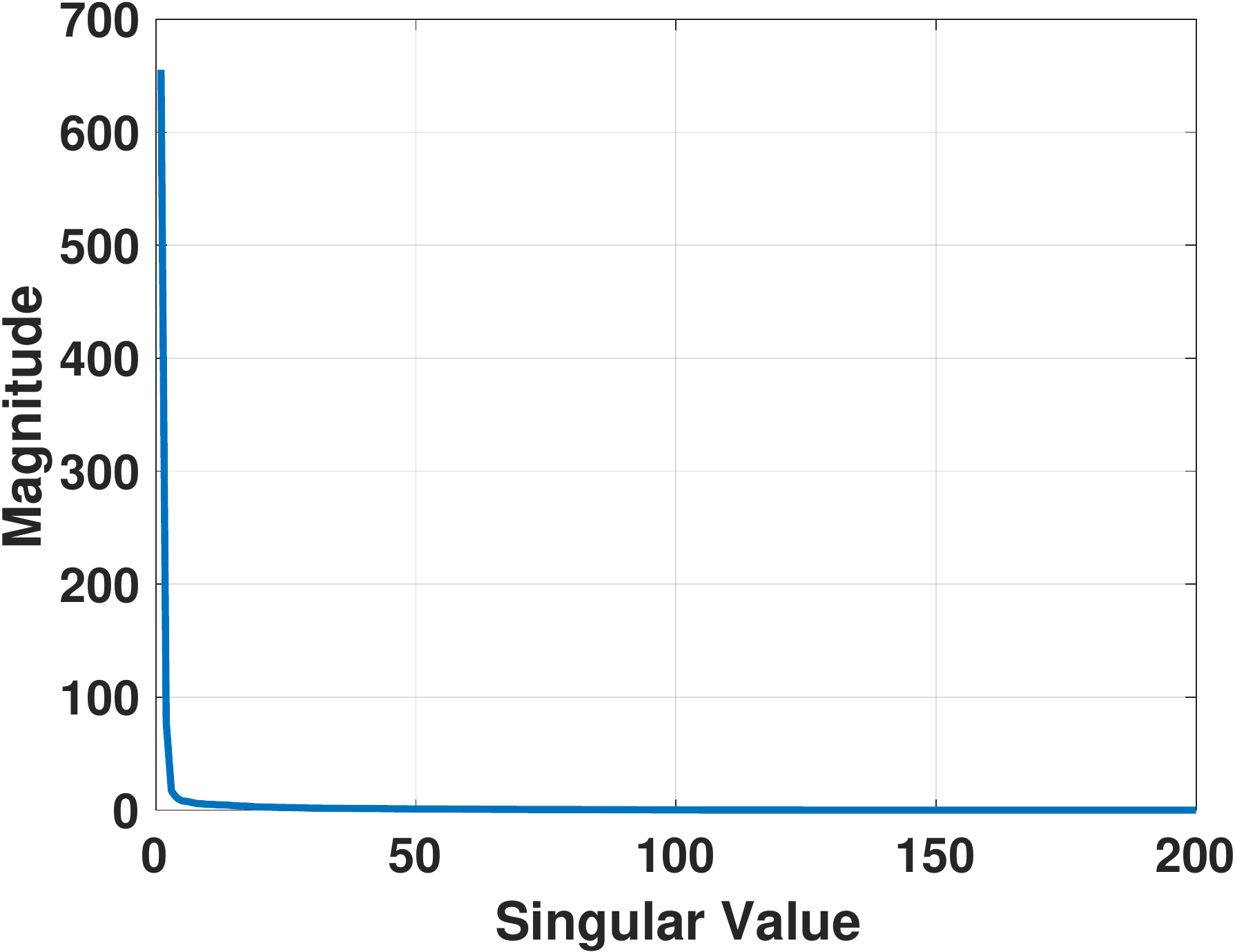}&\hspace{-14pt}
		\includegraphics[width=0.35\linewidth]{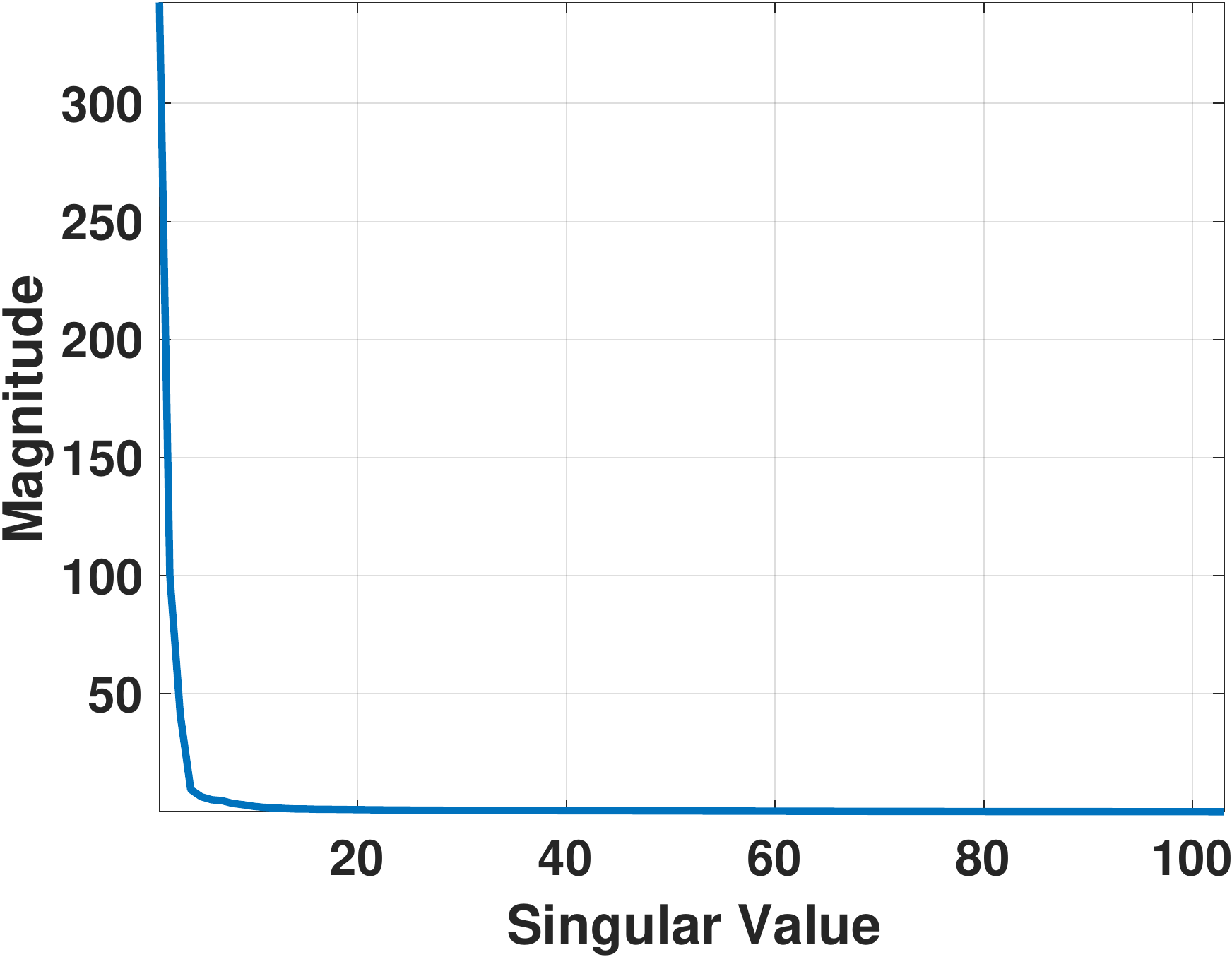}\\
		\hspace{-4pt} (a) Indian Pines Dataset & (b) Pavia University Dataset\\
	\end{tabular}
	\vspace{-4pt}
	\caption{Sorted Singular Values of the datasets. Panels (a) and (b) show the decay of singular values for the Indian Pines dataset \cite{HSdat} and Pavia University dataset \cite{PaviaUSdat}, respectively.} 
	\label{fig:sing_val_data} 
	\vspace{-8pt}
\end{figure}
\noindent\emph{Selecting parameters for the column-wise case:} Again, the choice of parameters \[\nu\] and \[\lambda_c\] is derived from the optimization problem shown in \eqref{Pc_apg}. In this case, the range of parameters \[\nu\] and \[\nu\lambda_c\] associated with the low-rank part \[\b{L}\] and the sparse coefficient matrix \[\b{S}\], respectively, lie in \[\nu \in \{0,\|\b{M}\|\}\] and \[\nu\lambda_e \in \{0,\|\b{D}^\top\b{M}\|_{\infty,2}\}\], i.e., for Algorithm~\ref{algo} \[\nu_0 = \|\b{M}\|\]. The range of regularization parameters are evaluated using the analysis similar to the entry-wise case, we use the optimality conditions for \eqref{Pc_apg}, instead of \eqref{Pe_apg}.
			\begin{table*}[!htbp]
			\caption{Entry-wise sparsity model for the Indian Pines Dataset. Simulation results are presented for our proposed approach \eqref{Pe}, robust-PCA based approach on transformed data \[\b{D^\dagger M}\] \eqref{RPCA}, matched filtering (\textcolor{blue}{MF}) on original data \[\b{M}\], and matched filtering on transformed data \[\b{D^\dagger M}\] (\textcolor{blue}{MF$^\dagger$}), across dictionary elements $d$, and the regularization parameter for initial dictionary learning procedure $\rho$; See \eqref{eq:dl}. Threshold selects columns with column-norm greater than threshold such that AUC is maximized.  For each case, the best performing metrics are reported in bold for readability. Further, $`` * "$ denotes the case where ROC curve was ``flipped'' (i.e. classifier output was inverted to achieve the best performance).}
			\vspace{-4pt}
			\label{res_tab}
		\captionsetup{justification=centering}
			\begin{subtable}{.5\linewidth}
			\captionsetup{font=small}
			 \centering
			 \caption{Learned dictionary, $d=4$}
			 \vspace{-5pt}
		   	\scalebox{0.9}{
		   	\begin{tabular}{|P{0.5cm}|c|c|c|c|c|c|}
			\hline
			\multirow{2}{*}{\textbf{$d$}} & \multirow{2}{*}{$\rho$} & \multirow{2}{*}{\textbf{Method}}& \multirow{2}{*}{\textbf{Threshold }}&\multicolumn{2}{G|}{\textbf{Performance at best operating point} }& \multirow{2}{*}{\textbf{AUC}}\\ \cline{5-6}
			&&&&\textbf{TPR}&\textbf{FPR}&\\\hline
			\multirow{12}{*}{4} &	\multirow{4}{*}{0.01}
			  &\textbf{D-RPCA(E)}					&0.300&\textbf{0.979}&\textbf{0.023}&\textbf{0.989}\\\cline{3-7}
			&&\textbf{RPCA$^\dagger$} &0.650&0.957&0.049&0.974\\\cline{3-7}
			&&\textbf{MF$_*$}						&N/A&0.957&0.036&0.994\\\cline{3-7}
			&&\textbf{MF$_*^\dagger$}	   &N/A&0.914&0.104&0.946\\\cline{2-7}
			&	\multirow{4}{*}{0.1}
		  &\textbf{D-RPCA(E)}				    	&0.800&\textbf{0.989}&0.017&0.997\\\cline{3-7}
			&&\textbf{RPCA$^\dagger$} &0.800&0.989&0.014&0.997\\\cline{3-7}
			&&\textbf{MF}						&N/A&0.989&0.016&0.998\\\cline{3-7}
			&&\textbf{MF$^\dagger$}	   &N/A&0.989&\textbf{0.010}&\textbf{0.998}\\\cline{2-7}
			&	\multirow{4}{*}{0.5}
		  &\textbf{D-RPCA(E)}					    &0.600&\textbf{0.968}&\textbf{0.031}&\textbf{0.991}\\\cline{3-7}
			&&\textbf{RPCA$^\dagger$} &0.600&0.935&0.067&0.988\\\cline{3-7}
			&&\textbf{MF}						&N/A&0.548&0.474&0.555\\\cline{3-7}
			&&\textbf{MF$_*^\dagger$}	   &N/A&0.849&0.119&0.939\\\hline
			\end{tabular}
			\label{dl_4}
			}
			\end{subtable} 
			\begin{subtable}{.5\linewidth}
			\captionsetup{font=small}
			 \centering
			 \caption{Learned dictionary, $d=10$}
			 \vspace{-5pt}
			   	\scalebox{0.9}{
			   	\begin{tabular}{|P{0.5cm}|c|c|c|c|c|c|}
				\hline
				\multirow{2}{*}{\textbf{$d$}} & \multirow{2}{*}{$\rho$} & \multirow{2}{*}{\textbf{Method}}& \multirow{2}{*}{\textbf{Threshold}}&\multicolumn{2}{G|}{\textbf{Performance at best operating point} }& \multirow{2}{*}{\textbf{AUC}}\\ \cline{5-6}
					&&&&\textbf{TPR}&\textbf{FPR}&\\\hline
				\multirow{12}{*}{10} &	\multirow{4}{*}{0.01}
				  &\textbf{D-RPCA(E)}					&0.600&0.935&0.060&0.972\\\cline{3-7}
				&&\textbf{RPCA$^\dagger$} &0.700&\textbf{0.978}&\textbf{0.023}&\textbf{0.990}\\\cline{3-7}
				&&\textbf{MF$_*$}						&N/A&0.624&0.415&0.681\\\cline{3-7}
				&&\textbf{MF$^\dagger_*$}	   &N/A&0.569&0.421&0.619\\\cline{2-7}
				&	\multirow{4}{*}{0.1}
			  &\textbf{D-RPCA(E)}					    &0.500&\textbf{0.968}&\textbf{0.029}&\textbf{0.993}\\\cline{3-7}
				&&\textbf{RPCA$^\dagger$} &0.500&0.871&0.144&0.961\\\cline{3-7}
				&&\textbf{MF$_*$}						&N/A&0.688&0.302&0.713\\\cline{3-7}
				&&\textbf{MF$^\dagger$}	   &N/A&0.527&0.469&0.523\\\cline{2-7}
				&	\multirow{4}{*}{0.5}
			  &\textbf{D-RPCA(E)}					    &1.000   &\textbf{0.978}&\textbf{0.031}&\textbf{0.996}\\\cline{3-7}
				&&\textbf{RPCA$^\dagger$} &2.200&0.849&0.113&0.908\\\cline{3-7}
				&&\textbf{MF}						&N/A&0.807&0.309&0.781\\\cline{3-7}
				&&\textbf{MF$^\dagger_*$}	   &N/A&0.527&0.465&0.539\\\hline
				\end{tabular}
				\label{dl_10}
				}
				\end{subtable}\vspace{5pt}
				\begin{subtable}{0.5\linewidth}
				\captionsetup{font=small}
				 \centering
				 \caption{Dictionary by sampling voxels, $d=15$}
				 \vspace{-5pt}
					   	\scalebox{0.9}{
					   	\begin{tabular}{|P{0.5cm}|c|c|c|P{1.5cm}|P{0.9cm}|}
						\hline 
						\multirow{2}{*}{\textbf{$d$}}  & \multirow{2}{*}{\textbf{Method}}& \multirow{2}{*}{\textbf{Threshold}}&\multicolumn{2}{>{\bfseries\centering\arraybackslash} m{2.3cm}|}{\textbf{Performance at best operating point} }& \multirow{2}{*}{\textbf{AUC}}\\ \cline{4-5}
						&&&TPR&FPR&\\\hline
						\multirow{4}{*}{15}
						  &\textbf{D-RPCA(E)}					&0.300&\textbf{0.989}&\textbf{0.021}&\textbf{0.998}\\\cline{2-6}
						&\textbf{RPCA$^\dagger$} &3.000&0.849&0.146&0.900\\\cline{2-6}
						&\textbf{MF}						&N/A&0.957&0.085&0.978\\\cline{2-6}
						&\textbf{MF$^\dagger$}	   &N/A&0.796&0.217&0.857\\\hline
						\end{tabular}
						\label{orig_dict}
						}
						\end{subtable} 
							\begin{subtable}{.5\linewidth}
								\captionsetup{font=small}
								\centering
						\caption{Average performance}
						\vspace*{-5pt}
						\scalebox{0.88}{	
						\begin{tabular}{|P{1.5cm}|c|c|c|c|c|c|}		
						\hline
						\multirow{2}{*}{\textbf{Method}}& \multicolumn{2}{G|}{\textbf{TPR}}& \multicolumn{2}{G|}{\textbf{FPR}}& \multicolumn{2}{G|}{\textbf{AUC}}\\ \cline{2-7}
						&Mean&St.Dev.&Mean&St.Dev.&Mean&St.Dev.\\ \hline
						\textbf{D-RPCA(E)}&	\textbf{0.972}&	\textbf{0.019}&	\textbf{0.030}&	\textbf{0.014}&	\textbf{0.991}&	\textbf{0.009}\\ \hline
						\textbf{RPCA$^\dagger$}&0.919&0.061&0.079&0.055&0.959&0.040\\ \hline
						\textbf{MF}&0.796&0.179&0.234&0.187&0.814&0.178\\ \hline
						\textbf{MF$^\dagger$}	&0.739&0.195&0.258&0.192&0.775&0.207\\ \hline
						\end{tabular}
						\label{overall_perf}
						}
						\end{subtable}
			\captionsetup{font=small}
			\vspace{-15pt}
			\end{table*}

\vspace*{-12pt}
\subsection{Experimental Evaluation}
\label{sec:exp}
We now evaluate the performance of the proposed technique on real-world HS data. We begin by introducing the dataset\footnote{Available via \url{http://www.ehu.eus/ccwintco/index.php?title=Hyperspectral\_Remote\_Sensing\_Scenes}.} used for these simulations, following which we describe the experimental set-up and present the results.

\vspace*{5pt}
\noindent\emph{Data} 
\\
\noindent\textbf{Indian Pines Dataset}:
We first consider the ``Indian Pines'' dataset \cite{HSdat}, which was collected over the Indian Pines test site in North-western Indiana in the June of 1992 using the Airborne Visible/Infrared Imaging Spectrometer (AVIRIS) \cite{AVIRIS} sensor, a popular choice for collecting HS images for various remote sensing applications. This dataset consists of spectral reflectances across $224$ bands in wavelength of ranges $400-2500$ nm from a scene which is composed mostly of agricultural land along with two major dual lane highways, a rail line and some built structures, as shown in Fig.~\ref{fig:gt_data}(a). The dataset is further processed by removing the bands corresponding to those of water absorption, which results in a HS data-cube with dimensions $\{145 \times 145 \times 200\}$ is as visualized in Fig.~\ref{fig:data}. Here, $n = 200$, $h = w = 145$, and therefore $m = hw= 145\times 145$. This modified dataset is available as ``corrected Indian Pines'' dataset \cite{HSdat}, with the ground-truth containing $16$ classes; henceforth, referred to as the ``Indian Pines Dataset". We form the data matrix  $\b{M} \in \mathbb{R}^{n \times m}$ by stacking each voxel of the image side-by-side, which results in a $\{200 \times 145^2\}$ data matrix $\b{M}$. We will analyze the performance of the proposed technique for the identification of the stone-steel towers (class $16$ in the dataset), shown in Fig.~\ref{fig:gt_data}(a), constituting $93$ voxels.  

\vspace{2pt}
\noindent\textbf{Pavia University Dataset}:
Acquired using Reflective Optics System Imaging Spectrometer (ROSIS) sensor, the Pavia University Dataset \cite{PaviaUSdat} consists of spectral reflectances across $103$ bands (in the range $430-860$ nm) of an urban landscape over northern Italy. The selected subset of the scene, a $\{201 \times 131 \times 103\}$ data-cube, mainly consists of buildings, roads, painted metal sheets and trees, as shown in Fig.~\ref{fig:gt_data}(b). Note that class-$3$ corresponding to ``Gravel'' is not present in the selected data-cube considered here. For our demixing task, we will analyze the localization of target class $5$, corresponding to the painted metal sheets, which constitutes $707$ voxels in the scene. Note that for this dataset $h= 201$, $w = 131$, $m = hw= 201 \times 131$ and $n = 103$.  

Further, in Fig.~\ref{fig:sing_val_data} we show the decay of singular values of the Indian Pines and the Pavia University dataset. We note that indeed the presence of a limited number of materials makes the these datasets approximately low-rank. 

\setlength{\textfloatsep}{5pt}

	    	\begin{table*}[!bhtp]
				\caption{Column-wise sparsity model and Indian Pines Dataset. Simulation results are presented for the proposed approach (\ref{Pc}), Outlier Pursuit (OP) based approach on transformed data (\ref{OP}), matched filtering (\textcolor{blue}{MF}) on original data $\b{M}$, and matched filtering on transformed data $\b{D^\dagger M}$ (\textcolor{blue}{MF$^\dagger$}), across dictionary elements $d$, and the regularization parameter for initial dictionary learning step $\rho$; See \eqref{eq:dl}. Threshold selects columns with column-norm greater than threshold such that AUC is maximized. For each case, the best performing metrics are reported in bold for readability. Further, $`` * "$ denotes the case where ROC curve was ``flipped'' (i.e. classifier output was inverted to achieve the best performance).}
				\vspace{-4pt}
				\label{res_tab_ip_cw}
				\captionsetup{justification=centering}
				\begin{subtable}{0.5\linewidth}
					\captionsetup{font=footnotesize}
					\centering
					\caption{Learned dictionary, $d=4$}
					\vspace{-5pt}
					\scalebox{0.9}{
						\begin{tabular}{|P{0.5cm}|c|c|c|c|c|c|}
							\hline
							\multirow{2}{*}{\textbf{$d$}} & \multirow{2}{*}{$\rho$} & \multirow{2}{*}{\textbf{Method}}& \multirow{2}{*}{\textbf{Threshold }}&\multicolumn{2}{G|}{\textbf{Performance at best operating point} }& \multirow{2}{*}{\textbf{AUC}}\\ \cline{5-6}
							&&&&\textbf{TPR}&\textbf{FPR}&\\\hline
							\multirow{12}{*}{4} &	\multirow{4}{*}{0.01}
							&\textbf{D-RPCA(C)}					&0.905&\textbf{0.989}&\textbf{0.014}&\textbf{0.998}\\\cline{3-7}
							&&\textbf{OP$^\dagger$}        &0.895&0.989&0.015&0.998\\\cline{3-7}
							&&\textbf{MF$_*$}                   &N/A&0.656&0.376&0.611\\\cline{3-7}
							&&\textbf{MF$_*^\dagger$}    &N/A&0.624&0.373&0.639\\\cline{2-7}
							&	\multirow{4}{*}{0.1}
							&\textbf{D-RPCA(C)}	              &0.805&\textbf{0.989}&\textbf{0.013}&\textbf{0.998}\\\cline{3-7}
							&&\textbf{OP$^\dagger_*$}      &1.100&0.720&0.349&0.682\\\cline{3-7}
							&&\textbf{MF$_*$}                 &N/A&0.742&0.256&0.780\\\cline{3-7}
							&&\textbf{MF$^\dagger$}     &N/A&0.828&0.173&0.905\\\cline{2-7}
							&	\multirow{4}{*}{0.5}
							&\textbf{D-RPCA(C)}				&1.800&\textbf{0.989}&\textbf{0.010}&\textbf{0.998}\\\cline{3-7}
							&&\textbf{OP$^\dagger$}    &1.300&0.989&0.012&0.998\\\cline{3-7}
							&&\textbf{MF}                      &N/A&0.548&0.474&0.556\\\cline{3-7}
							&&\textbf{MF$_*^\dagger$}&N/A&0.849&0.146&0.939\\\hline
						\end{tabular}
						\label{dl_4_ip_cw}
					
					}
				\end{subtable} 		\vspace*{-5pt}
				\begin{subtable}{0.5\linewidth}
					\captionsetup{font=footnotesize}
					\centering
					\caption{Learned dictionary, $d=10$}
					\vspace{-5pt}
					\scalebox{0.9}{
						\begin{tabular}{|P{0.5cm}|c|c|c|c|c|c|}
							\hline
							\multirow{2}{*}{\textbf{$d$}} & \multirow{2}{*}{$\rho$} & \multirow{2}{*}{\textbf{Method}}& \multirow{2}{*}{\textbf{Threshold}}&\multicolumn{2}{G|}{\textbf{Performance at best operating point} }& \multirow{2}{*}{\textbf{AUC}}\\ \cline{5-6}
							&&&&\textbf{TPR}&\textbf{FPR}&\\\hline
							\multirow{12}{*}{10} &	\multirow{4}{*}{0.01}
							&\textbf{D-RPCA(C)}					&0.800&\textbf{0.946}&\textbf{0.016}&\textbf{0.993}\\\cline{3-7}
							&&\textbf{OP$^\dagger$}         &1.300&0.946&0.060&0.988\\\cline{3-7}
							&&\textbf{MF$_*$}						&N/A&0.946&0.060&0.987\\\cline{3-7}
							&&\textbf{MF$^\dagger_*$}	   &N/A&0.527&0.468&0.511\\\cline{2-7}
							&	\multirow{4}{*}{0.1}
							&\textbf{D-RPCA(C)}					    &0.550&\textbf{0.979}&\textbf{0.029}&\textbf{0.997}\\\cline{3-7}
							&&\textbf{OP$^\dagger$}             &0.800&0.893&0.112&0.928\\\cline{3-7}
							&&\textbf{MF$_*$}						&N/A&0.688&0.302&0.714\\\cline{3-7}
							&&\textbf{MF$^\dagger$}	          &N/A&0.527&0.470&0.523\\\cline{2-7}
							&	\multirow{4}{*}{0.5}
							&\textbf{D-RPCA(C)}					    &1.400  &\textbf{0.989}&\textbf{0.037}&\textbf{0.997}\\\cline{3-7}
							&&\textbf{OP$^\dagger$}             &0.800&0.807&0.148&0.847\\\cline{3-7}
							&&\textbf{MF}						        &N/A&0.807&0.309&0.781\\\cline{3-7}
							&&\textbf{MF$^\dagger_*$}	      &N/A&0.527&0.468&0.539\\\hline
						\end{tabular}
						\label{dl_10_ip_cw}
					}
				\end{subtable}\vspace{10pt}\\
				\hspace{-2pt}
				\begin{subtable}{0.5\linewidth}
					\captionsetup{font=small}
					\centering
					\caption{Dictionary by sampling voxels, $d=15$}
					\vspace{-5pt}
					\scalebox{0.9}{
						   	\begin{tabular}{|P{0.5cm}|c|c|c|P{1.5cm}|P{0.9cm}|}
							\hline 
							\multirow{2}{*}{\textbf{$d$}}  & \multirow{2}{*}{\textbf{Method}}& \multirow{2}{*}{\textbf{Threshold}}&\multicolumn{2}{>{\bfseries\centering\arraybackslash} m{2.3cm}|}{\textbf{Performance at best operating point} }& \multirow{2}{*}{\textbf{AUC}}\\ \cline{4-5}
							&&&TPR&FPR&\\\hline
							\multirow{4}{*}{15}
							&\textbf{D-RPCA(C)}					&0.800&0.989&0.018&0.998\\\cline{2-6}
							&\textbf{OP$^\dagger$}           &2.200&0.882&0.126&0.900\\\cline{2-6}
							&\textbf{MF}						     &N/A&0.957&0.085&0.978\\\cline{2-6}
							&\textbf{MF$^\dagger$}	        &N/A&0.796&0.217&0.857\\\hline
						\end{tabular}
						\label{orig_dict_ip_cw}
					}
					\end{subtable}
					\begin{subtable}{0.5\linewidth}
					\captionsetup{font=small}
					\caption{Average performance}
					\vspace*{-5pt}
					\centering
					\scalebox{0.88}{	
						\begin{tabular}{|P{1.5cm}|c|c|c|c|c|c|}		
							\hline
							\multirow{2}{*}{\textbf{Method}}& \multicolumn{2}{G|}{\textbf{TPR}}& \multicolumn{2}{G|}{\textbf{FPR}}& \multicolumn{2}{G|}{\textbf{AUC}}\\ \cline{2-7}
							&Mean&St.Dev.&Mean&St.Dev.&Mean&St.Dev.\\ \hline
							\textbf{D-RPCA(C)}&	\textbf{0.981}&	\textbf{0.016}&	\textbf{0.020}&	\textbf{0.010}&	\textbf{0.997}&	\textbf{0.002}\\ \hline
							\textbf{OP$^\dagger$}&0.889&0.099&0.117&0.115&0.906&0.114\\ \hline
							\textbf{MF}                  &0.763&0.151&0.266&0.149&0.772&0.166\\ \hline
							\textbf{MF$^\dagger$}	&0.668&0.151&0.331&0.148&0.702&0.192\\ \hline
						\end{tabular}
						\label{overall_perf_ip_cw}
					}\vspace{16pt}
				\end{subtable}
				\captionsetup{font=small}
				\vspace*{-18pt}
			\end{table*}

	    \vspace*{2pt}
\noindent \emph{Dictionary}: We form the known dictionary \[\b{D}\] two ways: 1) where a (thin) dictionary is learned based on the voxels by solving \eqref{eq:dl}, and 2) when the dictionary is formed by randomly sampling voxels from the target class. This is to emulate the ways in which we can arrive at the dictionary corresponding to a target -- 1) where the \textit{exact signatures} are not available, and/or there is noise, and 2) where we have access to the exact signatures of the target, respectively. 

In our experiments for case 1), we learn a dictionary using the target class data \[\b{Y}\in \RR^{n \times p} \] by alternating between updating the sparse coefficients via FISTA \cite{Beck09} and dictionary via the Newton method \cite{Wright06}, approximately solving the following optimization problem \cite{Olshausen97,Aharon05, Mairal10,Lee2007}.

\vspace*{3pt}
\begin{align}\label{eq:dl}
\hat{\b{D}} = \underset{\b{D}:\|\b{D}_i\|=1, \b{A}}{{\rm arg. min}} ~ \|\b{Y} - \b{DA}\|_{\rm F}^2 + \rho \|\b{A}\|_1,
\end{align}
\vspace*{-1pt}

For case 2), the columns of the dictionary are set as the known data voxels of the target class. Specifically, instead of learning a dictionary based on a target class of interest, we set it as the exact signatures observed previously. Note that for this case, the dictionary is not normalized at this stage since the specific normalization depends on the particular demixing problem of interest, discussed shortly. In practice, we can store the un-normalized dictionary  \[\b{D}\] (formed from the voxels), consisting of actual \textit{signatures} of the target material, and can normalize it after the HS image has been acquired.\\

\vspace*{-8pt}

\noindent\emph{Experimental Setup}\label{sec:exp_setup}

\vspace{2pt}
\noindent\textbf{Normalization:} For normalizing the data, we divide each element of the data matrix $\b{M}$ by $\|\b{M}\|_{\infty}$ to preserve the inter-voxel scaling. For the dictionary, in the learned dictionary case, i.e., case 1), the dictionary already has unit-norm columns. Further, when the dictionary is formed from the data directly, i.e., for case 2), we divide each element of \[\b{D}\] by $\|\b{M}\|_{\infty}$, and then normalize the columns of \[\b{D}\], such that they are unit-norm.

\vspace{2pt}
\noindent\textbf{Dictionary selection for the Indian Pines Dataset}: For the learned dictionary case, we evaluate the performance of the aforementioned techniques for both entry-wise and column-wise settings for two dictionary sizes, $d=4$ and $d=10$, for three values of the regularization parameter $\rho$, used for the initial dictionary learning step, i.e., $\rho = 0.01,~0.1$ and $0.5$. Here, the parameter $\rho$ controls the sparsity during the initial dictionary learning step \eqref{eq:dl}.  For the case when dictionary is selected from the voxels directly, we randomly select $15$ voxels from the target class-$16$ to form our dictionary.

\vspace{2pt}
\noindent\textbf{Dictionary selection for the Pavia University Dataset}: Here, for the learned dictionary case, we evaluate the performance of the aforementioned techniques for both entry-wise and column-wise settings for a dictionary of size $d=30$ for three values of the regularization parameter $\rho$, used for the initial dictionary learning step, i.e., $\rho = 0.01,~0.1$ and $0.5$. Further, we randomly select $60$ voxels from the target class-$5$, when the dictionary is formed from the data voxels.

\vspace{2pt}
\noindent\textbf{Comparison with matched filtering (MF)-based approaches}: In addition to the robust PCA-based and OP-based techniques introduced in Section~\ref{sec:contribution}, we also compare the performance of our techniques with two MF-based approaches. These MF-based techniques are agnostic to our model assumptions, i.e., entry-wise or column-wise sparsity cases. Therefore, the following description applies to both sparsity cases.

For the first MF-based technique, referred to as \textcolor{blue}{MF}, we form the inner-product of the column-normalized data matrix $\b{M}$, denoted as $\b{M}_n$,  with the dictionary $\b{D}$, i.e., $\b{D^\top M}_n$, and select the maximum absolute inner-product per column. For the second MF-based technique, \textcolor{blue}{MF$^\dagger$}, we perform matched filtering on the pseudo-inversed data \[\b{\widetilde{M} = D^\dagger M}\]. Here, the matched filtering corresponds to finding maximum absolute entry for each column of the column-normalized $\b{\widetilde{\b{M}}}$. Next, in both cases we scan through $1000$ threshold values between $(0, 1]$ to generate the results.

\begin{table*}[!htbp]
		 			\caption{Entry-wise sparsity model and Pavia University Dataset. Simulation results are presented for the proposed approach (\ref{Pe}), robust-PCA based approach on transformed data  (\ref{RPCA}), matched filtering (\textcolor{blue}{MF}) on original data $\b{M}$, and matched filtering on transformed data $\b{D^\dagger M}$ (\textcolor{blue}{MF$^\dagger$}), across dictionary elements $d$, and the regularization parameter for initial dictionary learning step $\rho$; See \eqref{eq:dl}. Threshold selects columns with column-norm greater than threshold such that AUC is maximized. For each case, the best performing metrics are reported in bold for readability. Further, $`` * "$ denotes the case where ROC curve was ``flipped'' (i.e. classifier output was inverted to achieve the best performance). }
		 			\vspace{-4pt}
		 			\label{res_tab_pu_ew}
		 			\captionsetup{justification=centering}
		 			\begin{subtable}{.5\linewidth}
		 				\captionsetup{font=footnotesize}
		 				\centering
		 				\caption{Learned dictionary, $d=30$}
		 				\vspace{-5pt}
		 				\scalebox{0.93}{
		 					\begin{tabular}{|P{0.5cm}|c|c|c|c|c|c|}
		 						\hline
		 						\multirow{2}{*}{\textbf{$d$}} & \multirow{2}{*}{$\rho$} & \multirow{2}{*}{\textbf{Method}}& \multirow{2}{*}{\textbf{Threshold }}&\multicolumn{2}{G|}{\textbf{Performance at best operating point} }& \multirow{2}{*}{\textbf{AUC}}\\ \cline{5-6}
		 						&&&&\textbf{TPR}&\textbf{FPR}&\\\hline
		 						\multirow{12}{*}{30} &	\multirow{4}{*}{0.01}
		 						&\textbf{D-RPCA(E)}					&0.150&\textbf{0.989}&\textbf{0.015}&\textbf{0.992}\\\cline{3-7}
		 						&&\textbf{RPCA$^\dagger$} &0.700&0.849&0.146&0.925\\\cline{3-7}
		 						&&\textbf{MF}						&N/A&0.929&0.073&0.962\\\cline{3-7}
		 						&&\textbf{MF$^\dagger$}	   &N/A&0.502&0.498&0.498\\\cline{2-7}
		 						&	\multirow{4}{*}{0.1}
		 						&\textbf{D-RPCA(E)}				    	&0.050&\textbf{0.982}&\textbf{0.019}&\textbf{0.992}\\\cline{3-7}
		 						&&\textbf{RPCA$^\dagger$} &3.000&0.638&0.374&0.664\\\cline{3-7}
		 						&&\textbf{MF}						&N/A&0.979&0.053&0.986\\\cline{3-7}
		 						&&\textbf{MF$^\dagger$}	   &N/A&0.620&0.381&0.660\\\cline{2-7}
		 						&	\multirow{4}{*}{0.5}
		 						&\textbf{D-RPCA(E)}					    &0.080&\textbf{0.982}&\textbf{0.019}&\textbf{0.992}\\\cline{3-7}
		 						&&\textbf{RPCA$^\dagger$} &2.500&0.635&0.381&0.671\\\cline{3-7}
		 						&&\textbf{MF}						&N/A&0.980&0.159&0.993\\\cline{3-7}
		 						&&\textbf{MF$_*^\dagger$}	   &N/A&0.555&0.447&0.442\\\hline
		 					\end{tabular}
		 					\label{dl_30:pu_ew}
		 				}
		 			\end{subtable} 
		 			\begin{subtable}{.5\linewidth}
		 				\captionsetup{font=footnotesize}
		 				\centering
		 				\caption{Dictionary by sampling voxels, $d=60$}
		 				\vspace{-5pt}
		 				\scalebox{0.93}{
		 					\begin{tabular}{|P{0.5cm}|c|c|c|c|c|}
		 						\hline 
		 						\multirow{2}{*}{\textbf{$d$}}  & \multirow{2}{*}{\textbf{Method}}& \multirow{2}{*}{\textbf{Threshold}}&\multicolumn{2}{G|}{\textbf{Performance at best operating point} }& \multirow{2}{*}{\textbf{AUC}}\\ \cline{4-5}
		 						&&&\textbf{TPR}&\textbf{FPR}&\\\hline
		 						\multirow{4}{*}{60}
		 						&\textbf{D-RPCA(E)}					&0.060&\textbf{0.986}&0.016&\textbf{0.995}\\\cline{2-6}
		 						&\textbf{RPCA$^\dagger$} &1.000&0.799&0.279&0.793\\\cline{2-6}
		 						&\textbf{MF}						&N/A&0.980&\textbf{0.011}&0.994\\\cline{2-6}
		 						&\textbf{MF$^\dagger$}	   &N/A&0.644&0.355&0.700\\\hline
		 					\end{tabular}
		 					\label{orig_dict:pu_ew}
		 				}
		 				\vspace{0.17cm}
		 				\caption{Average performance}
		 				\vspace{-3px}
		 				\scalebox{0.8}{	
		 					\begin{tabular}{|P{1.5cm}|c|c|c|c|c|c|}		
		 						\hline
		 						\multirow{2}{*}{\textbf{Method}}& \multicolumn{2}{G|}{\textbf{TPR}}& \multicolumn{2}{G|}{\textbf{FPR}}& \multicolumn{2}{G|}{\textbf{AUC}}\\ \cline{2-7}
		 						&\textbf{Mean}&\textbf{St.Dev.}&\textbf{Mean}&\textbf{St.Dev.}&\textbf{Mean}&\textbf{St.Dev.}\\ \hline
		 						\textbf{D-RPCA(E)}&	\textbf{0.984}&	\textbf{0.003}&	\textbf{0.014}&	\textbf{0.002}&	\textbf{0.993}&	\textbf{0.001}\\ \hline
		 						\textbf{RPCA$^\dagger$}&0.730&0.110&0.295&0.110&0.763&0.123\\ \hline
		 						\textbf{MF}&0.967&0.025&0.074&0.062&0.983&0.0149\\ \hline
		 						\textbf{MF$^\dagger$}	&0.580&0.064&0.420&0.065&0.575&0.125\\ \hline
		 					\end{tabular}
		 					\label{overall_perf:pu_ew}
		 				}
		 			\end{subtable}
		 			\captionsetup{font=small}
		 			\vspace{-15pt}
		 		\end{table*}
\vspace{2pt}
\noindent\textbf{Performance Metrics}: We evaluate the performance of these techniques via the receiver operating characteristic (ROC) plots. ROC plots are a staple for classification performance analysis of a binary classifier in machine learning; see also \cite{James2013}. Specifically, it is a plot between the true positive rate (TPR) and the false positive rate (FPR), where a higher TPR (close to $1$) and a lower FPR (close to $0$) indicates that the classifiier detects all the elements in the class while rejecting those outside the class. 

A natural metric to gauge good performance is the area under the curve (AUC) metric. It indicates the area under the ROC curve, which is maximized when TPR \[=1\] and FPR \[=0\], therefore, a higher AUC is preferred. Here, an AUC of $0.5$ indicates that the performance of the classifier is roughly as good as a coin flip on average. As a result, if a classifier has an AUC $<0.5$, one can improve the performance by simply inverting the result of the classifier. This effectively means that AUC is evaluated after ``flipping'' the ROC curve. In other words, this means that the classifier is good at rejecting the class of interest, and taking the complement of the classifier decision can be used to identify the class of interest.  

In our experiments, MF-based techniques often exhibit this phenomenon. Specifically, when the dictionary contains element(s) which resemble the average behavior of the spectral signatures, the inner-product between the normalized data columns and these dictionary elements may be higher as compared to other distinguishing dictionary elements. Since MF-based techniques rely on the maximum inner-product between the normalized data columns and the dictionary, and further since the spectral signatures of even distinct classes are highly correlated; see, for instance Fig.~\ref{fig:data_corr}, where MF-based approaches in these cases can effectively reject the class of interest. This leads to an AUC \[<0.5\]. Therefore, as discussed above, we invert the result of the classifier (indicated as $(\cdot)_*$ in the tables) to report the best performance. If using MF-based techniques, this issue can potentially be resolved in practice by removing the dictionary elements which tend to resemble the average behavior of the spectral signatures.

\vspace{2pt}
\noindent \emph{Parameter Setup for the Algorithms}

\noindent\textbf{Entry-wise sparsity case}: 
We evaluate and compare the performance of the proposed method \ref{Pe} with \ref{RPCA} (described in Section~\ref{sec:contribution}), \textcolor{blue}{MF}, and \textcolor{blue}{MF$^\dagger$}. Specifically, we evaluate the performance of these techniques via the receiver \edit{operating} characteristic (ROC) plot for the Indian Pines dataset and the Pavia University dataset, with the results shown in Table~\ref{res_tab}(a)-(d) and Table~\ref{res_tab_pu_ew}(a)-(c), respectively. 

For the proposed technique, we employ the accelerated proximal gradient (APG) algorithm shown in Algorithm~\ref{algo} and discussed in Section~\ref{sec:optimization} to solve the optimization problem shown in \ref{Pe}. Similarly, for \ref{RPCA} we employ the APG algorithm with transformed data matrix $\widetilde{\b{M}}$, while setting $\b{D = I}$. 

With reference to selection of tuning parameters for the APG solver for \eqref{Pe} (\ref{RPCA}, respectively), we choose $v = 0.95$, $\nu =\|\mb{M}\|$ ($\nu =\|\b{\widetilde{M}}\|$), $\bar{\nu} = 10^{-4}$, and scan through $100$ values of $\lambda_e$ in the range $\lambda_e \in (0, {\|\b{D^\top M}\|_{\infty}}/{\|\b{M}\|} ]$ ($\lambda_e \in (0, {\|\b{\widetilde{M}}{\|_{\infty}}/{\|\b{\widetilde{M}}\|} }]$), to generate the ROCs. 
We threshold the resulting estimate of the sparse part $\b{S} \in \RR^{d \times m}$ based on its column norm. We choose the threshold such that the AUC metric is maximized for both cases (\ref{Pe} and \ref{RPCA}). 

\vspace{2pt}
\noindent\textbf{Column-wise sparsity case}: For this case, we evaluate and compare the performance of the proposed method \ref{Pc} with \ref{OP} (as described in Section~\ref{sec:contribution}), \textcolor{blue}{MF}, and \textcolor{blue}{MF$^\dagger$}. The results for the Indian Pines dataset and the Pavia University dataset as shown in Table~\ref{res_tab_ip_cw}(a)-(d) and Table~\ref{res_tab_pu_cw}(a)-(c), respectively. As in the entry-wise sparsity case, we employ the accelerated proximal gradient (APG) algorithm presented in Algorithm~\ref{algo} to solve the optimization problem shown in \ref{Pc}. Similarly, for \ref{OP} we employ the APG with transformed data matrix $\widetilde{\b{M}}$, while setting $\b{D = I}$. For the tuning parameters for the APG solver for \eqref{Pc} (\ref{OP}, respectively), we choose $v = 0.95$, $\nu =\|\mb{M}\|$ ($\nu =\|\b{\widetilde{M}}\|$), $\bar{\nu} = 10^{-4}$, and scan through $100$ $\lambda_c$s in the range $\lambda_c \in (0, {\|\b{D^\top M}\|_{\infty,2}}/{\|\b{M}\|} ]$ ($\lambda_c \in (0, {\|\b{\widetilde{M}}{\|_{\infty,2}}/{\|\b{\widetilde{M}}\|} }]$), to generate the ROCs. We threshold the resulting estimate of the sparse part $\b{S} \in \RR^{d \times m}$ based on its column norm.

 \captionsetup{justification=justified}	

			\begin{table*}[!bhtp]
				\caption{Column-wise sparsity model and Pavia University Dataset. Simulation results for the proposed approach (\ref{Pc}), Outlier Pursuit (OP) based approach (\ref{OP}), matched filtering (\textcolor{blue}{MF}) on original data $\b{M}$, and matched filtering on transformed data $\b{D^\dagger M}$ (\textcolor{blue}{MF$^\dagger$}), across dictionary elements $d$, and the regularization parameter for initial dictionary learning step $\rho$; See \eqref{eq:dl}. Threshold selects columns with column-norm greater than threshold such that AUC is maximized. For each case, the best performing metrics are reported in bold for readability.  Further, $`` * "$ denotes the case where ROC curve was ``flipped'' (i.e. classifier output was inverted to achieve the best performance). }
				\vspace{-4pt}
				\label{res_tab_pu_cw}
				\captionsetup{justification=centering}
				\begin{subtable}{.5\linewidth}
					\captionsetup{font=footnotesize}
					\centering
					\caption{Learned dictionary, $d=30$}
					\vspace{-5pt}
					\scalebox{0.9}{
						\begin{tabular}{|P{0.5cm}|c|c|c|c|c|c|}
							\hline
							\multirow{2}{*}{\textbf{$d$}} & \multirow{2}{*}{$\rho$} & \multirow{2}{*}{\textbf{Method}}& \multirow{2}{*}{\textbf{Threshold }}&\multicolumn{2}{G|}{\textbf{Performance at best operating point} }& \multirow{2}{*}{\textbf{AUC}}\\ \cline{5-6}
							&&&&\textbf{TPR}&\textbf{FPR}&\\\hline
							\multirow{12}{*}{30} &	\multirow{4}{*}{0.01}
							&\textbf{D-RPCA(C)}					&0.065&\textbf{0.990}&\textbf{0.015}&\textbf{0.991}\\\cline{3-7}
							&&\textbf{OP$^\dagger$} &0.800&0.7581&0.3473&0.705\\\cline{3-7}
							&&\textbf{MF}						&N/A&0.929&0.073&0.962\\\cline{3-7}
							&&\textbf{MF$^\dagger$}	   &N/A&0.502&0.50&0.498\\\cline{2-7}
							&	\multirow{4}{*}{0.1}
							&\textbf{D-RPCA(C)}				    	&0.070&\textbf{0.996}&\textbf{0.022}&\textbf{0.994}\\\cline{3-7}
							&&\textbf{OP$^\dagger$} &0.100&0.989&0.3312&0.904\\\cline{3-7}
							&&\textbf{MF}						&N/A&0.979&0.053&0.986\\\cline{3-7}
							&&\textbf{MF$^\dagger$}	   &N/A&0.62&0.3814&0.66\\\cline{2-7}
							&	\multirow{4}{*}{0.5}
							&\textbf{D-RPCA(C)}					    &0.035&\textbf{0.983}&\textbf{0.017}&\textbf{0.995}\\\cline{3-7}
							&&\textbf{OP$^\dagger$} &0.200&0.940&0.264&0.887\\\cline{3-7}
							&&\textbf{MF}						&N/A&0.980&0.160&0.993\\\cline{3-7}
							&&\textbf{MF$_*^\dagger$}	   &N/A&0.555&0.447&0.442\\\hline
						\end{tabular}
						\label{dl_30:pu_cw}
					}
				\end{subtable} 
				\begin{subtable}{.5\linewidth}
					\captionsetup{font=footnotesize}
					\centering
					\caption{Dictionary by sampling voxels, $d=60$ }
					\vspace{-5pt}
					\scalebox{0.9}{
						\begin{tabular}{|P{0.5cm}|c|c|c|c|c|}
							\hline 
							\multirow{2}{*}{\textbf{$d$}}  & \multirow{2}{*}{\textbf{Method}}& \multirow{2}{*}{\textbf{Threshold}}&\multicolumn{2}{G|}{\textbf{Performance at best operating point} }& \multirow{2}{*}{\textbf{AUC}}\\ \cline{4-5}
							&&&TPR&FPR&\\\hline
							\multirow{4}{*}{60}
							&\textbf{D-RPCA(C)}					&0.020&\textbf{0.993}&0.022&\textbf{0.994}\\\cline{2-6}
							&\textbf{OP$^\dagger$} &0.250&0.963&0.264&0.907\\\cline{2-6}
								&\textbf{MF}						&N/A&0.980&\textbf{0.011}&0.994\\\cline{2-6}
								&\textbf{MF$^\dagger$}	   &N/A&0.644&0.355&0.700\\\hline
						\end{tabular}
						\label{orig_dict:pu_cw}
					}
					\vspace{0.17cm}
					\caption{Average performance}
					\vspace{-3px}
					\scalebox{0.88}{	
						\begin{tabular}{|P{1.5cm}|c|c|c|c|c|c|}		
							\hline
							\multirow{2}{*}{\textbf{Method}}& \multicolumn{2}{G|}{\textbf{TPR}}& \multicolumn{2}{G|}{\textbf{FPR}}& \multicolumn{2}{G|}{\textbf{AUC}}\\ \cline{2-7}
							&Mean&St.Dev.&Mean&St.Dev.&Mean&St.Dev.\\ \hline
							\textbf{D-RPCA(C)}&	\textbf{0.990}&	\textbf{0.006}&	\textbf{0.015}&	\textbf{0.003}&	\textbf{0.993}&	\textbf{0.002}\\ \hline
							\textbf{OP$^\dagger$}&0.912&0.105&0.302&0.044&0.850&0.098\\ \hline
							\textbf{MF}&0.97&0.025&0.074&0.063&0.984&0.015\\ \hline
							\textbf{MF$^\dagger$}	&0.580&0.064&0.4208&0.065&0.575&0.124\\ \hline
						\end{tabular}
						\label{overall_perf:pu_cw}
					}
				\end{subtable}
				\captionsetup{font=small}
				\vspace*{-15pt}
			\end{table*}
\vspace{3pt}
\noindent\emph{Analysis:} Table~\ref{res_tab}--\ref{res_tab_pu_ew} and Table~\ref{res_tab_ip_cw}--\ref{res_tab_pu_cw} show the ROC characteristics and the classification performance of the proposed techniques \ref{Pe} and \ref{Pc}, for two datasets under consideration, respectively, under various choices of the dictionary $\b{D}$ and regularization parameter $\rho$ for \eqref{eq:dl}. We note that both proposed techniques \ref{Pe} and \ref{Pc} on an average outperform the competing techniques, emerging as the most reliable techniques across different dictionary choices; see Tables~\ref{res_tab}(d), \ref{res_tab_pu_ew}(c), \ref{res_tab_ip_cw}(d), and \ref{res_tab_pu_cw}(c). 

Further, the performance of \ref{Pc} is slightly better than \ref{Pe}. This can be attributed to the fact that the column-wise sparsity model does not require the columns of \[\b{S}\] to be sparse themselves. As alluded to in Section~\ref{sec:contribution}, this allows for higher flexibility in the choice of the dictionary elements for the thin dictionary case.

In addition, we see that the matched filtering-based techniques (and even \ref{OP} based technique for $d=4 $ and $\rho=0.1$ in Table~\ref{res_tab_ip_cw}) exhibit ``flip'' or inversion of the ROC curve. As described in Section~\ref{sec:exp_setup}, this phenomenon is an indicator that a classifier is better at rejecting the target class. In case of MF-based technique, this is a result of a dictionary that contains an element that resembles the average behavior of the spectral responses. A similar phenomenon is at play in case of the \ref{OP} for $d=4 $ and $\rho=0.1$ in Table~\ref{res_tab_ip_cw}. Specifically, here the inversion indicates that the dictionary is capable of representing the columns of the data \[\b{M}\] effectively, which leads to an increase in the corresponding column norms in their representation \[\widehat{\b{M}}\]. Coupled with the fact that the component \[\b{L}\] is no longer low-rank for this thin dictionary case (see our discussion in Section~\ref{sec:contribution}), this results in rejection of the target class. On the other hand, our techniques \ref{Pe} and \ref{Pc} do not suffer from this issue. Moreover, note that across all the experiments, the thresholds for \ref{RPCA} and \ref{OP} are higher than their D-RPCA counterparts. This can also be attributed to the pre-multiplication by the pseudo-inverse of the dictionary \[\b{D}^\dagger\], which increases column norms based on the leading singular values of \[\b{D}\]. Therefore, using \ref{Pe}, when the target spectral response admits a sparse representation, and \ref{Pc}, otherwise, yield consistent and superior results as compared to related techniques.

\begin{figure}[!t]
  \centering
  \begin{tabular}{cP{0.01cm}cc}
  {\footnotesize \textbf{Data}}&&{\footnotesize$\b{L}$ }& {\footnotesize $\b{DS}$} \vspace{-2pt}\\
    \epsfig{file=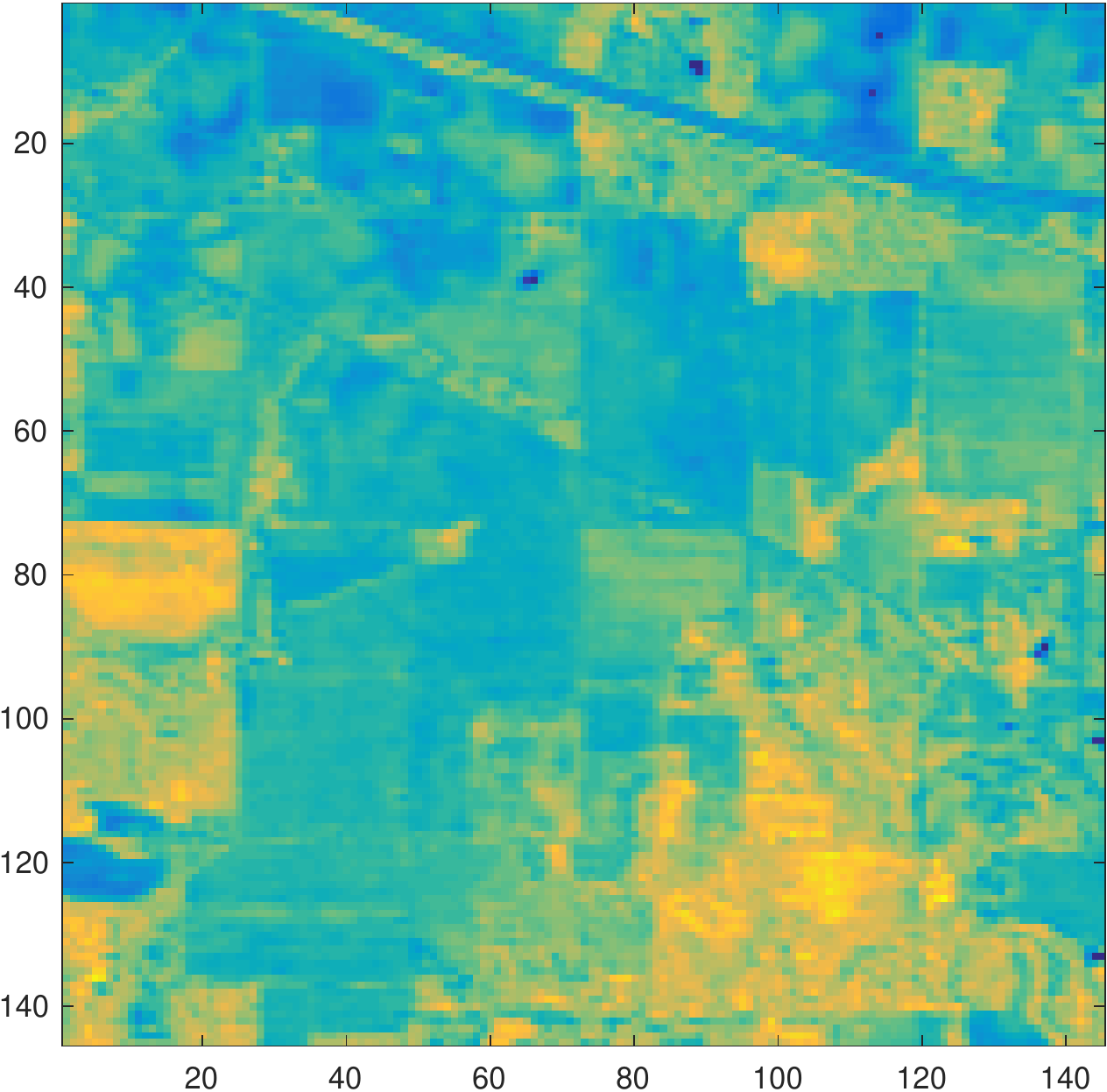,width=0.27\linewidth,clip=}&
     {\small\rotatebox{90}{ ~~~~~~~~~{Best $\lambda_e$}}} & 
     \epsfig{file=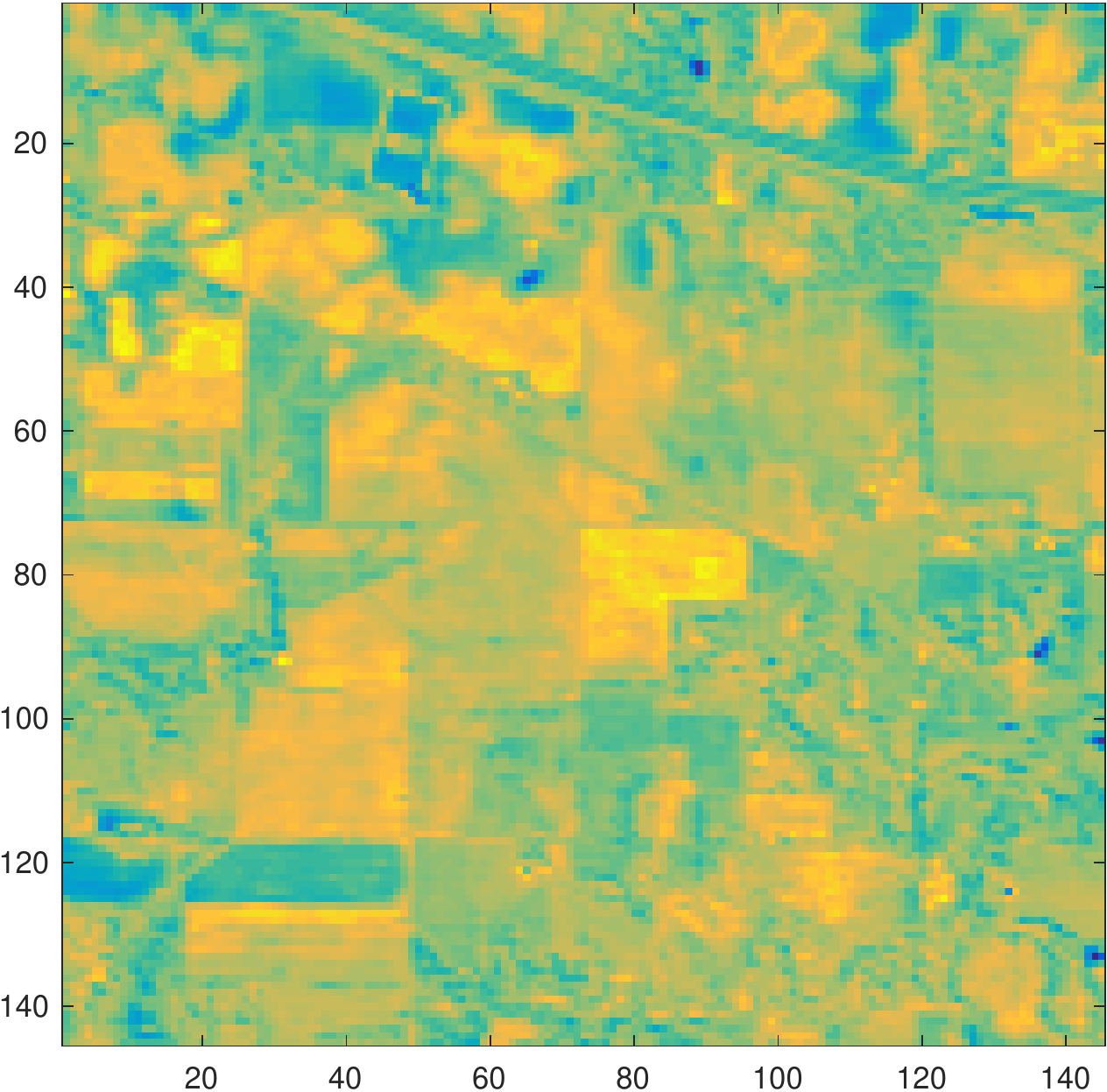,width=0.27\linewidth,clip=}&
     \epsfig{file=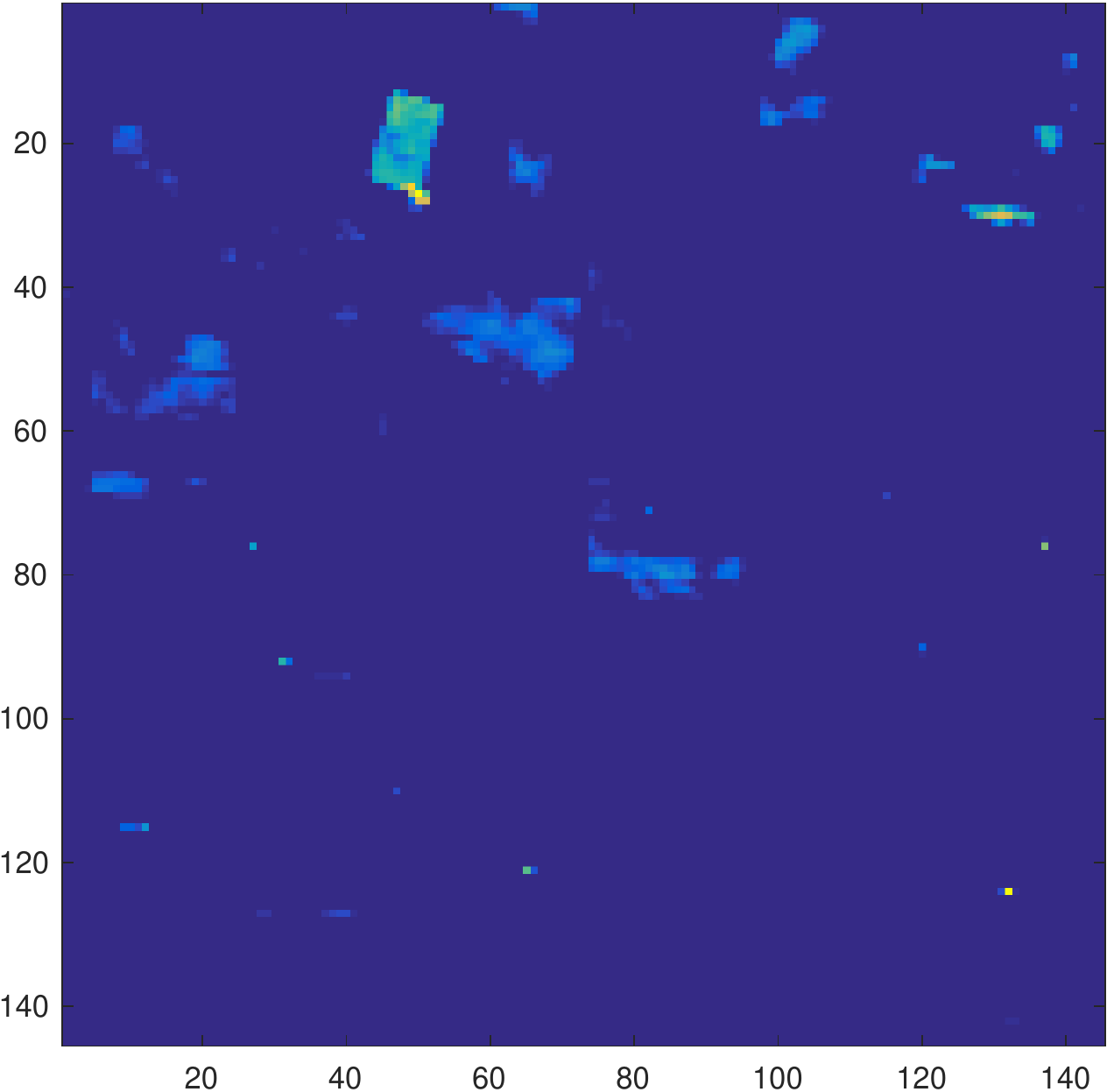,width=0.27\linewidth,clip=} \vspace{-5pt}\\ 
         (a) &&(c) & (d) \\
     \epsfig{file=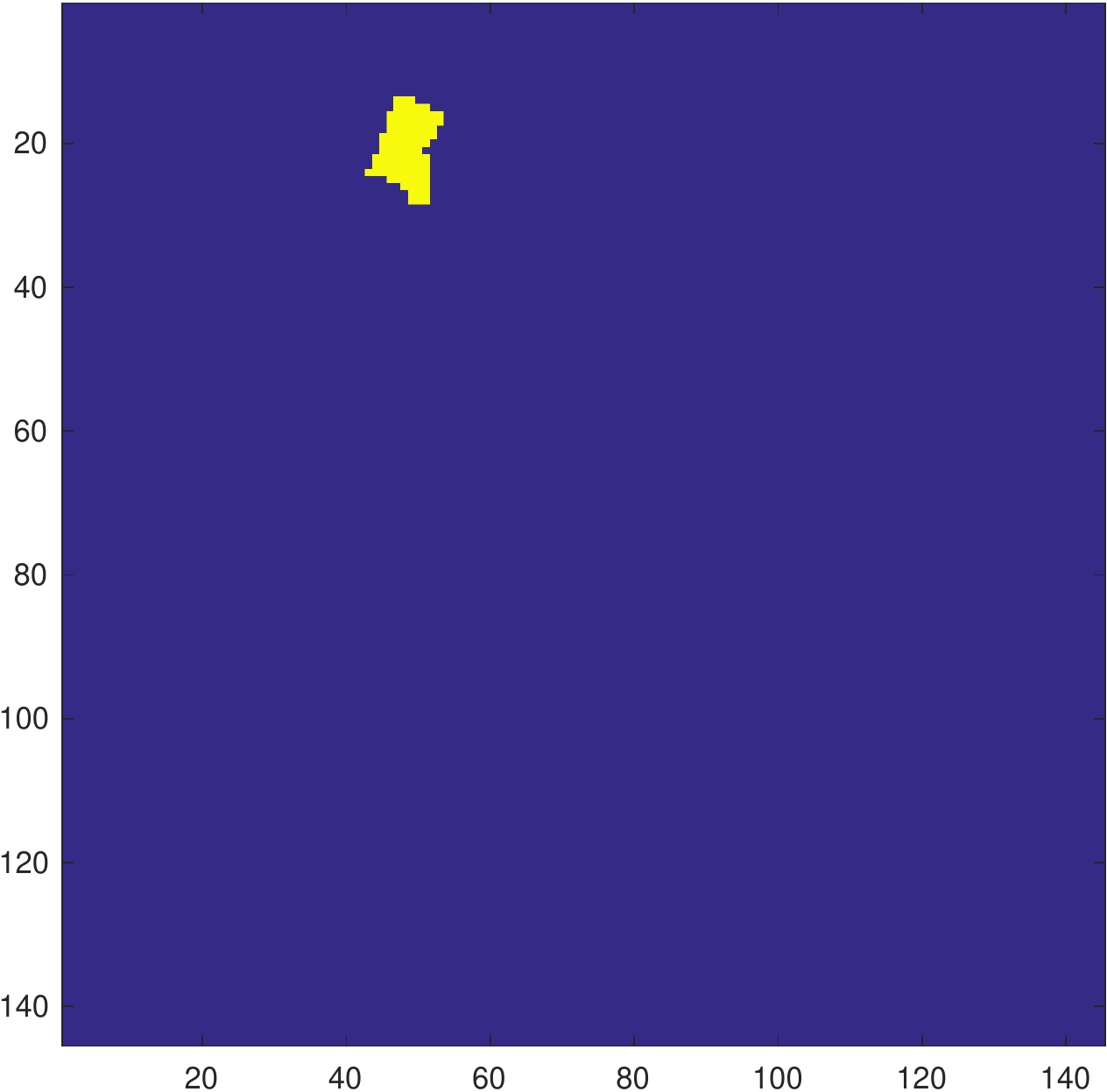,width=0.27\linewidth,clip=}&
     {\small\rotatebox{90}{ ~~~~{$85\%$ of $\lambda^{\max}_e$}}} &
     \epsfig{file=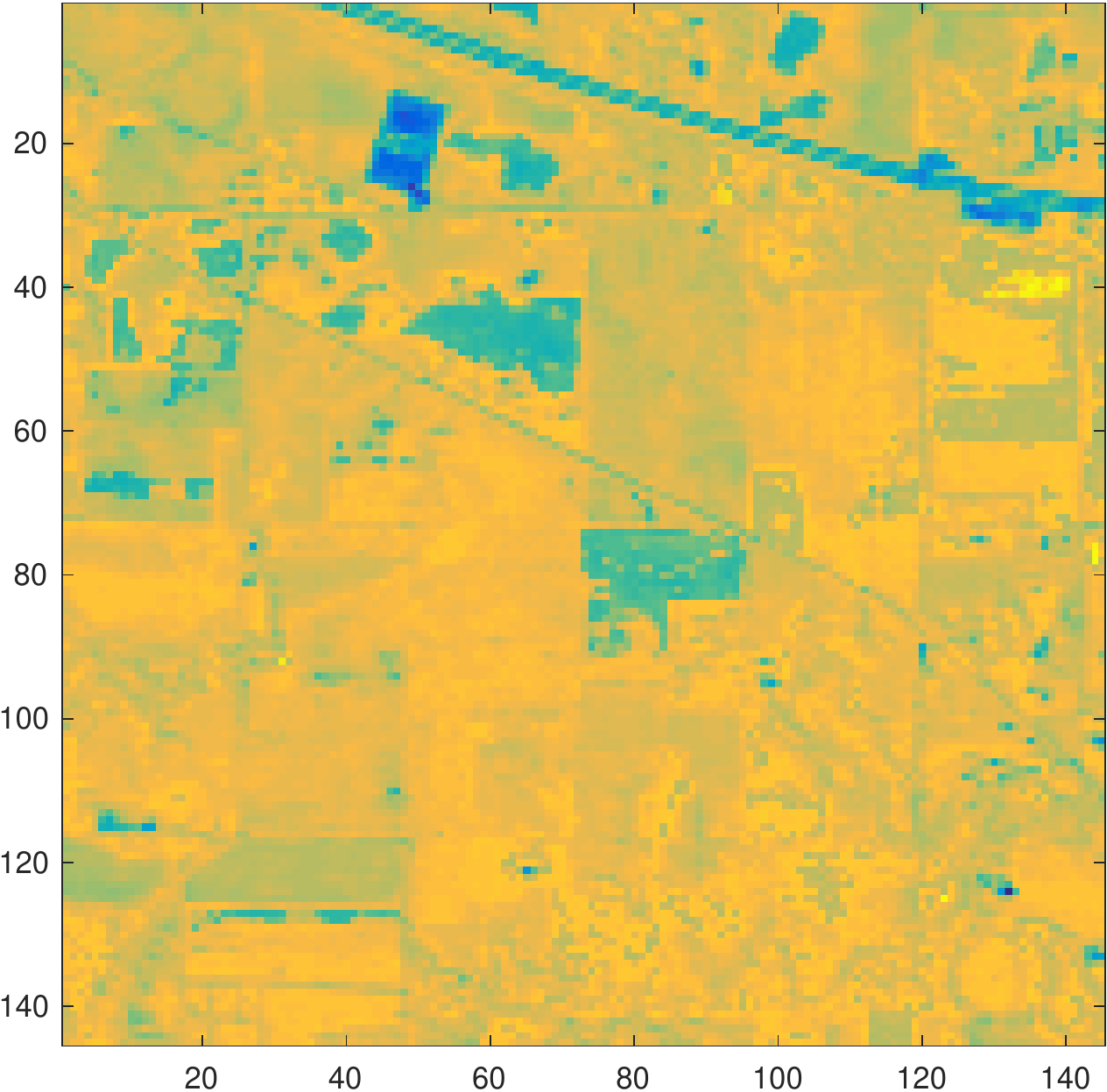,width=0.27\linewidth,clip=} &
     \epsfig{file=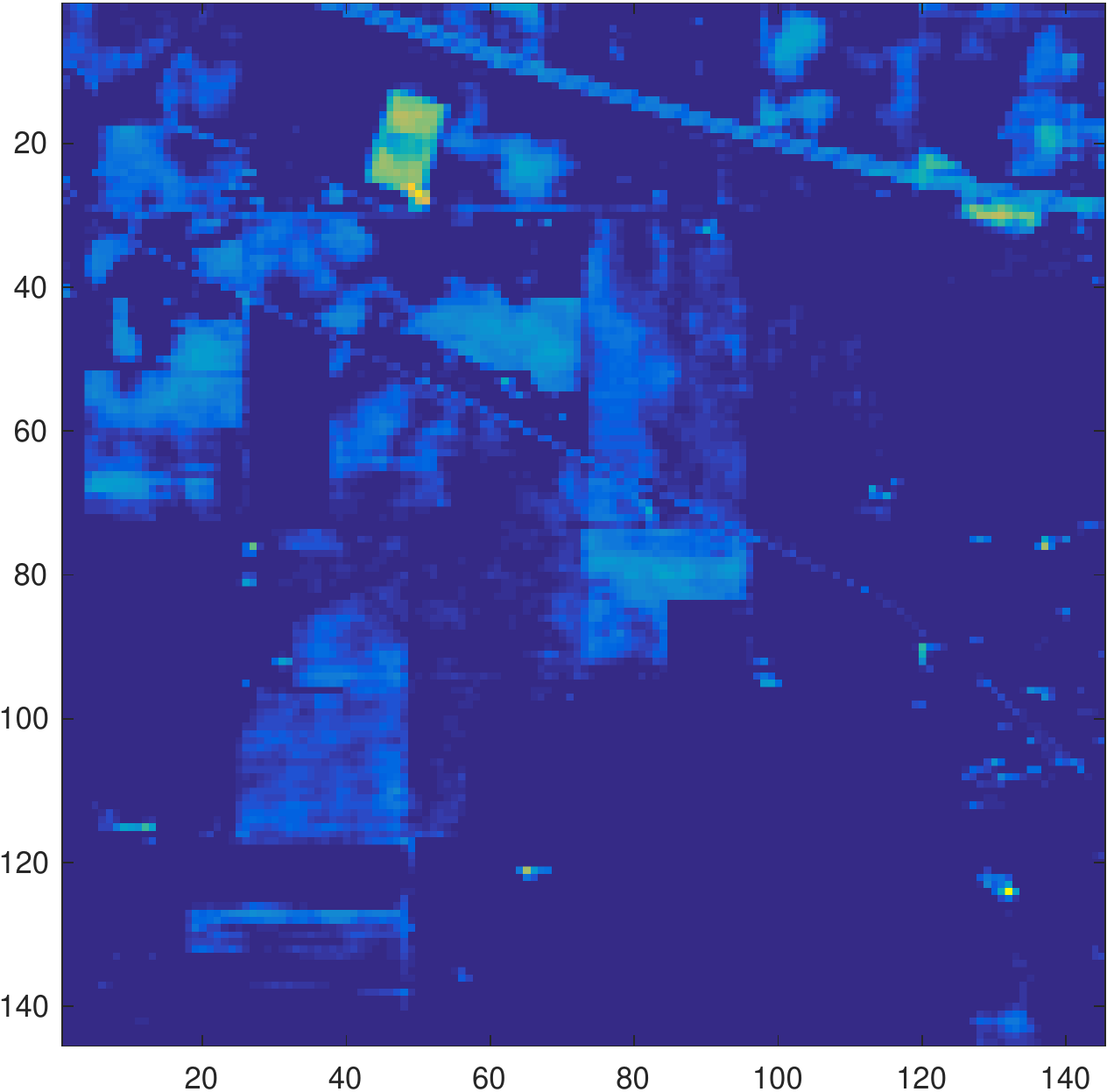,width=0.27\linewidth,clip=} \vspace{-5pt}\\ 
      (b)  && (e) & (f)
  \end{tabular}
  \vspace{-4pt}
  \caption{Recovery of the low-rank component \[\b{L}\] and the dictionary sparse component \[\b{DS}\] for different values of $\lambda$ for the proposed technique at { $n=50$-th} channel of the \cite{HSdat} (shown in panel (a)) corresponding to the results shown in Table~\ref{res_tab}(c). Panel (b) corresponds to the ground truth for class-$16$. Panel (c) and (d) show the recovery of the low-rank part and dictionary sparse part for a $\lambda$ at the best operating point. While, panels (e) and (f) show the recovery of these components at $\lambda_e = 85\%$ of $\lambda^{\max}_e$. Here, $\lambda^{\max}_e$ denotes the maximum value $\lambda_e$ can take; see Section~\ref{sec:opt_ew_param}.}\vspace*{-5pt}
  \label{figure:res_our}
  \end{figure}

There are other interesting recovery results which warrant our attention. Fig.~\ref{figure:res_our} shows the low-rank and the dictionary sparse component recovered by \ref{Pe} for two different values of $\lambda_e$, for the case where we form the dictionary by randomly sampling the voxels (Table~\ref{res_tab}(c)) for the Indian Pines Dataset \cite{HSdat}.  Interestingly, we recover the rail tracks/roads running diagonally on the top-right corner, along with some low-density housing; see Fig~\ref{figure:res_our} (f). This is because the \textit{signatures} we seek (stone-steel) are similar to the signatures of the materials used in these structures. This further corroborates the applicability of the proposed approach in detecting the presence of particular spectral \textit{signatures} as long as they are appropriately distinct. 
 \vspace{-2pt}
\section{Discussion}
\label{sec:conclusion}
\vspace{-3pt}
We analyze a dictionary-based generalization of Robust PCA, and use it for target localization in a hyperspectral (HS) image from the \textit{a priori} known spectral \textit{signature} of the material of interest. Here, we consider a case where the acquired data can be modeled as a superposition of a low-rank component and a dictionary sparse component, and analyze this model under two distinct sparsity modalities -- entry-wise and column-wise, respectively for both \emph{thin} and \emph{fat} dictionary cases.  

Our analysis shows that contrary to the existing intuition, in the \emph{thin} dictionary case, premultiplication with pseudo-inverse of the dictionary may not reduce the problem to that of Robust PCA. To this end, we theoretically analyze the \emph{thin} dictionary case while extending the analysis for the \emph{fat} dictionary case, while also analyzing the column-wise sparsity case. As a result, our results, to the best of our knowledge, are the most general for this model and facilitate use of this model for practical settings. Here, we consider the worst case analysis for the deterministic setting. Therefore, analysis of this model with additional randomness assumptions on the constituent factors constitutes the future work. Additionally, the recent results on non-convex low-rank matrix estimation formulations \cite{tu2015low,chen2015fast} may potentially lead to computationally efficient algorithms by replacing the expensive SVD step. 

In this work, we also leverage our theoretical results for a target localization task in hyperspectral imaging to demonstrate the applicability of the proposed approach on real-world demixing tasks. Here, we show how the entry-wise and column-wise sparsity modalities can be used to detect targets depending on the dictionary structure. Future work on this thread will aim to further exploit local similarities (potentially by group sparsity constraints) in HS images to improve localization. 

Overall, our algorithm agnostic theoretical guarantees and analysis of the corresponding application in HS image target detection task using the proposed dictionary-based generalization of Robust PCA opens up future theory-backed explorations of the model in various target detection applications.

\vspace*{-5pt}
\bibliographystyle{IEEEbib}
\bibliography{referLR_2,referLR_app}
 \setcounter{equation}{31}
 \appendix \label{sec:app}
 In the following appendices, we provide the proofs of the lemmata employed to establish our main results. We also summarize the notation in Table~\ref{tab:notation}.
 \begin{table}[h]
 \caption{Summary of important notation and parameters}\vspace{-5pt}
 \scalebox{0.92}{
 \begin{tabular}{|c|p{7cm}|}
 	\hline
 	\multicolumn{2}{|p{7cm}|}{\textbf{Matrices}} \\ \hline
 	\[\mathbf{M} \in \mathbb{R}^{n \times m}\] & The data matrix\\
 	\[\mathbf{L} \in \mathbb{R}^{n \times m}\] & The low-rank matrix with rank-\[r\] and singular value decomposition \[\mathbf{L} = \mathbf{U\Sigma V}^\top\]\\
 	\[\mathbf{D} \in \mathbb{R}^{n \times d}\] & The known dictionary either \textit{thin} (\[d \leq n\]) or \textit{fat} (\[d > n\])\\
 	\[\mathbf{S} \in \mathbb{R}^{d \times m}\] & The sparse component with the following properties --(1) in case of entry-wise sparsity: \[s_e\] non-zero entries and  when \[d>n\] has at most \[k\] non-zeros per column, and (2) in case of column-wise sparsity: \[s_c\] non-zero columns \\ \hline
 	\multicolumn{2}{|p{8cm}|}{\textbf{Regularization Parameters}} \\ \hline
 	\[\lambda_e \in \mathbb{R}\] & The regularization parameter for the entry-wise sparsity case\\
 	\[\lambda_c \in \mathbb{R}\] & The regularization parameter for  the column sparsity case\\ \hline
 	\multicolumn{2}{|p{8cm}|}{\textbf{Subspaces}} \\ \hline
 	\[\mathcal{L}\] & The set of matrices which span the same column or row space as \[\mathbf{L}\], i.e., \[\mathcal{L} := \{ \mathbf{UW}^\top_1 + \b{W}_2\b{V}^\top, \b{W}_1 \in \mathbb{R}^{m \times r}, \b{W}_2 \in \mathbb{R}^{n \times r} \] for \[\b{W}_1 \neq 0\] or \[\b{W}_2 \neq 0 \}\].\\
 	\[\mathcal{S}_e\] & The set of matrices with the same support as \[\mathbf{S}\] (for the entry-wise sparse case).\\
 	\[\mathcal{S}_c\] & The set of matrices with the same column support as \[\mathbf{S}\] (for the column-wise sparse case).\\
 	\[\mathcal{D}\] & The set of matrices whose columns span the subspace spanned by columns of \[\mathbf{D}\], i.e. \[	\mathcal{D} := \{ \b{Z} = \b{RH}, \b{H} \in \mathcal{S}_e ~\text{or}~ \b{H} \in \mathcal{S}_c \}\]\\ 
 	\[\c{U}\]& The column space of \[\b{L}\]\\
 	\[\c{V}\]& The row space of \[\b{L}\]\\\hline
 	\multicolumn{2}{|p{8cm}|}{\textbf{Index Sets}} \\ \hline
 	\[\mathcal{I}_{\c{S}_e}\] & Support of matrix \[\@S\@\] (entry-wise case)\\
 	\[\mathcal{I}_{\c{S}_c}\] & Column support of matrix \[\@S\@\] (the outliers)\\
 	\[\c{I}_{\b{L}}\] & Index set of the inliers (column-wise case)\\ \hline
 	\multicolumn{2}{|p{8cm}|}{\textbf{Projection}} \\ \hline
 	\[\mathcal{P}_{\mathcal{G}}(\cdot)\]&  Projection operator corresponding to any subspace \[\mathcal{G}\]\\
 	\[\mathbf{P}_{\mathbf{G}}\]&  Projection matrix corresponding to the operator \[\mathcal{P}_{\mathcal{G}}(\cdot)\]\\\hline
 	\multicolumn{2}{|p{8cm}|}{\textbf{Parameters for analysis}} \\ \hline
 	\[\mu\] & The incoherence parameter between the low-rank component and the dictionary, defined as \[	\mu := \underset{\b{Z} \in \mathcal{D} \backslash \{\b{0}_{d \times m}\}}{\max} \tfrac{\|\mathcal{P}_{\mathcal{L}}(\b{Z})\|_{\rm F}}{\|\b{Z}\|_{\rm F}}\] \\
 	\[\gamma_{\b{V}}\] & Defined as \[\gamma_{{\b{V}}} := \underset{i}{\max} \|\mathbf{P}_{\b{V}}\mathbf{e}_{i}\|^2\] \\
 	\[\gamma_{\mathbf{U}}\] & Defined as \[\gamma_{\b{U}}  :=  \underset{i}{\max} \tfrac{\|\b{P}_\b{U} \b{D}\b{e}_{i}\|^2}{\|\b{De}_{i}\|^2}\] \\
 	\[\beta_{\mathbf{U}}\] & Defined as \[\b{\beta}_{\b{U}}  :=  \underset{\|\b{u}\| = 1}{\max} \tfrac{\|(\b{I} - \b{P}_{\b{U}}) \b{D}\b{u}\|^2}{\|\b{Du}\|^2}\] \\
 	\[\xi_e\] & Defined as \[\xi_e := \|\b{D}^\top \b{UV}^\top\|_{\infty}\] \\
 	\[\xi_c\] & Defined as \[\xi_c := \|\b{D}^\top \b{UV}^\top\|_{\infty,2}\] \\
 	\[\alpha_\ell\] & Lower generalized frame bound \\
 	\[\alpha_u\] & Upper generalized frame bound\\
 	\hline
 \end{tabular}}
 \label{tab:notation}
 \vspace{-9pt}
 \end{table}
 \section{Proofs of Intermediate results} 
 \vspace{-10pt}
 \subsection{Proofs for Entry-wise Case}
 \label{app:entry}
 We present the details of the proofs in this section for the entry-wise case. We first start by deriving the optimality conditions. 
 
 \vspace{5pt}
 \begin{proof}[\hspace{-23pt} Proof of Lemma~\ref{DualCert}]
 	Let \[\{\b{L}_0, \b{S}_0\}\] be a solution of the problem posed above. Notice that this pair is not necessarily unique. For example, as shown in proof of Lemma 2 in \cite{Mardani2012}, \[\{\b{L}_0 + \b{DH}, \b{S}_0 - \b{H} \}\], with arbitrary \[\b{H}\], is another feasible solution of the problem satisfying the optimality conditions (derived in this section).
 	
 	We begin by writing the Lagrangian, \[\c{F}(\b{L}, \b{S}, \b{\Lambda})\], for the given problem as follows. 
 	\begin{align*}
 	\c{F}(\b{L}, \b{S}, \b{\Lambda}) =\|\b{L}\|_* + \lambda_e \|\b{S}\|_1 ~+ \langle\b{\Lambda}, ~\b{M} - \b{L} - \b{DS}\rangle,  		
 	\end{align*}
 	where \[\b{\Lambda} \in \mathbb{R}^{n \times m}\] are the Lagrange multipliers. 
 	
 	Let the singular value decomposition (SVD) of  \[\b{L}_0\]  be represented as \[\b{U\Sigma V^\top}\]. Then the sub-differential set of \[\|\b{L}\|_*\] can be represented as
 	\begin{align*}
 	\partial_\b{L} \|\b{L}\|_* \Bigr|_{\b{L} = \b{L}_0} = \{ \b{UV^\top} + \b{W} : \|\b{W}\| \leq 1, \c{P}_{\c{L}} (\b{W}) = \b{0} \},  		
 	\end{align*}
 	as shown in \cite{Watson1992}. Also, the subdifferential set corresponding to \[\|\b{S}\|_1\] is given by 
 	\begin{align*}
 	\partial_\b{S} \|\b{S}\|_1 \Bigr|_{\b{S} = \b{S}_0}  = \{ \text{sign}(\b{S}_0) + \b{F} : \|\b{F}\|_{\infty} \leq 1, \c{P}_{\c{S}_e} (\b{F}) = \b{0} \},  		
 	\end{align*}
 	Using these results, we write the sub-differential of the Lagrangian with respect to \[\b{L}\] and \[\b{S}\] at \[\{\b{L}_0, \b{S}_0\}\] as
 	\begin{align*}
 	\partial_\b{L} \c{F}(\b{L}_0, \b{S}_0, \b{\Lambda}) &=  \resizebox{0.32\textwidth}{0.015\textwidth}{$ \{ \b{UV^\top} + \b{W} - \b{\Lambda}: \|\b{W}\| \leq 1, \c{P}_{\c{L}} (\b{W}) = \b{0} \} $},  \\
 	\partial_\b{S} \c{F}(\b{L}_0, \b{S}_0, \b{\Lambda}) &= \{ \b{\lambda_e}\text{sign}(\b{S}_0) + \b{\lambda_e}\b{F} - \b{D^\top\Lambda}, \|\b{F}\|_{\infty}\leq 1, \\
 	&\hspace{1in}\c{P}_{\c{S}_e} (\b{F}) = \b{0} \}.  		
 	\end{align*}
 	Then optimality conditions are 
 	\begin{align*}
 	\b{0}_{n \times m } \in \partial_\b{L} \c{F}(\b{L}_0, \b{S}_0, \b{\Lambda}) \text{~and~}  \b{0}_{d \times m } \in \partial_\b{S} \c{F}(\b{L}_0, \b{S}_0, \b{\Lambda}), 
 	\end{align*}
 	which implies that the dual solution \[\b{\Lambda}\] must obey the following,
 	\begin{align*}
 	\b{\Lambda} &\in \b{UV^\top} + \b{W}, ~\|\b{W}\| \leq 1, ~\c{P}_{\c{L}} (\b{W}) = \b{0}_{n \times m } \text{~and~}\\
 	\b{D^\top\Lambda} &\in \b{\lambda_e}\text{sign}(\b{S}_0) + \b{\lambda_e}\b{F}, ~\|\b{F}\|_{\infty} \leq 1, ~\c{P}_{\c{S}_e} (\b{F}) = \b{0}_{d \times m }.
 	\end{align*}
 	Our aim here is to find the conditions on \[\b{W}\] and \[\b{F}\] such that the pair \[\{ \b{L}_0,~\b{S}_0\}\] is a unique solution to the problem at hand.

 	
 	Using these conditions, we see that \[\c{P}_{\c{L}}(\b{\Lambda}) = \b{UV^\top}\] and \[\c{P}_{\c{S}_e}(\b{D^\top\Lambda}) = \lambda_e\text{sign}(\b{S}_0)\]; these correspond to conditions \textbf{(C1)} and \textbf{(C2)}, respectively. Now consider a feasible solution \[\{\b{L_0 + DH}, \b{S_0 - H}\}\] for a non-zero \[\b{H} \in \mathbb{R}^{d\times m} \].  
     Now by duality of norms
 		\begin{align*}
 	 	\|\c{P}_{\c{L}^\perp}(\b{DH})\|_* =  \underset{\|\tilde{\b{W}} \| \leq 1}{\rm sup}	\langle\tilde{\b{W}} ,\c{P}_{\c{L}^\perp}(\b{DH})\rangle 
 	 	 	\end{align*}
 	 We can choose \[\b{W} := \c{P}_{\c{L}^\perp}( \tilde{\b{W}} )\] which implies	\[\|\b{W}\| \leq 1\] and \[\c{P}_{\c{L}} (\b{W}) = \b{0}\] and 
 	 	 	 		\begin{align*}
 	 	 	 	 	\langle\b{W},~\b{DH}\rangle ~=~ \|\c{P}_{\c{L}^\perp}(\b{DH})\|_*.
 	 	 	 	 	\end{align*}
 	Further, let \[\b{F}\], with \[\|\b{F}\|_\infty = 1\] and \[\c{P}_{\c{S}_e} (\b{F}) = \b{0}\], be such that
 	\begin{align*}
 	\b{F}_{ij} = \begin{cases}
 	-\text{sign}(\b{H}_{ij})&, \text{ if } \{i, ~j\} \not\in \c{S}_e \text{ and } \b{H}_{ij} \neq 0\\
 	~~0&, \text{ otherwise}
 	\end{cases},
 	\end{align*}
 	where \[\b{F}_{ij}\] denotes the \[(i,j)^{\text{th}}\] element of \[\b{F}\]. Then, we arrive at the following simplification for \[\langle\b{F},~\b{H}\rangle\] by duality of norms,
 	\begin{align*}
 	\langle\b{F},~\b{H}\rangle ~=~ \langle\b{F},~\c{P}_{\c{S}_e^\perp}(\b{H}) \rangle ~=~ -\|\c{P}_{\c{S}_e^\perp}(\b{H})\|_1.
 	\end{align*}
 	%
 	%
 	We first write the sub-gradient optimality condition,
 	\begin{align}
 	\label{subopt}
 	&\|\b{L}_0 + \b{DH}\|_* +\lambda_e~\|\b{S_0 - H}\|_1 \notag \geq~\|\b{L}_0\|_* + \lambda_e~\|\b{S}_0\|_1 ~ \\
 	&\hspace{0.2in} + \langle\b{UV^\top} + \b{W}, ~\b{DH}\rangle - \langle\lambda_e \text{sign}(\b{S}_0) + \lambda_e \b{F}, ~\b{H}\rangle.
 	\end{align}
 	Next, we use the relationships derived above to simplify the following term:
 	\begin{align*}
 	&\langle\b{UV^\top} + \b{W}, ~\b{DH}\rangle - \langle\lambda_e \text{sign}(\b{S}_0) + \lambda_e \b{F}, ~\b{H}\rangle\notag \\
 	&=\langle\b{W}, \b{DH}\rangle - \lambda_e \langle\b{F}, \b{H}\rangle + \langle\c{P}_{\c{L}}(\b{\Lambda}), \b{DH}\rangle \\
 	& \hspace{2.1in} - \langle\c{P}_{\c{S}_e}(\b{D^\top\Lambda}), \b{H}\rangle, \\
 	&=\|\c{P}_{\c{L}^\perp}(\b{DH})\|_* + \lambda_e \|\c{P}_{\c{S}_e^\perp}(\b{H})\|_1 + \langle\c{P}_{\c{L}}(\b{\Lambda}), \b{DH}\rangle \\
 	&\hspace{2.1in}- \langle\c{P}_{\c{S}_e}(\b{D^\top\Lambda}), \b{H}\rangle.
 	\end{align*}
 	
 	We now simplify \[\langle\c{P}_{\c{L}}(\b{\Lambda}), ~\b{DH}\rangle -  \langle\c{P}_{\c{S}_e}(\b{D^\top\Lambda}), ~\b{H}\rangle\] using Holder's inequality.
 	\begin{align*}
 	&\langle\c{P}_{\c{L}}(\b{\Lambda}), \b{DH}\rangle -  \langle\c{P}_{\c{S}_e}(\b{D^\top\Lambda}), \b{H}\rangle \notag\\
 	& = \langle\b{\Lambda}-\c{P}_{\c{L}^{\perp}}(\b{\Lambda}), ~\b{DH}\rangle - \langle\b{D^\top\Lambda} - \c{P}_{\c{S}_e^{\perp}}(\b{D^\top\Lambda}), \b{H}\rangle \notag\\
 	& \geq \hspace{-2pt}-\|\c{P}_{\c{L}^\perp}(\b{DH})\|_*\|\c{P}_{\c{L}^\perp}(\b{\Lambda})\| \hspace{-2pt}-\hspace{-2pt} \|\c{P}_{\c{S}_e^\perp}(\b{D^\top\Lambda})\|_\infty\|\c{P}_{\c{S}_e^\perp}(\b{H})\|_1.
 	\end{align*}
 	%
 	Finally, we simplify the optimality condition in shown in \eqref{subopt}, 
 	\begin{align*}
 	&\|\b{L}_0 + \b{DH}\|_* ~+~ \lambda_e~\|\b{S_0 - H}\|_1\\
 	& \geq~ \|\b{L}_0\|_* + \lambda_e~\|\b{S}_0\|_1 ~+  (1 - \|\c{P}_{\c{L}^\perp}(\b{\Lambda})\|)\|\c{P}_{\c{L}^\perp}(\b{DH})\|_*  \\
 	&\hspace{0.4in}+(\lambda_e - \|\c{P}_{\c{S}_e^\perp}(\b{D^\top\Lambda})\|_\infty) \|\c{P}_{\c{S}_e^\perp}(\b{H})\|_1.
 	\end{align*}
 	Here, we note that if  \[\|\c{P}_{\c{L}^\perp}(\b{\Lambda})\| < 1\] and \[ \|\c{P}_{\c{S}_e^\perp}(\b{D^\top\Lambda})\|_\infty < \lambda_e\], then the pair \[\{\b{L}_0, \b{S}_0\}\] is the unique solution of the problem.
 	%
 	Consequently, these are the required necessary conditions \textbf{(C3)} and \textbf{(C4)}, respectively. 
 \end{proof} 
  \vspace{5pt}
  \begin{proof}[\hspace{-23pt} Proof of Lemma \ref{lower_sigmaMin}]
 	First, note that we need \[\b{A}_{\c{S}_e}\] to have full row rank, i.e, its smallest singular value should be greater than zero. To this end, we first derive a lower bound on the smallest singular value, \[\sigma_{\min}{(\b{A}_{\c{S}_e})} \] of \[\b{A}_{\c{S}_e}\] as follows:
 	\begin{align*}
 	\sigma_{\min}{(\b{A}_{\c{S}_e})} & = \underset{\b{H} \in {\c{S}_e} \backslash  \{\b{0}\}}{\min} ~\tfrac{\|\b{A}^\top \text{vec}(\b{H})\|}{\|\text{vec}(\b{H})\|}.
 	\end{align*} 
 	Now, using the definition of \[\b{A}^\top\] and properties of Kronecker products namely, transpose and vectorization of product of three matrices, we have
 	\begin{align*}
 	\sigma_{\min}{(\b{A}_{\c{S}_e})} 
 	&=  \underset{\b{H} \in {\c{S}_e} \backslash \{\b{0}\}}{\min} \tfrac{\|(\b{I} - \b{P_U}) \b{DH} (\b{I} - \b{P_V})\|_{\Fr}}{\|\b{H}\|_{\Fr}}.
 	\end{align*} 
 	Now, since \[(\b{I} - \b{P_U}) \b{DH} (\b{I} - \b{P_V}) = \c{P}_{\c{L}^\perp}(\b{DH})\], 
 	\begin{align*}
 	\sigma_{\min}{(\b{A}_{\c{S}_e})} 
 	&= \underset{\b{H} \in {\c{S}_e} \backslash \{\b{0}\}}{\min} \tfrac{\|\c{P}_{\c{L}^\perp}(\b{DH})\|_{\Fr}}{\|\b{DH}\|_{\Fr}} \tfrac{\|\b{DH}\|_{\Fr}}{\|\b{H}\|_{\Fr}}.
 	\end{align*} 
 	Using the GFP, we have the following lower bound:
 	\begin{align*}
 	\sigma_{\min}{(\b{A}_{\c{S}_e})} 
 	&\geq \sqrt{\alpha_\ell}~\underset{\b{Z} \in {\c{D}} \backslash \{\b{0}\}}{\min} \tfrac{\|\c{P}_{\c{L}^\perp}(\b{Z})\|_{\Fr}}{\|\b{Z}\|_{\Fr}}.
 	\end{align*} 
 	Further, simplifying using properties of the projection operator, the reverse triangle inequality and the definition of \[\mu\],
 	\begin{align*}
 	\sigma_{\min}{(\b{A}_{\c{S}_e})} 
 	&= \sqrt{\alpha_\ell}~\underset{\b{Z} \in {\c{D}} \backslash \{\b{0}\}}{\min} \tfrac{\|\b{Z} - \c{P}_{\c{L}}(\b{Z})\|_{\Fr}}{\|\b{Z}\|_{\Fr}}\\	
 	&\geq \sqrt{\alpha_\ell}~\big( 1 - \underset{\b{Z} \in {\c{D}} \backslash \{\b{0}\}}{\max} \tfrac{\|\c{P}_{\c{L}}(\b{Z})\|_{\Fr}}{\|\b{Z}\|_{\Fr}}\big) = \sqrt{\alpha_\ell} (1- \mu).
 	\end{align*} 
 	Therefore, we note that if \[\mu <1\] and \[\alpha_\ell >0 \], \[\b{A}_{\c{S}_e}\] has full row rank, and the lower bound on the smallest singular value is given by \[\sqrt{\alpha_\ell} (1- \mu)\].
 \end{proof}
 
   \vspace{5pt}
   \begin{proof}[\hspace{-23pt} Proof of Lemma~\ref{upper_bOmega}]
 	We begin with the definition of \[\b{b}_{\c{S}_e}\]. Since \[\|\b{b}_{\c{S}_e}\|_2 = \|\b{B}_{\c{S}_e}\|_{\Fr}\] and \[\b{B_{\c{S}_e}} :=\lambda_e \text{sign}(\b{S}_0) - \mathcal{P}_{\c{S}_e} (\b{D^\top UV^\top})\],
 	\begin{align*}
 	\|\b{b}_{\c{S}_e}\|_2 &= \|\lambda_e \text{sign}(\b{S}_0) - \c{P}_{\c{S}_e} (\b{D^\top UV^\top })\|_{\Fr}, \\
 	&\leq \lambda_e \sqrt{s_e} + \|\c{P}_{\c{S}_e} (\b{D^\top UV^\top })\|_{\Fr}.	
 	\end{align*}
 	Now for an upper bound on \[\|\c{P}_{\c{S}_e} (\b{D^\top UV^\top })\|_{\Fr}\] we start by analyzing \[\|\c{P}_{\c{S}_e} (\b{D^\top UV^\top })\|_{\Fr}^2\], 	
 	\begin{align*}
 	\|\c{P}_{\c{S}_e} (\b{D^\top UV^\top })\|_{\Fr}^2 
 	& = |\langle\b{D^\top UV^\top }, ~\c{P}_{\c{S}_e} (\b{D^\top UV^\top })\rangle|.
 	\end{align*}
 	Using properties of the inner products and using the fact that \[\c{P}_{\c{L}} (\b{UV^\top }) = \b{UV^\top} \],
 	\begin{align*}
 	\|\c{P}_{\c{S}_e} (\b{D^\top UV^\top })\|_{\Fr}^2 
 	& = |\langle\c{P}_{\c{L}} (\b{UV^\top }), ~\b{D}\c{P}_{\c{S}_e} (\b{D^\top UV^\top })\rangle|.
 	\end{align*}
 	Further simplifying using Cauchy Schwarz inequality and the definition of \[\mu\] we have
 	\begin{align*}
 	\|\c{P}_{\c{S}_e} (&\b{D^\top UV^\top })\|_{\Fr}^2 \\
 	&\leq \|\c{P}_{\c{L}} (\b{UV^\top })\|_{\Fr} \|\c{P}_{\c{L}} (\b{D}\c{P}_{\c{S}_e} (\b{D^\top UV^\top }))\|_{\Fr}\\
 	&\leq \mu\|\b{UV^\top }\|_{\Fr} \|\b{D}\c{P}_{\c{S}_e} (\b{D^\top UV^\top })\|_{\Fr}
 	\end{align*}
 	Now, since \[\|\b{UV^\top }\|_{\Fr} = \sqrt{r}\] and using the GFP we have $\|\c{P}_{\c{S}_e} (\b{D^\top UV^\top })\|_{\Fr} \leq  \mu\sqrt{r\alpha_u}$. Therefore, an upper bound for \[\|\b{b}_{\c{S}_e}\|_2\] is given by $\|\b{b}_{\c{S}_e}\|_2 \leq \lambda_e \sqrt{s_e} + \sqrt{r \alpha_u} \mu$.
 \end{proof}
 
 
  \vspace{5pt}
  \begin{proof}[\hspace{-24pt} Proof of Lemma~\ref{lbOmegaInf}]
 	Since \[ \|\b{b}_{\c{S}_e}\|_\infty =	 \|\b{B}_{\c{S}_e}\|_\infty\] and \[\b{B_{\c{S}_e}} :=\lambda_e \text{sign}(\b{S}_0) - \mathcal{P}_{\c{S}_e} (\b{D^\top UV^\top})\], we have the upper bound $\|\b{b}_{\c{S}_e}\|_\infty 
 	\leq  \lambda_e  + \|\c{P}_{\c{S}_e} (\b{D}^\top\b{UV}^\top)\|_\infty$.
 \end{proof}
 

  \vspace{5pt}
  \begin{proof}[\hspace{-23pt} Proof of Lemma~\ref{Q_inf}]
 	We begin by simplifying the quantity of interest as follows:
 	\begin{align}
 	\label{Q_inf_eq}
 	\|\b{Q}\|_{\infty, \infty} &= \|\b{A}_{{\c{S}_e}^\perp} \b{A}_{\c{S}_e}^\top  (\b{A}_{\c{S}_e}\b{A}_{\c{S}_e}^\top )^{-1}\|_{\infty, \infty} \notag \\
 	&\leq \|\b{A}_{{\c{S}_e}^\perp} \b{A}_{\c{S}_e}^\top \|_{\infty, \infty} \|( \b{I} - (\b{I} - \b{A}_{\c{S}_e}\b{A}_{\c{S}_e}^\top ))^{-1}\|_{\infty, \infty} \notag \\
 	&\leq \tfrac{\|\b{A}_{{\c{S}_e}^\perp} \b{A}_{\c{S}_e}^\top \|_{\infty, \infty} }{1 - \|\b{I} - \b{A}_{\c{S}_e}\b{A}_{\c{S}_e}^\top \|_{\infty, \infty}}.
 	\end{align}
 	
 	Now, we derive appropriate bounds on the numerator and the denominator of \eqref{Q_inf_eq} separately. 
 	Consider the numerator \[{\|\b{A}_{{\c{S}_e}^\perp} \b{A}_{\c{S}_e}^\top \|_{\infty, \infty} }\]. Here, we are interested in the maximum \[\ell_1\]-norm of the rows of \[\b{A}_{{\c{S}_e}^\perp} \b{A}_{\c{S}_e}^\top\], i.e., 
 	\begin{align*}
 	\|\b{A}_{{\c{S}_e}^\perp} \b{A}_{\c{S}_e}^\top \|_{\infty, \infty} & = \underset{i}{\max}~\|\b{e}^\top _i\b{A}_{{\c{S}_e}^\perp} \b{A}_{\c{S}_e}^\top \|_1.
 	\end{align*}
 	Let \[\c{I}_{\c{S}_e }\] refer to the support of \[\b{S}_0\], and  \[\bar{\c{I}}_{\c{S}_e}\] to its complement. Then, the expression can be written in terms of \[\c{I}_{\c{S}_e }\] and \[\bar{\c{I}}_{\c{S}_e}\]:
 	\begin{align*}
 	\|\b{A}_{{\c{S}_e}^\perp} \b{A}_{\c{S}_e}^\top \|_{\infty, \infty} 
 	&=\underset{j\in \bar{\c{I}}_{\c{S}_e}}{\max}~\textstyle\sum\limits_{\ell \in \c{I}_{\c{S}_e }} |\b{e}_l^\top\b{A}\b{A}^\top \b{e}_j|.
 	\end{align*}
 	Now, \[\b{A}\] is defined as \[\b{(I - P_V)} \otimes \b{D^\top}(\b{I- P_U})\], therefore using the property of the product of two Kronecker products and product of projection matrices, \[\b{AA}^\top\] can be written as 
 	\begin{align*}
 	\b{A}\b{A}^\top 
 	=  \b{(I - P_V)} \otimes \b{D^\top}(\b{I- P_U})\b{D}.
 	\end{align*}
 	We are interested in the \[\{\ell, j\}\] entry of \[\b{AA}^\top\]. Since, \[\b{AA}^\top\] has a Kronecker product structure, an entry of  \[\b{AA}^\top\] is given by the product of elements of the matrices in the Kronecker product, therefore
 	\begin{align}
 	\label{sumG}
 	\hspace{-5pt}&\underset{j\in \bar{\c{I}}_{\c{S}_e}}{\max} \hspace{-2pt}\textstyle\sum\limits_{\ell \in \c{I}_{\c{S}_e }}\hspace{-3pt} |\b{e}_l^\top\b{A}\b{A}\hspace{-2pt}^\top \b{e}_j|
 	\hspace{-3pt}=\hspace{-8pt}\underset{j_1,j_2 \in \bar{\c{I}}_{\c{S}_e}}{\max}\hspace{-2pt} \textstyle\sum\limits_{\ell_1,\ell_2 \in \c{I}_{\c{S}_e }} \hspace{-6pt}g(j_1, j_2, \ell_1, \ell_2),
 	\end{align}
 	where {$g(j_1, j_2, \ell_1, \ell_2)$} is given by
 	\begin{align*}
 	g(j_1, &j_2, \ell_1, \ell_2) \\
 	&= |{\rm Tr}(\b{e}_{\ell_2}\b{e}_{\ell_1}^\top\b{D^\top}(\b{I- P_U})\b{D} \b{e}_{j_1}\b{e}_{j_2}^\top\b{(I - P_V)})|.
 	\end{align*}
 	Now, consider \[g(j_1, j_2, \ell_1, \ell_2)\], which can be simplified as
 	\begin{align*}
 	g(j_1,j_2, \ell_1, &\ell_2) \\
 	&= |{\rm Tr}(\b{e}_{\ell_2}\b{e}_{\ell_1}^\top\b{D^\top}(\b{I- P_U})\b{D} \b{e}_{j_1}\b{e}_{j_2}^\top) \\ 
 	&\hspace{0.3in}- {\rm Tr}(\b{e}_{\ell_2}\b{e}_{\ell_1}^\top\b{D^\top}(\b{I- P_U})\b{D} \b{e}_{j_1}\b{e}_{j_2}^\top\b{P_V})|.
 	\end{align*}
 	Since trace is invariant under cyclic permutations, we have
 	\begin{align*}
 	g(j_1, j_2, \ell_1, \ell_2) 
 	&= |\b{e}_{\ell_1}^\top\b{D^\top}(\b{I- P_U})\b{D} \b{e}_{j_1}  \mathbbm{1}_{\{j_2 = \ell_2\}} \\ &~~~~~~~~~
 	- \b{e}_{\ell_1}^\top\b{D^\top}(\b{I- P_U})\b{D} \b{e}_{j_1}\b{e}_{j_2}^\top\b{P_V}\b{e}_{\ell_2}|.
 	\end{align*}
 	Denote \[x := \b{e}_{\ell_1}^\top\b{D^\top}(\b{I- P_U})\b{D} \b{e}_{j_1}\] and \[y := \b{e}_{j_2}^\top\b{P_V}\b{e}_{\ell_2}\], then we have
 	\vspace{-7pt}
 	\begin{align*}
 	g(j_1, j_2, \ell_1, \ell_2) &= |x  \mathbbm{1}_{\{j_2 = \ell_2\}} - xy|.
 	\end{align*}
 	Now, the following upper bound on \[ g(j_1, j_2, \ell_1, \ell_2)\] can be evaluated by squaring both sides and simplifying
 	\begin{align}
 	\label{gjl}
 	g(j_1, j_2, \ell_1, \ell_2) &\leq x \sqrt{\mathbbm{1}_{\{j_2 = \ell_2\}} + y^2}.
 	\end{align}
 	First consider \[x\], which can be written as $x = x\mathbbm{1}_{\{j_1 = \ell_1\}} + x\mathbbm{1}_{\{j_1 \neq \ell_1\}}$.
 	Here, \[x\mathbbm{1}_{\{j_1 = \ell_1\}}\] can be upper bounded as shown below using the GFP
 	\begin{align*}
 	x= (\b{e}_{\ell_1}^\top \b{D}^\top (\b{I} - \b{P_U}) \b{D}\b{e}_{\ell_1})
 	\leq \b{e}_{\ell_1}^\top \b{D}^\top \b{D}\b{e}_{\ell_1} 
 	\leq \alpha_u.
 	\end{align*}
 	Further,  we can derive an upper bound on \[ x\mathbbm{1}_{\{j_1 \neq \ell_1\}}\] using the paraflelogram law for inner-products as follows.
 	\begin{align*}
 	~~~x~&\leq  |\b{e}_{j_1}^\top \b{D}^\top \b{D}\b{e}_{\ell_1}| + |\b{e}_{j_1}^\top \b{D}^\top \b{P_U} \b{D}\b{e}_{\ell_1}|\\
 	&\leq  \tfrac{\alpha_u -  \alpha_\ell}{2} +  \alpha_u \gamma_{\b{U}}  = \tfrac{ \alpha_u(1 + 2{ \gamma_{\b{U}}} ) }{2} - \tfrac{\alpha_\ell}{2}.
 	\end{align*}
 	%
 	Therefore, we have 
 	\begin{align*}
 	x \leq  \alpha_u\mathbbm{1}_{\{j_1 = \ell_1\}} + ( \tfrac{ \alpha_u(1 + 2 \gamma_{\b{U}} ) }{2} - \tfrac{\alpha_\ell}{2})\mathbbm{1}_{\{j_1 \neq \ell_1\}}.
 	\end{align*}

 	\noindent
 	Now, consider \[\sqrt{\mathbbm{1}_{\{j_2 = \ell_2\}} + y^2}\], since \[ y = \b{e}_{j_2}^\top\b{P_V}\b{P_V}\b{e}_{\ell_2}\], and further, since \[\sqrt{a^2 + b^2} < (a+b) \text{~for~} a>0 \text{~and~} b>0\], we have $\sqrt{\mathbbm{1}_{\{j_2 = \ell_2\}} + y^2} 
 	\leq \mathbbm{1}_{\{j_2 = \ell_2\}} +  \gamma_{\b{V}}.$
 	Now, substituting in (\ref{gjl}), i.e., the expression for \[g(j_1, j_2, \ell_1, \ell_2)\], we have,
 	\begin{align*}
 	& g(j_1, j_2, \ell_1, \ell_2) \leq\\
 	&(\alpha_u\mathbbm{1}_{\{j_1 = \ell_1\}} + ( \tfrac{ \alpha_u(1 + 2 \gamma_{\b{U}} ) }{2} - \tfrac{\alpha_\ell}{2})\mathbbm{1}_{\{j_1 \neq \ell_1\}} )(\mathbbm{1}_{\{j_2 = \ell_2\}} +  \gamma_{\b{V}}),
 	\end{align*}
 	and finally substituting in (\ref{sumG}) and noting that since \[j_1, j_2 \in \bar{\c{I}}_{\c{S}_e}\] and \[\ell_1, \ell_2 \in \bar{\c{I}}_{\c{S}_e}\],  \[ \mathbbm{1}_{\{j_1 = \ell_1\}}\mathbbm{1}_{\{j_2 = \ell_2\}} = 0\], 
 	{\small\begin{align}
 	&\|\b{A}_{{\c{S}_e}^\perp} \b{A}_{\c{S}_e}^\top \|_{\infty, \infty} \leq \underset{j_1,j_2 \in {\bar{\c{I}}_{\c{S}_e}}}{\max}\textstyle\sum\limits_{ \ell_1,\ell_2 \in \c{I}_{\c{S}_e }} 
 	\hspace{-5pt}( \tfrac{ \alpha_u(1 + 2 \gamma_{\b{U}} ) }{2} - \tfrac{\alpha_\ell}{2})\mathbbm{1}_{\substack{\{j_1 \neq \ell_1\},\\ \{j_2 = \ell_2\}}} \notag \\
 	& 
 	\hspace{0.1in} +  \alpha_u \gamma_{\b{V}}\mathbbm{1}_{\{j_1 = \ell_1\}} +  
 	( \tfrac{ \alpha_u(1 + 2 \gamma_{\b{U}} ) \gamma_{\b{V}} }{2} - \tfrac{\alpha_\ell \gamma_{\b{V}}}{2})\mathbbm{1}_{\{j_1 \neq \ell_1\}}.
 	\end{align}}
 	Now, for \[\b{A}_0 \in \mathbb{R}^{d \times m}\], the maximum number of non-zeros per row is \[\text{min}(s_e, m)\], while those in a column are \[\text{min}(s_e, d)\] for the \textit{thin} case and \[\text{min}(s_e, k)\] for the \textit{fat} case. Then we have
 	\begin{align}\label{bound_num_AScAS}
 	\|\b{A}_{{\c{S}_e}^\perp} \b{A}_{\c{S}_e}^\top \|_{\infty, \infty} \leq c.
 	%
 	\end{align}
 	Here, the constant \[c\] is as defined in \eqref{eqconst:c_entry}.
%
 	Now, to bound  the denominator of \eqref{Q_inf_eq}, we have 
 	\begin{align}
 	\label{IminAinfinf}
 	&\|\b{I} - \b{A}_{\c{S}_e}\b{A}_{\c{S}_e}^\top \|_{\infty, \infty} = \underset{i}{\max} \|\b{e}_i^\top (\b{I} - \b{A}_{\c{S}_e} \b{A}^\top _{\c{S}_e})\|_1 \notag \\
 	&= \underset{j,\ell \in \c{S}}{\max}|1 - \|\b{e}_j^\top \b{A}\|^2| + \textstyle\sum\limits_{j\neq \ell} |\langle \b{e}_j^\top \b{A}, \b{e}_l^\top \b{A} \rangle|
 	\end{align}
 	We proceed to bound \[|1 - \|\b{e}_j^\top \b{A}\|^2|\]. For this, we derive a lower bound on \[\|\b{e}_j^\top \b{A}\|^2\]. Note that \[\b{e}_j^\top \b{A}\] selects the \[j\]-th row of \[\b{A}\], which has a Kronecker product structure. Therefore,
 	\begin{align*}
 	\|\b{e}_j^\top \b{A}\| 
 	& = \resizebox{0.4\textwidth}{0.015\textwidth}{$ \|(\b{I} - \b{P_U})\b{D}\b{e}_{j_1}\b{e}_{j_2}^\top (\b{I} - \b{P_V})\|_{\Fr}  = \|\c{P}_{\c{L}^\perp}(\b{D}\b{e}_{j_1}\b{e}_{j_2}^\top )\|_{\Fr} $}  \\
 	&\geq \|\b{D}\b{e}_{j_1}\b{e}_{j_2}^\top \| - \|\c{P}_{\c{L}}(\b{D}\b{e}_{j_1}\b{e}_{j_2}^\top )\|_{\Fr} \notag 
 	\geq \sqrt{ \alpha_\ell }(1 - \mu).
 	\end{align*}
 	Therefore, since \[\mu < 1\] and \[\alpha_\ell > 0\], then if \[\alpha_\ell \leq \tfrac{1}{(1-\mu)^2}\], we have $|1 - \|\b{e}_j^\top \b{A}\|^2| \leq 1-  \alpha_\ell(1 - \mu)^2$.
 	The analysis for deriving an upper bound for the second term in \eqref{IminAinfinf} closely follows that used in \eqref{bound_num_AScAS}, as shown below
 	\begin{align*}
 	\textstyle\sum\limits_{j\neq \ell} |\langle \b{e}_j^\top \b{A}_{\c{S}_e}, \b{e}_l^\top \b{A}_{\c{S}_e} \rangle| 
 	&= \hspace{-0.1in}\textstyle\sum\limits_{ \substack{(\ell_1, \ell_2) \in \c{S} \backslash \{(j_1, j_2)\}} } \hspace{-0.1in} g(j_1, j_2, \ell_1, \ell_2) \leq c.
 	\end{align*}
 	Combining these results, we have the following bound for 
 	\begin{align*}
 	\|\b{I} - \b{A}_{\c{S}_e}\b{A}_{\c{S}_e}^\top \|_{\infty, \infty} \leq 1 - \alpha_\ell(1 - \mu)^2 + c.
 	\end{align*}
 	Finally, substituting these results in \eqref{Q_inf_eq} we have $
 	\|\b{Q}\|_{\infty, \infty} 
 	\leq C_e := \tfrac{c}{\alpha_\ell(1 - \mu)^2 - c}$,
 	where \[c\] is given by \eqref{eqconst:c_entry}. 
 \end{proof}
  
  \subsection{Proofs for Column-wise Case}
  \label{app:col}
    \vspace{5pt}
    \begin{proof}[\hspace{-23pt} Proof of Lemma \ref{lem:unique}]\label{pf:lem:unique}
  We show that for any \[(\b{L}_0, \b{S}_0) \in \{ \b{M}, \c{U}, \c{I}_{\c{S}_c} \}\], if \[\spann\{\col(\b{L}_0) \} = \c{U}$ and $\csupp(\b{D} \b{S}_0) = \c{I}_{\c{S}_c}\] do not hold simultaneously, then $\mu = 1$.
  
  \noindent
  Let \[\b{L} + \b{DS} = \b{M}\], as per our model shown in \eqref{Prob}.  
  Now, let \[(\b{L}_0, \b{S}_0)\] be any other pair in our Oracle Model \[ \{ \b{M}, \c{U}, \c{I}_{\c{S}_c} \}\], 
  \begin{align*}
  \b{L}_0 = \b{L} + \b{\Delta}_1 \in \c{U}~~\text{and}~~\b{D}\b{S}_0 = \b{D}\b{S} + \b{\Delta}_2 \in \c{S}_c,
  \end{align*}
  for some \[\b{\Delta}_1\] and \[\b{\Delta}_2\], then we have that $\b{\Delta}_1 + \b{\Delta}_2 = \b{0}$.
  This implies that \[\csupp(\b{\Delta}_1) \in \c{S}_c\]. Further, this implies that \[\b{L}\] and \[\b{L}_0\] at least match in the columns indexed by the inliers, i.e., $\c{P}_{\c{I}_{\b{L}}}(\b{L}) = \c{P}_{\c{I}_{\b{L}}}(\b{L}_0)$, and we have
 \begin{align*}
 \c{U}&=\spann\{\col(\b{L}_0) \} = \spann\{\col(\c{P}_{\c{I}_{\b{L}}}(\b{L}_0)) \} \\
 &= \spann\{\col(\c{P}_{\c{I}_{\b{L}}}(\b{L})) \}.
 \end{align*}
 Therefore, \[\csupp( \b{D}\b{S}_0) \subseteq  \c{I}_{\c{S}_c} \]. Specifically, this implies that there may exist a \[j \in \c{I}_{\c{S}_c} \] for which $\b{D} \b{S}_{:,j} - (\b{\Delta}_1)_{:,j} = 0$, which will imply that  $\c{P}_{\c{U}^\perp}(\b{D} \b{S}_{:,j}) = 0$. This condition implies that \[\mu = 1\]. Therefore, we require \[\spann\{\col(\b{L}_0) \} = \c{U}\] and \[\csupp(\b{D} \b{S}_0) = \c{I}_{\c{S}_c}\] to hold simultaneously for \[\mu < 1\].
\end{proof}

\vspace{5pt}
  \begin{proof}[\hspace{-23pt} Proof of Lemma ~\ref{lem:dualcertify}] Let \[(\b{L}_0,\b{S}_0)\] be an optimal solution pair of \eqref{Pc}. From the optimality conditions \eqref{eqn:optL} and \eqref{eqn:optC}, we seek \[\b{\Lambda}\] such that 
	\begin{align}
	\b{\Lambda} \in  \b{U} \b{V}^\top + \b{W}~~\text{and}~~\b{\b{D}^\top\Lambda} \in \lambda_c \b{H}+ \lambda_c \b{F}.  \label{eqn:subdiff}
	\end{align}  
	Now consider a feasible solution \[\{\b{L_0 + D\Delta}, \b{S_0 - \Delta}\}\] for a non-zero \[\b{\Delta} \in \mathbb{R}^{d\times m} \]. Then by the optimality of \[(\b{L}_0,\b{S}_0)\] using the subgradient inequality, we have
	\begin{align*}
	\|\b{L}_0 + \b{D} \b{\Delta} \|_* + \lambda_c \| \b{S}_0 - \b{\Delta} \|_{1,2} &\geq \| \b{L}_0 \|_* + \lambda_c \| \b{S}_0 \|_{1,2} \\
	&\hspace{-0.8in}+ \langle \b{U} \b{V}^\top + \b{W},\b{D} \b{\Delta} \rangle - \lambda_c \langle \b{H} + \b{F}, \b{\Delta} \rangle.
	\end{align*}
	Let \[G(\b{\Delta}) = \langle \b{U} \b{V}^\top + \b{W},\b{D} \b{\Delta}  \rangle - \lambda_c \langle \b{H} + \b{F}, \b{\Delta}  \rangle\]. We will show that if \textbf{(q1)}-\textbf{(q4)} hold, then \[G(\b{\Delta})>0\], which proves the optimality of \[(\b{L}_0,\b{S}_0)\]. Rewrite \[G(\b{\Delta})\] as
	\begin{align}
	G(\b{\Delta}) = \langle \b{W}, \b{D} \b{\Delta} \rangle - \lambda_c \langle \b{F}, \b{\Delta} \rangle + \langle \b{D}^\top \b{U} \b{V}^\top - \lambda_c \b{H}, \b{\Delta} \rangle.\label{eqn:GDelta}
	\end{align}
	Let \[\b{W}\], with \[\|\b{W}\| = 1\] and \[\c{P}_{\c{L}} (\b{W}) = \b{0}\], then by duality of norms,
	\begin{align}\label{eqn:GDelta1}
	\langle\b{W},\b{D\Delta}\rangle = \langle\b{W},\c{P}_{\c{L}^\perp}(\b{D\Delta}) \rangle = \|\c{P}_{\c{L}^\perp}(\b{D\Delta})\|_*.
	\end{align}
	Further, let \[\b{F}\], with \[\|\b{F}\|_{\infty,2} = 1\] and \[\c{P}_{\c{S}_c} (\b{F}) = \b{0}\], be such that
	\begin{align*}
	\b{F}_{{:,j}} = \begin{cases}
	-\tfrac{\b{\Delta}_{:,j}}{\|\b{\Delta}_{:,j}\|},& \text{ if } j \not\in \c{I}_{\c{S}_c}  \text{ and } \b{\Delta}_{:,j} \neq 0\\
	0,& \text{ otherwise}
	\end{cases},
	\end{align*}
	\noindent where \[\b{F}_{{:,j}}\] denotes the \[j^{\text{th}}\] column of \[\b{F}\]. Then, we arrive at the following simplification for \[\langle\b{F},~\b{\Delta}\rangle\] by duality of norms,
	\begin{align}\label{eqn:GDelta2}
	\langle\b{F},~\b{\Delta}\rangle ~=~ \langle\b{F},~\c{P}_{\c{S}_c^\perp}(\b{\Delta}) \rangle ~=~ -\|\c{P}_{\c{S}_c^\perp}(\b{\Delta})\|_{1,2}.
	\end{align}
	%
	Since \[\c{P}_{\c{L}} (\b{\Lambda}) = \b{UV}^\top\] and \[ \c{P}_{\c{S}_c} (\b{D}^\top\b{\Lambda}) = \lambda_c \b{H}\] by optimality conditions of \eqref{eqn:subdiff},
	\begin{align}
	&\langle \b{D}^\top \b{U} \b{V}^\top - \lambda_c \b{H}, \b{\Delta} \rangle \\
	&= -\langle \c{P}_{\c{L}^\perp} (\b{\Lambda}), \b{D\Delta}\rangle + \langle \c{P}_{\c{S}_c^\perp} (\b{D}^\top\b{\Lambda}) , \b{\Delta} \rangle \notag\\
	&\geq - \|\c{P}_{\c{L}^\perp} (\b{D\Delta})\|_*\|\c{P}_{\c{L}^\perp} (\b{\Lambda})\| \nonumber\\
	&\hspace{1in}- \|\c{P}_{\c{S}_c^\perp} (\b{\Delta}) \|_{1,2} \|\c{P}_{\c{S}_c^\perp} (\b{D}^\top\b{\Lambda}) \|_{\infty,2}, \label{eqn:GDelta3}
	\end{align}
	where we use Holder's inequality in the last step.
	
	Combining \eqref{eqn:GDelta}, \eqref{eqn:GDelta1}, \eqref{eqn:GDelta2}, and \eqref{eqn:GDelta3}, we have
	\begin{align*}
	G(\b{\Delta}) 
	&\geq (1-\|\c{P}_{\c{L}^\perp} (\b{\Lambda})\|)\|\c{P}_{\c{L}^\perp} (\b{D\Delta})\|_* \\
	&\hspace{0.2in}+ ( \lambda_c- \|\c{P}_{\c{S}_c^\perp} (\b{D}^\top\b{\Lambda}) \|_{\infty,2})\|\c{P}_{\c{S}_c^\perp}(\b{\Delta}) \|_{1,2}
	\end{align*}
	Since we have an arbitrary \[\b{\Delta}\] with \[\b{\Delta} \neq \b{0}\] and \[(\b{L}_0+\b{D} \b{\Delta}, \b{S}_0 - \b{\Delta}) \notin \{ \c{U}, \c{I}_{\c{S}_c}  \}\], \[\| \c{P}_{\c{L}^\perp}(\b{D} \b{\Delta}) \|_* = \| \c{P}_{{\c{S}_c}^\perp}(\b{\Delta}) \|_{1,2} = 0\] does not hold. Therefore, to ensure the uniqueness of the solution \[(\b{L}_0,\b{S}_0)\], we need \[\|\c{P}_{\c{L}^\perp} (\b{\Lambda})\| < 1\] and \[\|\c{P}_{\c{S}_c^\perp} (\b{D}^\top\b{\Lambda}) \|_{\infty,2}< \lambda_c\]. Hence, any dual certificate which obeys the conditions \textbf{(C1)}-\textbf{(C4)}  guarantees optimality of the solution. 
\end{proof}


  \vspace{5pt}
  \begin{proof}[\hspace{-23pt} Proof of Lemma~\ref{lower_sigmaMin_col}]
	We begin by writing the definition of \[\sigma_{\min}(\b{A}_{\c{S}_c}^\top) \] as
	\begin{align*}
	\sigma_{\min}(\b{A}_{\c{S}_c}^\top) &= \min_{\b{H} \in \c{S}_c / \{\b{0}_{d\times m} \}} \dfrac{\| \b{A}^\top \vect(\b{H}) \|_2}{\| \vect(\b{H}) \|_2}. 
	%
	\end{align*}
	By the definition of \[\b{A}\] and using the property of Kronecker product for multiplication by a vector we have
	\begin{align*}
	\sigma_{\min}(\b{A}_{\c{S}_c}^\top) 
	&= \min_{\b{H} \in \c{S}_c / \{\b{0}_{d\times m} \}} \dfrac{\| \left( \b{I} - \b{P}_{\b{U}} \right) \b{D}   \b{H} \left( \b{I} - \b{P}_{\b{V}} \right) \|_{\rm F}}{\| \b{H} \|_{\rm F}}.
	\end{align*}
	Further \[\left( \b{I} - \b{P}_{\b{U}} \right) \b{D}   \b{H} \left( \b{I} - \b{P}_{\b{V}} \right) = \c{P}_{\c{L}^\perp}(\b{DH})\], and we can write that expression above as follows
	\begin{align*}
	&\sigma_{\min}(\b{A}_{\c{S}_c}^\top) 
	%
	= \min_{\b{H} \in \c{S}_c / \{\b{0}_{d\times m} \}} \dfrac{\| \b{D} \b{H} \|_{\rm F}}{\| \b{H} \|_{\rm F}} \cdot \dfrac{\| (\b{I} - \c{P}_{\c{L}})(\b{D} \b{H}) \|_{\rm F}}{\| \b{D} \b{H} \|_{\rm F}} \notag \\
	&\overset{(i)}{\geq}\sqrt{\alpha_{\ell}}  (1 - \max_{\b{Z} \in \c{D} / \{\b{0}_{n \times m} \}} \dfrac{\| \c{P}_{\c{L}}(\b{Z}) \|_{\rm F}}{\| \b{Z} \|_{\rm F}}) \overset{(ii)}{\geq} \sqrt{\alpha_{\ell}} (1 - \mu). 
	\end{align*}
	Here (i) is due to the GFP condition \textcolor{blue}{\textbf{D.\ref{frame}}} and the reverse triangle inequality, and (ii) from the incoherence property in \eqref{eqn:mu}.
\end{proof}

  \vspace{5pt}
  \begin{proof}[\hspace{-23pt} Proof of Lemma~\ref{upper_bOmega_col}]
	We start by using the correspondence between the vector \[\b{b}_{\c{S}_c}\] and the matrix  \[\b{B}_{\c{S}_c}\], i.e.,
	\begin{align*}
	\| \b{b}_{\c{S}_c}\|_2 &= \| \b{B}_{\c{S}_c} \|_{\rm F} = \| \lambda_c \tilde{\b{S}} - \c{P}_{\c{S}_c}(\b{D}^\top \b{U} \b{V}^\top) \|_{\rm F}.
	\end{align*}
	Now, since \[\tilde{\b{S}}_{:,j} = \b{S}_{:,j} / \| \b{S}_{:,j} \|_2\] for all \[j \in \c{I}_{\c{S}_c}\]; and is \[\b{0}\] otherwise (i.e., when \[j \notin \c{I}_{\c{S}_c}\]), using triangle inequality, we have
	\begin{align}
	\| \b{b}_{\c{S}_c}\|_2 
	 \leq \lambda_c \sqrt{s_c} + \| \c{P}_{\c{S}_c}(\b{D}^\top \b{U} \b{V}^\top) \|_{\rm F}. \label{eqn:vecY1}
	\end{align}
	\noindent 
	Since we have
	{\small\begin{align}
	&\| \c{P}_{\c{S}_c}(\b{D}^\top \b{U} \b{V}^\top) \|_{\rm F}^2 
	\leq  \| \c{P}_{\c{L}}(\b{U} \b{V}^\top) \|_{\rm F} \| \c{P}_{\c{L}}(\b{D} \c{P}_{\c{S}_c}(\b{D}^\top \b{U} \b{V}^\top)) \|_{\rm F} \nonumber \\
	&\overset{(i)}{\leq}\hspace{-2pt} \mu\| \b{U} \b{V}^\top \|_{\rm F} \| \b{D} \c{P}_{\c{S}_c}(\b{D}^\top \b{U} \b{V}^\top) \|_{\rm F} \hspace{-2pt}
	\overset{(ii)}{\leq} \hspace{-2pt}\sqrt{r \alpha_{u}} \mu \| \c{P}_{\c{S}_c}(\b{D}^\top \b{U} \b{V}^\top) \|_{\rm F}, \label{eqn:DUV}
	\end{align}}
	
\noindent where (i) is from subspace incoherence property and (ii) is from the GFP \textcolor{blue}{\textbf{D.\ref{frame}}}. Combining \eqref{eqn:vecY1} and \eqref{eqn:DUV}, we have
	\begin{align*}
	\| \vect(\b{B}_{\c{S}_c}) \|_2 \leq \lambda_c \sqrt{s_c} + \sqrt{r \alpha_{u}}\mu. 
	\end{align*}
\end{proof}
%

  \vspace{5pt}
  \begin{proof}[\hspace{-23pt} Proof of Lemma~\ref{lem:step3}]
	We begin by analyzing the quantity of interest -- \[\|\c{P}_{\c{S}_c^\perp}(\b{Z}) \|_{\infty,2}\], i.e., we are interested in the maximum column norm of the matrix \[\c{P}_{\c{S}_c^\perp}(\b{Z})\]. Note that \[\b{Z}\] is defined as 
	\begin{align*}
	\b{Z} = \b{D}^\top \left( \b{I} - \b{P}_{\b{U}} \right) {\b{X}} \left( \b{I} - \b{P}_{\b{V}} \right),
	\end{align*}
	and we have \[\vect(\b{Z}) = \b{A} \vect(\b{X})\]. Further, we have that
	\begin{align*}
	\c{P}_{\c{S}_c^\perp} \left(\vect(\b{Z}) \right) = \b{A}_{\c{S}_c^\perp}\text{vec}(\b{X}).
	\end{align*}
	Now, observe that the columns of matrix  \[\c{P}_{\c{S}_c^\perp}(\b{Z})\] appear as blocks of size \[n \times 1\] in the vector \[ \c{P}_{\c{S}_c^\perp} \left(\vect(\b{Z}) \right) \]. Moreover, the elements of vector \[ \c{P}_{\c{S}_c^\perp} \left(\vect(\b{Z}) \right) \] are formed due to the inner product between the rows of Kronecker product structured matrix \[\b{A}_{\c{S}_c^\perp}\] and \[\vect(X)\]. Therefore, to identify a column of \[ \c{P}_{\c{S}_c^\perp} \left(\b{Z} \right) \] we need to focus on the interaction between correponding rows of \[\b{A}_{\c{S}_c^\perp}\] and \[\vect(\b{X})\].
	
	Consider the Kronecker product structured matrix \[\b{A}_{\c{S}_c^\perp}\]. Since the rows in \[\b{A}_{\c{S}_c^\perp}\] correspond to all rows outside the column support \[\c{S}_c\], this corresponds to selecting those rows of \[m \times m\] matrix \[ (\b{I}-\b{P}_{\b{V}})\] which correspond to \[\c{S}_c^\perp\], which we denote by \[(\b{I}-\b{P}_{\b{V}})_{\c{S}_c^\perp} \] i.e.,
	\begin{align*}
	\b{A}_{\c{S}_c^\perp} = (\b{I}-\b{P}_{\b{V}})_{\c{S}_c^\perp} \otimes \b{D}^\top(\b{I}-\b{P}_{\b{U}}).
	\end{align*}
	For simplicity of the upcoming analysis, we denote the matrix \[(\b{I}-\b{P}_{\b{V}})\] as 
	\vspace{-0.08in}
	\begin{align*}
	(\b{I}-\b{P}_{\b{V}}) = \resizebox{0.15\textwidth}{!}{$
	\left[ {\begin{array}{*{20}{c}}
		{{v_{11}}} &\cdots & {{v_{1m}}}  \\
		\vdots & \ddots & \vdots \\
		{{v_{m 1}}} &\cdots &{{v_{m m}}}  \\
		\end{array}} \right] $}.
	\end{align*}
Using this notation, the \[j\]-th block of vector \[ \c{P}_{\c{S}_c^\perp} \left(\vect(\b{Z}) \right) \] (which is also the \[j\]-th column of \[ \c{P}_{\c{S}_c^\perp} \left(\b{Z} \right) \]), can be written as 
	\begin{align*}
	\b{Z}_{:,j} = (v_{j,:} \otimes \b{D}^\top(\b{I}-\b{P}_{\b{U}})) \vect(\b{X})
	\end{align*}
	for some \[j \in \c{I}_{\c{S}_c^\perp}\]. Now, further since  \[\vect(\b{X}) := \b{A}_{\c{S}_c}^\top (\b{A}_{\c{S}_c} \b{A}_{\c{S}_c}^\top)^{-1} \vect(\b{B}_{\c{S}_c})\], therefore we are interested in maximum \[2\]-norm of 
	\begin{align*}
	\b{Z}_{:,j} = (v_{j,:} \otimes \b{D}^\top(\b{I}-\b{P}_{\b{U}}))  \b{A}_{\c{S}_c}^\top (\b{A}_{\c{S}_c} \b{A}_{\c{S}_c}^\top)^{-1} \vect(\b{B}_{\c{S}_c}),
	\end{align*}
	for some \[j \in \c{I}_{\c{S}_c^\perp}\].
	Note that \[\b{A}_{\c{S}_c}^\top\] itself is a Kronecker product structured matrix given by
	\begin{align*}
	\b{A}_{\c{S}_c} = (\b{I}-\b{P}_{\b{V}})_{\c{S}_c}^\top \otimes (\b{I}-\b{P}_{\b{U}})\b{D}.
	\end{align*}
	Using the mixed product rule for Kronecker products we have
	\begin{align*}
	\b{Z}_{:,j} = (v_{j,:}(\b{I}-\b{P}_{\b{V}})_{\c{S}_c}^\top \otimes \b{D}^\top(\b{I}-\b{P}_{\b{U}})\b{D}) (\b{A}_{\c{S}_c} \b{A}_{\c{S}_c}^\top)^{-1} \b{b}_{\c{S}_c},
	\end{align*}
	for some \[j \in \c{I}_{\c{S}_c^\perp}\]. 
	\noindent 
	Further, since for two matrices \[\b{A}\] and \[\b{B}\], \[\|\b{A} \otimes \b{B}\| = \|\b{A}\|\|\b{B}\|\], we have 
	\begin{align}
	\|\b{Z}_{:,j}\| &\leq \|{\b{e}_j^\top}(\b{I}-\b{P}_{\b{V}})_{\c{S}_c^\perp}(\b{I}-\b{P}_{\b{V}})_{\c{S}_c}^\top\|\notag\\
	&\hspace{0.2in}\times \|\b{D}^\top(\b{I}-\b{P}_{\b{U}})\b{D}\| \|(\b{A}_{\c{S}_c} \b{A}_{\c{S}_c}^\top)^{-1}\| \|\b{b}_{\c{S}_c}\|, \label{eqn:Z_j}
	\end{align}
	where we also use the fact that \[v_{j,:} = {\b{e}_j^\top}(\b{I}-\b{P}_{\b{V}})_{\c{S}_c^\perp}\]. 
	\noindent
	We will now proceed to bound the first term in \eqref{eqn:Z_j}. Note that
	\begin{align*}
	&\underset{j \in \c{S}_c^\perp }{\max}\|{\b{e}_j^\top}(\b{I}-\b{P}_{\b{V}})_{\c{S}_c^\perp} (\b{I}-\b{P}_{\b{V}})_{\c{S}_c}^\top\|^2 \\
	& = \underset{j \in \c{S}_c^\perp }{\max} \textstyle\sum\limits_{i \in \c{S}_c} \langle (\b{I}-\b{P}_{\b{V}})^\top {\b{e}_j}, (\b{I}-\b{P}_{\b{V}})^\top {\b{e}_i}\rangle^2.
	\end{align*}
	Now, each term in the summation can be bounded as 
	\begin{align*}
&\underset{\substack{j \in \c{S}_c^\perp, i \in \c{S}_c }}{\max}
	|\langle (\b{I}-\b{P}_{\b{V}})^\top {\b{e}_j}, (\b{I}-\b{P}_{\b{V}})^\top {\b{e}_i}\rangle|
	\\
	& = \underset{\substack{j \in \c{S}_c^\perp, i \in \c{S}_c }}{\max}|-\langle\b{P}_{\b{V}} {\b{e}_j}, \b{P}_{\b{V}}{\b{e}_i} \rangle| 
	\leq \| \b{P}_{\b{V}}{\b{e}_j}\|\| \b{P}_{\b{V}}{\b{e}_i}\| \leq \gamma_{\b{V}}.
	\end{align*}
	This implies $\|{\b{e}_j^\top}(\b{I}-\b{P}_{\b{V}})_{\c{S}_c^\perp}(\b{I}-\b{P}_{\b{V}})_{\c{S}_c}^\top\| \leq \sqrt{s_c} \gamma_{\b{V}}$. Further, note that $\|(\b{A}_{\c{S}_c} \b{A}_{\c{S}_c}^\top)^{-1}\| \leq \|\b{A}_{\c{S}_c}^{-1}\|^2 = \tfrac{1}{\sigma_{\min}(\b{A}_{\c{S}_c})^2}$. Substituting this into \eqref{eqn:Z_j}, for a \[j \in \c{S}_c^\perp\], we have
	\begin{align}
	\label{Z_j_1}
	\|\b{Z}_{:,j}\| \leq   \tfrac{\sqrt{s_c} \gamma_{\b{V}}}{\sigma_{\min}(\b{A}_{\c{S}_c})^2}\|\b{D}^\top(\b{I}-\b{P}_{\b{U}})\b{D}\| \|\b{b}_{\c{S}_c}\|.
	\end{align}
	We can further write \[\|\b{D}^\top(\b{I}-\b{P}_{\b{U}})\b{D}\|\] as follows
	\begin{align*}
	\|\b{D}^\top(\b{I}-\b{P}_{\b{U}})\b{D}\| =  \underset{\|u\| = 1}{\max} \tfrac{\|(\b{I}-\b{P}_{\b{U}})\b{D} u\|^2}{\|\b{D}u\|^2}{\|\b{D}u\|^2} \leq \beta_{\b{U}} \alpha_u.
	\end{align*}
	Substituting this result in \eqref{Z_j_1}, using Lemma~\ref{lower_sigmaMin_col} and Lemma~\ref{upper_bOmega_col},
	\begin{align*}
	\|\c{P}_{\c{S}_c^\perp}(\b{Z}) \|_{\infty,2} 
	&\leq   \sqrt{s_c} C_c( \lambda_c \sqrt{s_c} + \sqrt{r \alpha_{u}}\mu).
	\end{align*}

 \end{proof}

\end{document}